\documentclass{article}

\PassOptionsToPackage{sort, numbers}{natbib}

\usepackage[final]{neurips_2021}

\usepackage[utf8]{inputenc} 
\usepackage[T1]{fontenc}    
\usepackage[table,xcdraw,dvipsnames]{xcolor}
\usepackage[hypertexnames=false,colorlinks=true,citecolor=NavyBlue]{hyperref}
\usepackage{url}            
\usepackage{booktabs}       
\usepackage{amsfonts}       
\usepackage{nicefrac}       
\usepackage{microtype}      
\usepackage[pdftex]{graphicx}
\usepackage{amsmath}
\usepackage{xspace}
\usepackage{stmaryrd}
\usepackage[ruled,vlined]{algorithm2e}
\usepackage{mathtools}
\usepackage{tikz}
\usepackage{upgreek}
\usepackage{multirow}
\usepackage{array}
\usepackage{floatrow}
\usepackage{arydshln}
\usepackage{enumitem}
\usepackage{tabularx}
\usepackage[export]{adjustbox}

\newfloatcommand{capbtabbox}{table}[][\FBwidth]

\newcommand{\encoder}{\mathcal{E}}
\newcommand{\decoder}{\mathcal{D}}
\newcommand{\quantizer}{\mathcal{Q}}
\newcommand{\unquantizer}{\mathcal{U}}
\newcommand{\shiftandadd}{\mathcal{S}}

\newcommand{\imagedis}{\Delta_i}
\newcommand{\videodis}{\Delta_t}
\newcommand{\transformer}{\mathcal{T}}
\newcommand{\flowestimator}{\mathcal{F}}
\newcommand{\stimes}{\hspace{-0.15em}\times\hspace{-0.15em}}
\newcommand{\loss}{\mathcal{L}}
\newcommand{\ts}{\textsuperscript}
\newcommand{\ltau}{\scalebox{1.15}{$\uptau$}}
\newcommand{\stau}{\scalebox{0.95}{$\uptau$}}
\newcommand{\method}{CCVS}
\newcommand{\image}{\mathcal{X}}
\newcommand{\feature}{\mathcal{Z}}

\usepackage{pifont}
\usepackage{amssymb}

\newcommand{\revision}[1]{\textcolor{black}{#1}}

\makeatletter
\DeclareRobustCommand\onedot{\futurelet\@let@token\@onedot}
\def\@onedot{\ifx\@let@token.\else.\null\fi\xspace}
\def\eg{\emph{e.g}\onedot} 
\def\ie{\emph{i.e}\onedot} 
\def\RR{\mathbb{R}}

\DeclareMathOperator{\warp}{W}
\DeclareMathOperator{\round}{round}
\DeclareMathOperator{\dilate}{D}
\DeclareMathOperator{\blur}{B}
\DeclareMathOperator{\aug}{A}

\DeclareMathOperator{\sg}{sg}
\DeclareMathOperator{\VGG}{VGG}

\newcolumntype{C}[1]{>{\centering\arraybackslash}p{#1}}	
\newcolumntype{M}[1]{>{\centering\arraybackslash}m{#1}}
\newcolumntype{L}[1]{>{\arraybackslash}p{#1}}
 
\title{CCVS: Context-aware Controllable Video Synthesis}

\author{Guillaume Le Moing\textsuperscript{1, }\thanks{corresponding author: guillaume.le-moing@inria.fr} \hspace{0.8cm} Jean Ponce\textsuperscript{1, 2} \hspace{0.8cm} Cordelia Schmid\textsuperscript{1}\\
\textsuperscript{1}Inria and Department of Computer Science, ENS, CNRS, PSL Research University \\
\textsuperscript{2}Center for Data Science, New York University
}

\begin{document}

\renewcommand*{\thefootnote}{\fnsymbol{footnote}}
\maketitle

\renewcommand*{\thefootnote}{\arabic{footnote}}
\setcounter{footnote}{0} 

\begin{abstract}
  This presentation introduces a self-supervised learning approach to the synthesis of new video clips from old ones, with several new key elements for improved spatial resolution and realism: It conditions the synthesis process on contextual information for temporal continuity and ancillary information for fine control. The prediction model is doubly autoregressive, in the latent space of an autoencoder for forecasting, and in image space for updating contextual information, which is also used to enforce spatio-temporal consistency through a learnable optical flow module. Adversarial training of the autoencoder in the appearance and temporal domains is used to further improve the realism of its output. A quantizer inserted between the encoder and the transformer in charge of forecasting future frames in latent space (and its inverse inserted between the transformer and the decoder) adds even more flexibility by affording simple mechanisms for handling multimodal ancillary information for controlling the synthesis process (\eg, a few sample frames, an audio track, a trajectory in image space) and taking into account the intrinsically uncertain nature of the future by allowing multiple predictions. Experiments with an implementation of the proposed approach give very good qualitative and quantitative results on multiple tasks and standard benchmarks.
\end{abstract}

\section{Introduction}
\label{sec:introduction}

Feeding machines with extensive video content, and teaching them to create new samples on their own, may deepen their understanding of both the physical and social worlds. Video synthesis has numerous applications from content creation (\eg, deblurring, slow motion) to human-robot interaction (\eg, motion prediction). Despite the photo-realistic results of modern image synthesis models~\citep{karras2020analyzing}, video synthesis is still lagging behind due to the increased complexity of the additional temporal dimension.

An emerging trend is to use autoregressive models, for example transformer architectures~\citep{vaswani2017attention}, for their simplicity, and their ability to model long-range dependencies and learn from large volumes of data~\citep{devlin2019bert, brown2020language}. \revision{First introduced for natural language processing (NLP) and then succesfully applied to visual data~\citep{dosovitskiy2021an},} the strength of transformers is grounded in a self-attention mechanism which considers all pairwise interactions within the data. The price to pay is a computational complexity which grows quadratically with the data size, which itself depends linearly on the temporal dimension and quadratically on the spatial resolution in the image domain. Although there have been some efforts to reduce the complexity of self-attention~\citep{kitaev2020reformer, katharopoulos2020transformers, choromanski2021rethinking}, using such methods directly on visual data is still limited to low resolutions and impractical without considerable computational power~\citep{chen2020generative, weissenborn2020scaling}.

Some recent works~\citep{esser2020taming, rakhimov2020latent} address this problem by using an autoencoder to compress the visual data, and apply the autoregressive model in the latent space. This greatly reduces the memory footprint and computational cost, yet, the greater the compression, the harder it is to faithfully reconstruct frames. The corresponding trade-offs may undermine the practical usability of these approaches. GANs~\citep{goodfellow2014generative} mitigate this issue by ``hallucinating'' plausible local details in image synthesis~\citep{esser2020taming}. But latent video transformers~\citep{rakhimov2020latent} decode frames independently, which prevents local details from being temporally coherent. Hence, using GANs in this setting may result in flickering effects in the synthesized videos.

We follow the problem decomposition from~\citep{rakhimov2020latent, esser2020taming}, but introduce a more elaborate compression strategy with \method{} (for \emph{Context-aware Controllable Video Synthesis}), an approach that takes advantage of ``context'' frames (\ie, both input images and previously synthesized ones) to faithfully reconstruct new ones despite lossy compression. As shown in Figure~\ref{fig:skip-ae}, \method{} relies on optical flow estimation between context and new frames, within temporal skip connections, to let information be shared across timesteps. New content, which cannot be retrieved from context, is synthesized directly from latent compressed features, and adversarial training~\citep{goodfellow2014generative} is used to make up realistic details. Indeed, information propagates in \method{} to new frames as previously synthesized ones become part of the context. Like other video synthesis architectures based on autoregressive models~\citep{weissenborn2020scaling, rakhimov2020latent}, \method{} can be used in many tasks besides future video prediction. Any data which can be expressed in the form of a fixed-size sequence of elements from a finite set (\textit{aka}, tokens) can be processed by a transformer. This applies to video frames (here, via compression and quantization) and to other types of data. Hence, one can easily fuse modalities without having to build complex or task-specific architectures. This idea has been used to control image synthesis~\citep{esser2020taming, ramesh2021zero}, and we extend it to a variety of video synthesis tasks by guiding the prediction with different annotations as illustrated in Figure~\ref{fig:gpt}. Code, pretrained models, and video samples synthesized by our approach are available at the url \url{https://16lemoing.github.io/ccvs}. Our main contributions are as follows:
\begin{enumerate}[noitemsep,topsep=0pt,left=7pt]
    \item an optical flow mechanism within an autoencoder to better reconstruct frames from context,
    \item the use of ancillary information to control latent temporal dynamics when synthesizing videos,
    \item a performance on par with or better than the state of the art, while being more memory-efficient.
\end{enumerate}
\section{Related Work}
\paragraph{Video synthesis.} In its simplest form, videos are produced without prior information about their content. GAN-based approaches map Gaussian noise into a visually plausible succession of frames. For example, VGAN~\citep{vondrick2016generating} and ProgressiveVGAN~\citep{acharya2018towards}, adapt the traditional GAN framework~\citep{goodfellow2014generative} from image to video synthesis by simply using 3D instead of 2D convolutions. These approaches, including recent attempts such as G$^3$AN~\citep{wang2020disentangling}, are computationally expensive, and, by nature, restricted to synthesizing a fixed number of frames due to the constraints of their architecture. Other approaches predict latent motion vectors with a CNN~\citep{saito2017temporal}, or a recurrent neural network (RNN)~\citep{tulyakov2018mocogan, clark2019adversarial, saito2020train}, and generate frames with individual 2D operations. To avoid the shortcomings of information loss in the sequential processing of RNNs, we use an attention-based autoregressive model instead. Since we forecast temporal dynamics in a compressed space, a large temporal window can be used when predicting new frames, without having to resort to expensive 3D computations. Previous works have attempted to scale video synthesis to higher resolution by using progressive training~\citep{acharya2018towards}, sub-sampling~\citep{saito2017temporal, saito2020train}, reducing the dimension of the discriminator~\citep{kahembwe2020lower}, or redefining the task as finding a trajectory in the latent space of a pre-trained image generator~\citep{tian2021good}. Compression, together with efficient context-aware reconstruction allows us to synthesize videos at high resolution.

\paragraph{Controllable video synthesis.} Some of the aforementioned works~\citep{saito2017temporal, clark2019adversarial, saito2020train, wang2020disentangling} handle synthesis conditioned on a class label. Another popular control is to use a few priming frames to set off the generation process. This task, known as \textit{future video prediction}, has received a lot of attention recently. Methods based on variational autoencoders (VAE)~\citep{babaeizadeh2018stochastic, denton2018stochastic} have been proposed to account for the stochastic nature of future forecasting, \ie, the plurality of possible continuations, disregarded in deterministic predictive models~\citep{finn2016unsupervised, mathieu2016deep, vondrick2016anticipating}. Yet, their blurry predictions have motivated the incorporation of adversarial training~\citep{lee2018stochastic}, hierarchical architectures~\citep{castrejon2019improved, wu2021greedy}, fully latent temporal models~\citep{franceschi2020stochastic}, or normalizing flows~\citep{kumar2020video}. Another line of work infers spatial transformation parameters (\eg, optical flow), and predicts the future by warping past frames (\ie, grid sampling and interpolation as in~\citep{jaderberg2015spatial}) in RGB space~\citep{finn2016unsupervised, vondrick2017generating, hao2018controllable, gao2019disentangling, wu2020future} or in feature space~\citep{luc2020transformation}, typically using a refinement step to handle occlusions. Lately, autoregressive methods~\citep{weissenborn2020scaling, rakhimov2020latent} leveraging a self-attention mechanism~\citep{vaswani2017attention} have also been applied to this task. Our method benefits from the context-efficiency of approaches based on spatial transformation modules and the modeling power of autoregressive networks. In the meantime, other forms of control with interesting applications have emerged. Point-to-point generation~\citep{wang2019point}, a variant of future video prediction, specifies both start and end frames. State or action-conditioned synthesis~\citep{hao2018controllable, menapace2021playable} guides the frame-by-frame evolution with high-level commands. Additional works consider video synthesis based on one~\citep{pan2019video} or multiple layouts~\citep{wang2018video, mallya2020world}, another video~\citep{chan2019everybody}, or sound~\citep{vougioukas2018end, jamaludin2019you,  cherian2020sound}. To account for the variety of potential controls, we leverage the flexibility of transformers, and propose an unifying approach to tackle all of these tasks.
\section{Context-aware controllable video synthesis}

\subsection{Overview of the proposed approach}

\begin{figure}
    \centering
    \includegraphics[width=0.60\linewidth, valign=c]{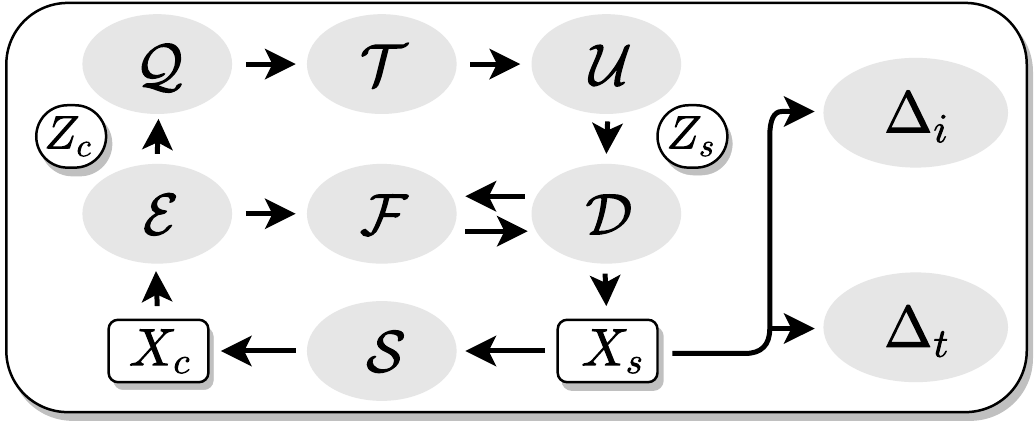}
    \includegraphics[width=0.20\linewidth, valign=c]{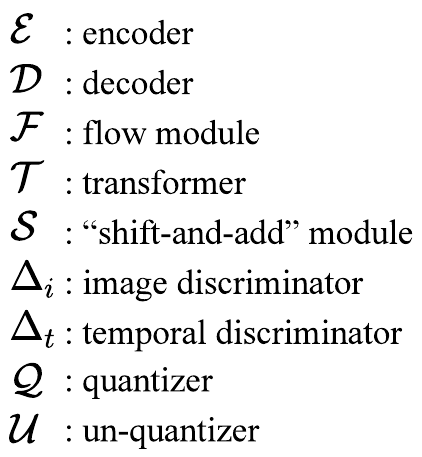}
	\caption{Proposed architecture. Here $X_c$ and $X_s$ respectively stand for the (input) context and (output) synthesized video. Learnable encoder and decoder modules $\encoder$ and $\decoder$ are linked by a learnable flow estimation module $\flowestimator$ ensuring spatio-temporal consistency between context and synthesized frames. The architecture is doubly autoregressive, with a transformer $\transformer$ responsible for predicting the features $Z_s$ associated with future frames $X_s$ from the features $Z_c$ associated with context frames $X_c$, and a simple, parameterless, ``shift-and-add'' module $\shiftandadd$ updating $X_c$ as each new frame is generated. The architecture is trained in two steps: The parameters of $\encoder$, $\decoder$ and $\flowestimator$ are first estimated (without any future forecasting from the transformer) in an adversarial manner using two discriminators $\imagedis$ and $\videodis$ to ensure that the frames synthesized are both realistic ($\imagedis$) and temporally coherent ($\videodis$). The two discriminators are then discarded, and the parameters of the transformer $\transformer$ are estimated with $\encoder$, $\flowestimator$ and $\decoder$ frozen. At inference time, the transformer is used only once, latter frames being estimated in an autoregressive manner by the quadruple $\decoder$ $\shiftandadd$, $\encoder$, $\flowestimator$. See text for more details.}
	\label{fig:skip-ae}
\end{figure}

\label{sec:overview}
\revision{We consider video data, such that an individual frame $x$ is natively an element of $\image=\RR^{H \times W \times 3}$, which can be encoded as some feature $z$ in $\feature=\RR^{h \times w \times F}$ with reduced spatial resolution and increased number of channels. We assume that we are given $X_c$ in $\image^m$ corresponding to $m$ successive frames, and our goal is to {\em synthesize} a plausible representation of the following $n$ frames $X_s$ which lies in $\image^n$. Given $X_c$, we can compute the corresponding features $Z_c$ in $\feature^m$ using some encoder $\encoder:\image\rightarrow\feature$ on each frame individually, then use an autoregressive model (a transformer coupled with a quantization step in our case) to predict features $Z_s$ in $\feature^n$ formed by the $n$ features corresponding to time ticks $m+1$ to $m+n$. (Note that the values of $m$ and $n$ can be arbitrary, using [by now] traditional masking and sliding window techniques.) These features can finally be converted one by one into the corresponding frames $X_s$ using some decoder $\decoder:\feature\rightarrow \image$.}

\revision{The overall approach is illustrated by Figure~\ref{fig:skip-ae}. The use of a quantizer over a learned codebook in our implementation complicates the architecture a bit, but has several advantages, including the reuse of familiar NLP technology~\citep{vaswani2017attention} and, perhaps more importantly, affording simple mechanisms for handling different types of inputs (from images to sound for example~\citep{chen2020generative, huang2018music}) and the intrinsically uncertain and multimodal nature of future prediction (by sampling different codebook elements according to their likelihood~\citep{holtzman2020curious}).
Concretely, this choice simply amounts to inserting in our architecture, right after the encoder $\encoder$, a nearest-neighbor quantizer $\quantizer:\RR^F\rightarrow\llbracket 1,q\rrbracket$ parameterized by an $F \times q$ codebook which, given the embedding $z$ of a frame, returns a $h \times w$ matrix of corresponding {\em tokens}, that is, the indices of the closest entry in the codebook for feature vector $z_{i,j}$ for all spatial locations $(i, j)$ in $\llbracket 1,h\rrbracket\times\llbracket 1,w\rrbracket$. We abuse notation, and identify $\quantizer$ with its parameterization by this codebook, so we can optimize over $\quantizer$ just as we optimize over $\encoder$ instead of naming explicitly its parameters in the rest of this presentation. An ``un-quantizer'' $\unquantizer: \llbracket 1,q\rrbracket\rightarrow \RR^F$, also parameterized (implicitly) by the codebook and associating with each token the corresponding entry of the codebook, is also inserted right before the decoder $\decoder$.
In this setting, $\transformer:\llbracket 1,q\rrbracket^{m\times h\times w} \rightarrow \llbracket 1,q\rrbracket^{n\times h \times w}$ takes as input a sequence of tokens, and outputs the tokens for subsequent frames, each one of them chosen among $q$ possibilities as either the one with the highest score, or drawn randomly from the top-$k$ scores to account for the multimodal nature of future forecasting (here $k\le q$ is some predefined constant, see~\citep{holtzman2020curious} for related approaches).}

\revision{The elements of the architecture described so far are by now (individually) rather classical, with learnable, parametric functions $\encoder$, $\decoder$, $\quantizer$ and $\transformer$. Besides putting them all together, and as detailed in the rest of this section, we add several original elements: {\bf (a)} The encoding/decoding scheme is improved by the use of two discriminators, $\imagedis$ and $\videodis$ respectively, trained in an adversarial manner to ensure that the predicted frames are realistic and temporally consistent. {\bf (b)} The context frames $X_c$ are themselves updated each time a new frame is predicted in an autoregressive manner (iteratively fill the sequence up to some predefined capacity, then shift to the left, forgetting the first frame and adding the latest synthetic one on the right). {\bf (c)} The encoder and decoder are linked through a learnable {\em flow module} $\flowestimator$, allowing the context frames to guide the prediction of the synthetic ones. {\bf (d)} Additional {\em control variables}, ranging from object trajectories to audio tracks, can be used in the form of sequence- or frame-level annotations to drive the synthesis by adding the corresponding tokens to the ones passed on to the autoregressive model~$\transformer$.}

\revision{We detail in this section the concrete components of the approach sketched above, including the autoencoder and quantizer architectures and their training procedure (Section~\ref{sec:ae}), and the implementation of the autoregressive model by a transformer~\citep{vaswani2017attention}, illustrated in Figure~\ref{fig:gpt}, which we adapt to account for outside control signals (Section~\ref{sec:transformer}), once again with the corresponding procedure. Further architectural choices are also detailed in Appendix~\ref{sec:architecture}.}

\subsection{First stage: training the context-aware autoencoder and the quantizer}
\label{sec:ae}

\paragraph{Architecture.} \revision{$\encoder$ and $\decoder$ respectively decreases and increases the spatial resolution by using convolutional residual blocks~\citep{he2016deep}, with $(r_k)_{k\in \llbracket 1;K \rrbracket}$ the $K$ corresponding resolution levels ($r_k=h_k \times w_k$). It is common practice to augment, as in U-Net~\citep{ronneberger2015unet}, the autoencoder with long skip connections between $\encoder$ and $\decoder$ to share information across the two models at these intermediate levels and escape the lossy compression bottleneck. Although we cannot apply this directly to our setting since information only flows through $\decoder$ for predicted timesteps, such skip connections can be established from the encoding stage of a context frame $x_c$ to the decoding stage of a new frame $x_s$ (resulting from features $z_s$). Similar mechanisms~\citep{ebert2017self, denton2018stochastic} have been proposed in the past for video synthesis but they are only copying static background features from a single context frame.
We follow works on semantic segmentation~\citep{gadde2017semantic} and face frontalization~\citep{wei2020learning} and use a flow module $\flowestimator$ to warp features and produce temporally consistent outputs despite motion. We extend this to multi-frame contexts, with significant performance gains and no additional parameter to be learned.}

\revision{Concretely, let $e_c^k$ be features being encoded from $x_c$, and $d_s^k$ features being decoded from $z_s$ at a given intermediate resolution~$r_k$. We first compute all intermediate context features $e_c^k$ by applying $\encoder$ to $x_c$. We then progressively decode features $d_s^k$ for the new frame from low ($k=1$) to high resolution ($k=K$) by iterating over the following steps: {\bf (a)} apply one decoding sub-module to get $d_s^k$ from $d_s^{k-1}$, {\bf (b)} use $\flowestimator$ to refine the optical flow $f_c^k$ (in $\RR^{2 \times r_k}$) which estimates the displacement field from $e_c^k$ to $d_s^k$ (as a proxy to the one from $x_c$ to $x_s$), and a fusion mask $m_c^k$ (in $\RR^{1 \times r_k}$) which indicates the expected similarity between aligned features $e_c'^k=\warp(e_c^k, f_c^k)$ and $d_s^k$ (also as a proxy for the one in image domain) with $\warp$ corresponding to a standard warping operation, {\bf (c)} use $e_c'^k$ and $m_c^k$ to update $d_s^k$ with context information (see update rule~(\ref{eq:update}) below), {\bf (d)} move to resolution level~$r_{k+1}$ by going back to {\bf (a)}. We note that $\flowestimator$ estimates $f_c^k$ and $m_c^k$ in a coarse-to-fine fashion by refining $f_c^{k-1}$ and $m_c^{k-1}$ (see Appendix~\ref{sec:architecture} for further details).} Temporal skip connections at a given resolution level $r_k$ are defined as the following in-place modification of $d_s^k$:
\begin{equation}
    d_s^k=\sigma(m_c^k) \otimes d_s^k+(\mathbf{1}-\sigma(m_c^k)) \otimes e_c'^k,
\label{eq:update}
\end{equation}
where $\sigma$ is the Sigmoid function, and $\otimes$ the element-wise product. \revision{We note that update rule~(\ref{eq:update}) is quite standard for warping and fusing two streams of spatial information~\citep{wang2018video, hao2018controllable, gadde2017semantic}.} For concrete implementation of $\flowestimator$, we build upon LiteFlowNet~\citep{hui2018liteflownet}, an optical flow estimation model which also combines pyramidal extraction and progressive warping of features. For simple integration into our framework, we use features from $\encoder$ and $\decoder$ both in the mask and flow estimation, and in the update~(\ref{eq:update}). This process readily generalizes to multi-frame contextual information (see Appendix~\ref{sec:multi} for details). \revision{Similar to spatial transformers~\citep{jaderberg2015spatial}, the warping operation $\warp$ is differentiable. As a result, gradients from the training losses can backpropagate from $\decoder$ to $\encoder$ through $\flowestimator$. This allows end-to-end training of the autoencoder even with information from different timesteps in $\encoder$ and $\decoder$.}

\paragraph{Training procedure.} The global objective is the linear combination of four auxiliary ones:
\begin{equation}
\label{eq:ae}
    \loss = \lambda_q\loss_q + \lambda_r\loss_r + \lambda_a\loss_a + \lambda_c\loss_c,
\end{equation}
namely a quantization loss ($\loss_q$), a reconstruction loss ($\loss_r$), an adversarial loss ($\loss_a$), and a contextual loss ($\loss_c$), detailed in the next paragraphs.

The codebook is trained by minimizing the reconstruction error between encoded features $z=\encoder(x)$ and quantized features $z_q=\unquantizer(\quantizer(z))$ (with notations introduced in Section~\ref{sec:overview}):
\begin{equation}
\label{eq:q}
    \loss_q(\encoder, \quantizer)=\|\sg(z)-z_q\|_2^2 + \beta \|\sg(z_q)-z\|_2^2,
\end{equation}
where $\sg(.)$ is the stop gradient operation which constrains its operand to remain constant during backpropagation. The first part moves the codebook entries closer to the encoded features. The second part, known as the \textit{commitment loss}~\citep{vandenoord2017neural}, reverses the roles played by the two variables.

In regions with complex textures and high frequency details, local patterns shifted by a few pixels in $x$ and its reconstruction $\widehat{x}=\decoder(z_q)$ may result in large pixel-to-pixel errors, while being visually satisfactory. We thus define the recovery objective as the $L_1$ loss in both RGB space and between features from a $\VGG$ network~\citep{simonyan2015very} pretrained on ImageNet~\citep{deng2009imagenet}:
\begin{equation}
\label{eq:recovery}
    \loss_r(\encoder, \quantizer, \flowestimator, \decoder)=\|x-\widehat{x}\|_1+\|\VGG(x)-\VGG(\widehat{x})\|_1.
\end{equation}

To tackle cases where information cannot be recovered from context due to occlusion, and the compressed features are insufficient to create plausible reconstructions due to lossy compression, we supplement our architecture with an image discriminator $\imagedis$, made of downsampling residual blocks~\citep{he2016deep}, to encourage realistic outputs. $\imagedis$ tries to distinguish real images from reconstructed ones ($\loss_d$), while $\encoder$ and $\decoder$ fools $\imagedis$ into assuming reconstructed images are as good as real ones ($\loss_a$):
\vspace{-0.35cm}
\begin{align}
\begin{minipage}{.5\linewidth}
\begin{equation}
    \loss_d(\imagedis)=\ln(1+e^{\imagedis(x)})+\ln(1+e^{-\imagedis(\widehat{x})}),
\label{dis-loss}
\end{equation}
\end{minipage}%
\begin{minipage}{.5\linewidth}
\vspace{0.20cm}
\begin{equation}
    \loss_a(\encoder, \quantizer,  \flowestimator, \decoder)=\ln(1+e^{\imagedis(\widehat{x})}).
\label{adv-loss}
\end{equation}
\end{minipage} 
\end{align}
We employ a similar strategy on sequences of consecutive frames to improve the temporal consistency using a 3D temporal discriminator $\videodis$, a direct extension of 2D image discriminator $\imagedis$.

The success of our method relies on accurate motion estimation in $\flowestimator$, a difficult task which benefits from self-supervision~\citep{jonschkowski2020what}. Therefore, we train the autoencoder with augmented views of the input frames as context. Custom augmentations functions $\aug:\image\rightarrow\image$ include: rotation, scaling, translation, elastic deformation, and combinations of these. Augmented views are obtained by warping $x$ by the suitable flow $a_c$: $\aug(x)=\warp(x,a_c)$,
and the inverted flow $f_c$ from $\aug(x)$ to $x$ can be approximated.\footnote{Our implementation of flow inversion approximation is detailed in Appendix~\ref{sec:inversion}.} We resort to flow inversion because directly reconstructing distorted views may encourage similar defects during inference. Moreover, we balance between self-recovery and context-recovery objectives by additionally applying a blurring function $\blur:\image\rightarrow\image$ and an occlusion mask $o_c$ to the augmented frames $x_c=o_c \otimes \blur(\aug(x))$. This augmentation strategy is illustrated in Appendix~\ref{sec:augmentation}. We define the contextual loss as:
\begin{equation}
    \loss_c(\encoder, \quantizer, \flowestimator, \decoder)=\lVert f_c - \widehat{f}_c \rVert_2^2+\lVert o_c' - \sigma(\widehat{m}_c) \rVert_2^2,
\label{flow-loss}
\end{equation}
where $o_c'=\warp(o_c,f_c)$, and $\widehat{f}_c$ and $\widehat{m}_c$ are the flow and mask estimated by $\flowestimator$. In practice, this loss is applied at intermediate resolution levels $r_k$ for improved training.

\subsection{Second stage: predicting temporal dynamics with transformers}
\label{sec:transformer}

\begin{figure}
    \centering
    \includegraphics[width=0.76\linewidth, valign=c]{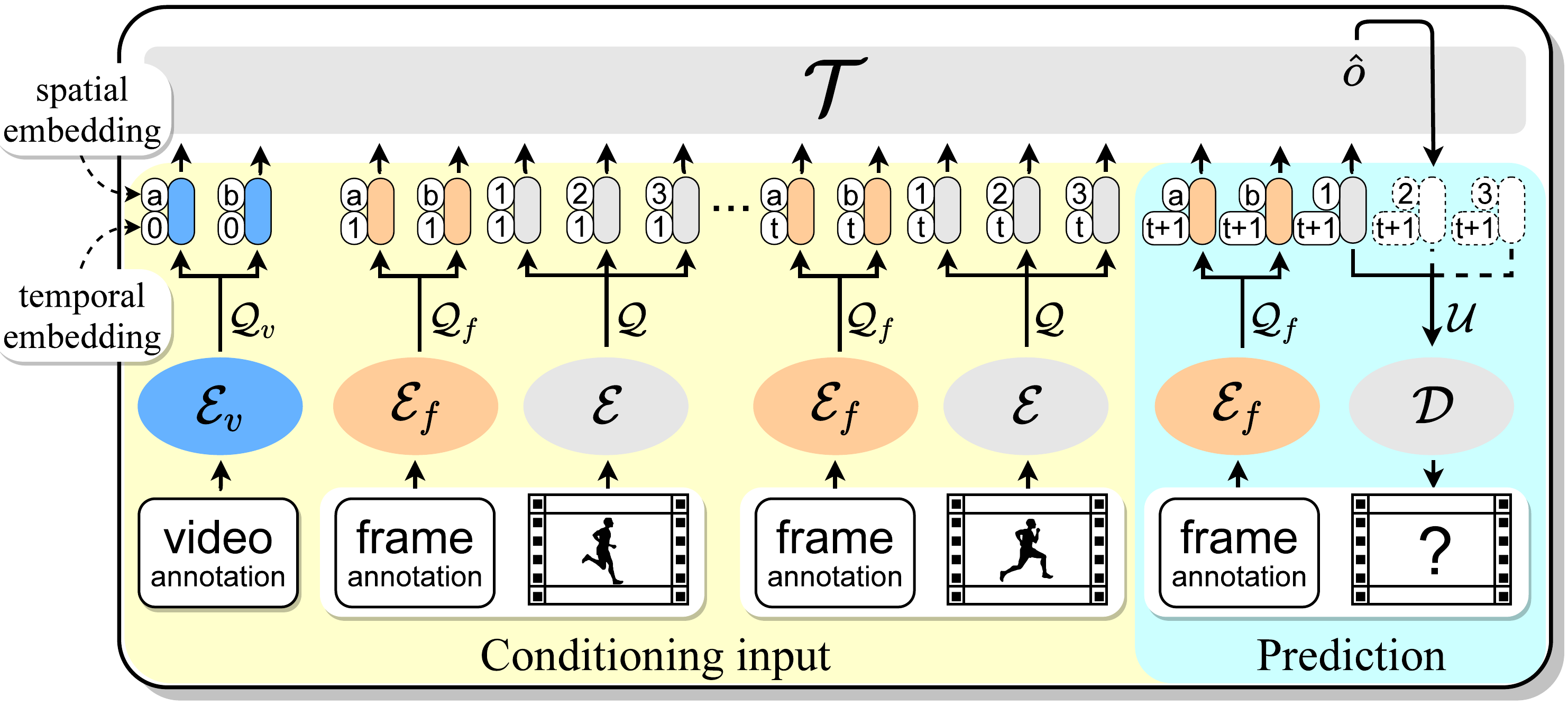}
    \includegraphics[width=0.23\linewidth, valign=c]{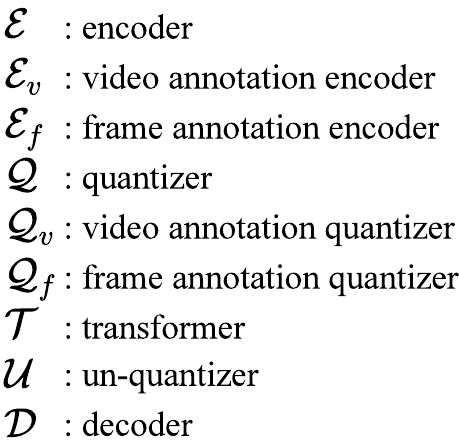}
	\caption{Illustration of the transformer to predict latent temporal dynamics with control-related data. For clarity, we omit the flow module $\flowestimator$ between $\encoder$ and $\decoder$. Input video-level annotation, frame-level annotations, and conditioning frames are encoded and quantized to form the initial token sequence, and mapped to corresponding embeddings. Subsequent frame tokens are obtained autoregressively with $\transformer$, while the ones corresponding to video- and frame-level annotations guide the prediction.}
    \label{fig:gpt}
\end{figure}

\paragraph{Architecture.} We follow~\citep{esser2020taming} and adopt an architecture similar to Image-GPT~\citep{chen2020generative} for $\transformer$. Instead of modeling a single annotated frame as in~\citep{ramesh2021zero}, we design our model to account for sequences of $N$ such frames to allow prediction of temporal dynamics controlled by ancillary information in the form of video- and frame-level annotations. We have shown, in Section~\ref{sec:ae}, how to represent a frame as a sequence of $h \times w$ tokens (indices in $\llbracket 1,q\rrbracket$) through encoding and quantization. Similar strategies can be applied to cater to other types of data, with or without compression depending on their complexity, and, thereby, turn ancillary information into tokens as well. The capacity of the transformer (the maximum sequence length it can process) is $L=l_v+N*(l_f + h*w)$, where $l_v$ and $l_f$ are the size of video- and frame-level annotations respectively. The final layer of the model predicts, for every $i$ in $\llbracket 1,L-1\rrbracket$, a vector $\widehat{o}_{i}$ (of size $q$, the number of possible tokens) which scores the likelihood of the $i+1$\textsuperscript{th} token given all preceding ones.

\paragraph{Training procedure.} To learn the parameters of $\transformer$, we load a complete sequence ($N$ frames corresponding to $L$ tokens), and try to predict the $L-1$ last tokens based on the $L-1$ first ones. This is done by maximizing the log-likelihood of the data using the cross-entropy loss:
\begin{equation}
    \loss(\transformer)=-\sum_{i=2}^{L}\log\left(\frac{\exp(\langle\,\widehat{o}_{i-1},e_{\stau(i)}\rangle)}{\sum_{j}\exp(\langle\,\widehat{o}_{i-1},e_j\rangle)}\right),
\end{equation}
where $\ltau(i)$ is the $i$\ts{th} ground-truth token, and $e_j$ a vector of 0's with a $1$ in its $j$\ts{th} coordinate. 

\paragraph{Inference.} During inference we use $\transformer$ to complete autoregressively an input sequence of tokens. To avoid known pitfalls (\eg, repetitive synthesis) and allow diverse outcomes, we use top-$k$ sampling whereby the next token is randomly chosen among the $k$ most likely ones, weighted by their scores $\widehat{o}$. Although $\transformer$ processes sequences of $N$ annotated frames, we predict ones of arbitrary length by using a temporal sliding window. Tokens corresponding to video- and frame-level annotations are given as input to $\transformer$ (not predicted, even for future frames) to guide the synthesis. We show in the experiments how this control can translates into a variety of interesting tasks.
\section{Experiments}
\label{sec:experiments}
We assess the merits of \method{} in the light of extensive experiments on various video synthesis tasks.

\paragraph{Datasets.} BAIR Robot Pushing~\citep{ebert2017self} consists of 43k training and 256 test videos of a robotic arm interacting with objects from a fixed viewpoint. A high-resolution version has recently been released. We manually annotate, in $500$ frames, the $(x,y)$ location of the arm in image space to train a position estimator, which we use for state-conditioned synthesis. To account for real world scenarios, we evaluate on Kinetics-600~\citep{carreira2018a}, a large and diverse action-recognition dataset with approximately 500K videos. We also test our method on AudioSet-Drums~\citep{gemmeke2017audio} for sound-conditioned synthesis on music performance, containing 6k and 1k video clips in train and test splits respectively. Other datasets and tasks are covered in Appendix~\ref{sec:additional-results}.

\paragraph{Metrics.} We use the Fréchet video distance (FVD)~\citep{unterthiner2018towards} which measures the distribution gap between real and synthetic videos in the feature space of an Inception3D network~\citep{carreira2017quo} pretrained on Kientics-400~\citep{kay2017the}. \revision{It estimates the visual quality and temporal consistency of samples as well as the diversity in unconditioned scenarios. In conditioned ones, we use another metric for diversity (DIV) which is the mean pixel-wise distance among synthetic trajectories conditioned on the same input.} For near-deterministic motions (\eg, in reconstructions, or constrained tasks), there is a one-to-one mapping between real video frames and synthetic ones, and we include pairwise image quality assessments: the structural similarity index measure (SSIM)~\citep{wang2004image} which evaluates a per-frame conformity (combination of luminance, contrast and structure), and the peak signal-to-noise ratio (PSNR) which is directly related to the root mean squared error. For each metric, we compute the mean and standard deviation (std) over $5$ evaluation runs (2 for Kinetics due to its voluminous 50k video test set). For clarity, the std value is shown only when it is greater than the reported precision.

\paragraph{Training details.}
All our models are trained on 4 Nvidia V100 GPUs (32GB VRAM each), using ADAM~\citep{kingma2014adam} optimizer, for multiple 20 hour runs. We adapt the batch size to fill the memory available. We use a learning rate of $0.02$ to train the autoencoder, and exponential moving average~\citep{yaz2019unusual} to obtain its final parameters. We use weighting factors $(1, 10, 1, 1)$ and $0.25$ for $(\lambda_q, \lambda_r, \lambda_a, \lambda_c)$ and $\beta$ in Equations (\ref{eq:ae}) and (\ref{eq:q}) respectively. We use a learning rate of $10^{-5}$ to train the transformer.

\subsection{Ablation study}
\label{sec:ablation}
\begin{table}
 \setlength\heavyrulewidth{0.25ex}
 \aboverulesep=0ex
 \belowrulesep=0.3ex
 \centering
 \caption{Ablation study of the autoencoder on BAIR ($256\stimes256$). We evaluate self- and context-recovery modules in different scenarios: synthesizing 16-frame videos from known compressed features (``Reconstruction''), by inferring compressed features with $\transformer$ given the real trajectory of the robotic arm (``State-conditioned''), or without the trajectory (``Pred.'' and ``Unc.''). The first real frame is used as initial context in all cases, except for ``Unc.'' where it is synthesized by StyleGAN2~\citep{karras2020analyzing}.}
 \label{tab:ae-ablation}
 \footnotesize
 \begin{tabular}{@{}C{0.80cm}@{}C{0.80cm}@{}C{0.80cm}@{}C{0.80cm}@{}C{0.80cm}@{}C{0.80cm}@{}C{0.80cm}@{}C{0.10cm}@{}C{1.30cm}@{}C{1.30cm}@{}C{0.10cm}@{}C{1.30cm}@{}C{1.30cm}@{}C{0.10cm}@{}C{1.30cm}@{}C{0.10cm}@{}C{1.30cm}@{}}
 \toprule
 \multicolumn{4}{c}{\footnotesize Self-recovery} & \multicolumn{3}{c}{\footnotesize Ctxt.-recovery} & \multicolumn{3}{c}{ \footnotesize Reconstruction} & \multicolumn{3}{c}{ \footnotesize State-conditioned} & & {\footnotesize Pred.} & & {\footnotesize Unc.} \\
 \cmidrule(r){1-4} \cmidrule(lr){5-7} \cmidrule(lr){8-10} \cmidrule(lr){11-13} \cmidrule(lr){14-15} \cmidrule(l){16-17}
 {\scriptsize RGB} & {\scriptsize VGG} & {\scriptsize $\imagedis$} & {\scriptsize $\videodis$} & {\scriptsize $\flowestimator$} & {\scriptsize Sup.} & {\scriptsize Ctxt.} & & {\scriptsize FVD \hspace{-1.0ex} $\downarrow$} & {\scriptsize PSNR \hspace{-1.0ex} $\uparrow$} & & {\scriptsize FVD \hspace{-1.0ex} $\downarrow$} & {\scriptsize PSNR \hspace{-1.0ex} $\uparrow$} & & {\scriptsize FVD \hspace{-1.0ex} $\downarrow$} & & {\scriptsize FVD \hspace{-1.0ex} $\downarrow$} \\
 \midrule
 \checkmark &   &   &   &   &   & 0 &  & $1200_{\pm6}$ & $20.0$ & & $1238_{\pm24}$ & $17.8$ & & $1265_{\pm22}$ & & $1321_{\pm10}$\\
 \rowcolor[HTML]{EFEFEF}[0pt][0pt]
 \checkmark & \checkmark &   &  &   &   & 0 &  & $700_{\pm11}$ & $17.9$ & & $714_{\pm5}$  & $17.4$ & & $704_{\pm7}$ & & $765_{\pm12}$\\
 \checkmark & \checkmark & \checkmark   &   &   &   & 0 & & $323_{\pm3}$ & $17.8$ & & $355_{\pm5}$ & $16.6$ & & $377_{\pm11}$ & & $566_{\pm22}$\\
 \rowcolor[HTML]{EFEFEF}[0pt][0pt]
 \checkmark & \checkmark & \checkmark & \checkmark   &   &   & 0 & & $389_{\pm12}$ & $18.2$ & & $401_{\pm4}$ & $16.9$ & & $407_{\pm10}$ & & $627_{\pm9}$\\
 \checkmark & \checkmark & \checkmark & \checkmark & \checkmark &   & 1 & & $98_{\pm2}$ & $22.7$ & & $106_{\pm2}$ & $22.2$ & & $142_{\pm6}$ & & $350_{\pm11}$\\
 \rowcolor[HTML]{EFEFEF}[0pt][0pt]
 \checkmark & \checkmark & \checkmark & \checkmark & \checkmark & \checkmark   & 1 &  & $87_{\pm3}$ & $24.4$ & & $97_{\pm4}$ & $22.1$ & & $128_{\pm4}$ & & $301_{\pm10}$\\
 \checkmark & \checkmark & \checkmark & \checkmark & \checkmark & \checkmark   & 8 & & $62_{\pm1}$ & $25.4$ & & $76_{\pm3}$ & $22.7$ & & $109_{\pm6}$ & & $299_{\pm4}$\\
 \rowcolor[HTML]{EFEFEF}[0pt][0pt]
 \checkmark & \checkmark & \checkmark & \checkmark & \checkmark & \checkmark   & 15 & & $60_{\pm1}$ & $25.6$ & & $75_{\pm2}$ & $\mathbf{22.8}$ & & $110_{\pm3}$ & & $297_{\pm7}$\\
 \multicolumn{7}{@{}c@{}}{\textit{Training longer ~~~~ (num. epochs $\stimes \, 3$)}} & &  $\mathbf{45}_{\pm1}$ & $\mathbf{26.8}$ & & $\mathbf{67}_{\pm1}$ & $22.3$ & & $\mathbf{100}_{\pm2}$ & & $\mathbf{293}_{\pm7}$\\
 \bottomrule
 \multicolumn{17}{c}{\scriptsize ``Sup.'': self-supervision of $\flowestimator$; \hspace{2ex} ``Ctxt.'': number of context frames taken into account (in $\flowestimator$) when decoding current frame (in $\decoder$).} \\
 \end{tabular}
\end{table}
\begin{figure}\CenterFloatBoxes
\begin{floatrow}

\ttabbox{
 \setlength\heavyrulewidth{0.25ex}
 \aboverulesep=0ex
 \belowrulesep=0.3ex
 \centering
 \footnotesize
 \begin{tabular}{@{}C{0.65cm}@{}C{0.65cm}@{}C{0.65cm}@{}C{0.65cm}@{}C{0.65cm}@{}C{1.15cm}@{}C{1.15cm}@{}C{1.15cm}@{}C{1.15cm}@{}}
 \toprule
 \multicolumn{3}{@{}c@{}}{\footnotesize Architecture} & \multicolumn{2}{@{}c@{}}{\footnotesize Top-k} & \multicolumn{2}{@{}c@{}}{\footnotesize State-conditioned} & {\footnotesize Pred.} & {\footnotesize Unc.} \\
 \cmidrule(r){1-3} \cmidrule(lr){4-5} \cmidrule(lr){6-7} \cmidrule(lr){8-8} \cmidrule(l){9-9}
 {\scriptsize Layer} & {\scriptsize Head} & {\scriptsize Dec.} & {\scriptsize Frame} & {\scriptsize State} & {\scriptsize FVD \hspace{-1.0ex} $\downarrow$} & {\scriptsize PSNR \hspace{-1.0ex} $\uparrow$} & {\scriptsize FVD \hspace{-1.0ex} $\downarrow$} & {\scriptsize FVD \hspace{-1.0ex} $\downarrow$} \\
 \midrule
 6 & 4 &  & 1 & 1 & $73_{\pm2}$ & $23.0$ & $281_{\pm7}$ & $474_{\pm17}$\\
 \rowcolor[HTML]{EFEFEF}[0pt][0pt]
 12 & 8 &  & 1 & 1 & $73_{\pm3}$ & $23.1$ & $262_{\pm7}$ & $435_{\pm8}$\\
 24 & 16 &  & 1 & 1 & $70_{\pm3}$ & $\mathbf{23.3}$ & $331_{\pm9}$ & $521_{\pm19}$\\
 \rowcolor[HTML]{EFEFEF}[0pt][0pt]
 24 & 16 & \checkmark & 1 & 1 & ${69}_{\pm2}$ & $23.2$ & $321_{\pm9}$ & $479_{\pm34}$\\
 24 & 16 & \checkmark & 10 & 1 & $\mathbf{65}_{\pm2}$ & $22.4$ & $127_{\pm7}$ & $308_{\pm19}$\\
 \rowcolor[HTML]{EFEFEF}[0pt][0pt]
 24 & 16 & \checkmark & 100 & 1 & $67_{\pm1}$ & $22.3$ & $121_{\pm2}$ & $314_{\pm12}$\\
 24 & 16 & \checkmark & 100 & 10 & $67_{\pm1}$ & $22.3$ & $\mathbf{100}_{\pm2}$ & $\mathbf{293}_{\pm7}$\\
 \bottomrule
 \multicolumn{9}{c}{\scriptsize ``Dec.'': spatio-temporal decomposition of positional embeddings.} \\
 \end{tabular}
}{\caption{\makebox[1.0\textwidth][l]{Ablation~ study~ of~ the~ transformer~ on~ BAIR.} We~adopt
notations and evaluation setups from Table~\ref{tab:ae-ablation}.}\label{tab:transformer-ablation}}

\ffigbox[\FBwidth]{%
  \includegraphics[width=0.39\textwidth]{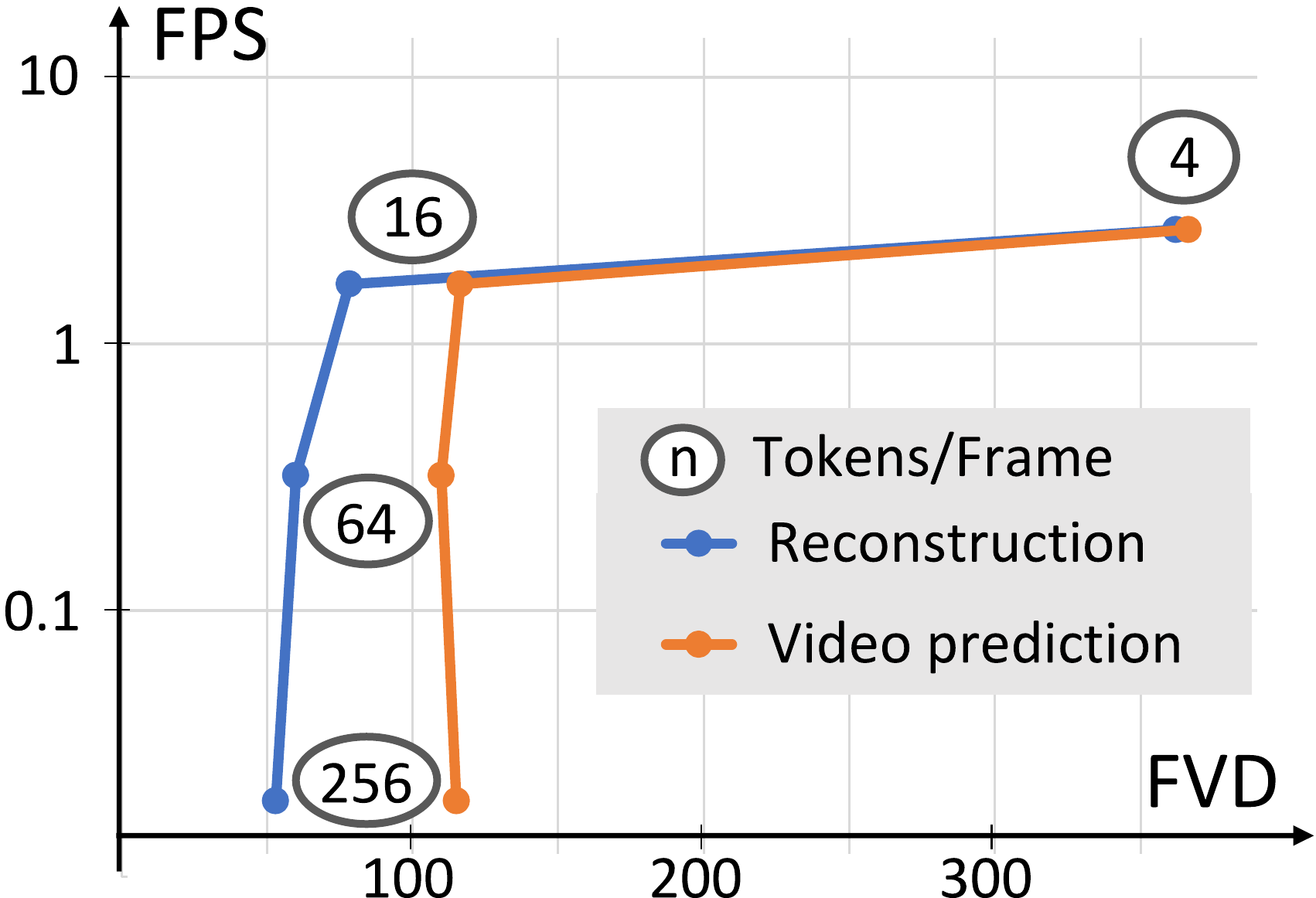}
}{\caption{Quality and speed of synthesis vs. compression on a Nvidia V100 GPU.}\label{fig:compression}}

\end{floatrow}
\end{figure}

We conduct an ablation study of \method{} to show the individual contribution of the key components of the proposed autoencoder (Table~\ref{tab:ae-ablation}), transformer (Table~\ref{tab:transformer-ablation}), and the effect of compression (Figure~\ref{fig:compression}). 

First, we fix the transformer and observe the incremental improvements in synthesis quality when adding self- and context-recovery modules to the autoencoder (Table~\ref{tab:ae-ablation}). In particular, $L_1$ loss in the feature space of a $\VGG$ net~\citep{simonyan2015very} produces sharper videos than using the same loss in the RGB space alone. Predicted frames using image and temporal discriminators ($\imagedis$ and $\videodis$) display greater realism and temporal consistency. The flow module $\flowestimator$ significantly improves the performance on all metrics by allowing context frames to guide the reconstruction of synthetic ones. The self-supervision of $\flowestimator$, the use of larger context windows, and longer training times, further improve the quality of the synthesis. $\videodis$ seems to deteriorate the FVD at first, but when all modules are combined it improves FVD by almost a factor $2$ as it encourages better temporal consistency in the presence of context.

We fix the autoencoder, and compare different architectures and sampling strategies for the transformer (Table~\ref{tab:transformer-ablation}). Increasing the model capacity by adding layers and increasing expressivity (number of attention heads) along with a spatio-temporal decomposition of positional embeddings (shown in Figure~\ref{fig:gpt}, detailed in supplementary material) yields small improvements. Top-$k$ sampling is beneficial for stochastic tasks (video prediction and unconditional synthesis), whereas always selecting the most probable tokens results in little to no motion. Guiding temporal dynamics with state-conditioned synthesis reduces the advantage of sampling by narrowing down possible outcomes. 

Finally, we explore the effect of compression (in terms of the number of tokens for each frame) on synthesis speed (FPS) and quality (FVD) of reconstructed and predicted videos (Figure~\ref{fig:compression}). We use a compression of $64$ tokens in our default setup since it gives the best FVD while retaining a reasonable speed. The critical drop of FPS for low compression ratios is due to the pairwise consideration of all input tokens in $\transformer$. Note that using a smaller temporal window in $\transformer$ may allow additional speed-ups.

\subsection{Quantitative and qualitative studies}
\label{sec:qua}
\begin{table}
    \begin{minipage}[t]{0.800\textwidth}
    \vspace{0pt}
    \setlength\heavyrulewidth{0.25ex}
	\aboverulesep=0ex
    \belowrulesep=0.3ex
	\centering
	\caption{Future video prediction on BAIR ($64\stimes64$), synthesizing 16-frame videos given a few conditioning frames (``Cond.''). We include some extensions of our method at higher resolution.}
	\footnotesize
	\begin{tabular}{@{}L{4.00cm}@{}C{1.10cm}@{}C{1.10cm}@{}C{2.35cm}@{}C{2.35cm}@{}C{0.01cm}@{}}
    	\toprule
        Method                                           & Cond. & FVD $\downarrow$ & Code Avail. & Memory, compute  \\
        \cmidrule(){1-5}
        MoCoGAN~\citep{tulyakov2018mocogan}              & 4     & $503$            & \checkmark  & 16GB, 23h$^*$    \\
        SVG-FP~\citep{denton2018stochastic}              & 2     & $315$            & \checkmark  & 12GB, 6 to 24h$^*$              \\
        CDNA~\citep{finn2016unsupervised}                & 2     & $297$            & \checkmark  & 10GB, 20h$^*$    \\
        SV2P~\citep{babaeizadeh2018stochastic}           & 2     & $263$            & \checkmark  & 16GB, 24 to 48h$^*$    \\
        SRVP~\citep{franceschi2020stochastic}            & 2     & $181$            & \checkmark  & 36GB, 168h$^*$   \\
        VideoFlow~\citep{kumar2020video}                 & 3     & $131$            &             & 128GB, 336h$^*$  \\
        LVT~\citep{rakhimov2020latent}                   & 1     & $126_{\pm3}$     & \checkmark  & 128GB, 48h      \\
        SAVP~\citep{lee2018stochastic}                   & 2     & $116$            & \checkmark  & 32GB, 144h       \\
        DVD-GAN-FP~\citep{clark2019adversarial}          & 1     & $110$            &             & 2TB, 24h$^*$    \\
        Video Transformer (S)~\citep{weissenborn2020scaling} & 1 & $106_{\pm3}$     &             & 256GB, 33h$^*$ \\
        TriVD-GAN-FP~\citep{luc2020transformation}           & 1 & $103$            &             & 1TB, 280h$^*$   \\
        \textit{Low res.} \method{} (\emph{ours})        & 1     & $99_{\pm2}$      & \checkmark  & 128GB, 40h  \\
        Video Transformer (L)~\citep{weissenborn2020scaling} & 1 & $94_{\pm2}$      &             & 512GB, 336h$^*$ \\
        \cmidrule(){1-5}
        & & & SSIM $\uparrow$ ($t=8$)  & SSIM $\uparrow$ ($t=15$)  \\
        \cmidrule(){1-5}
        \textit{High res.$^{**}$} \method{} (\emph{ours}) & 1 & $80_{\pm3}$ & 0.729 & 0.683 \\
        ~~+ end frame   & 2 & $81_{\pm2}$  & 0.766 & 0.839  \\
        ~~+ state       & 1 & $50_{\pm1}$  & 0.885 & 0.863 \\
        \bottomrule \\
        [-0.37cm] 
        \multicolumn{6}{c}{\scriptsize~~~~~~~~~~~~~~~~~~~~~~~$^*$: value confirmed by authors; \hspace{10ex} $^{**}$: training / inference / SSIM at $256 \stimes 256$, FVD at $64 \stimes 64$.~~~~~~~~~~~~~~~~~~~~~~~} \\
    \end{tabular}
	\label{tab:bair-prediction-64}
	\end{minipage}
	\hspace{0.0ex}
	\begin{minipage}[t]{0.178\textwidth}
	\vspace{0.3ex}
	\centering
	\includegraphics[width=\textwidth]{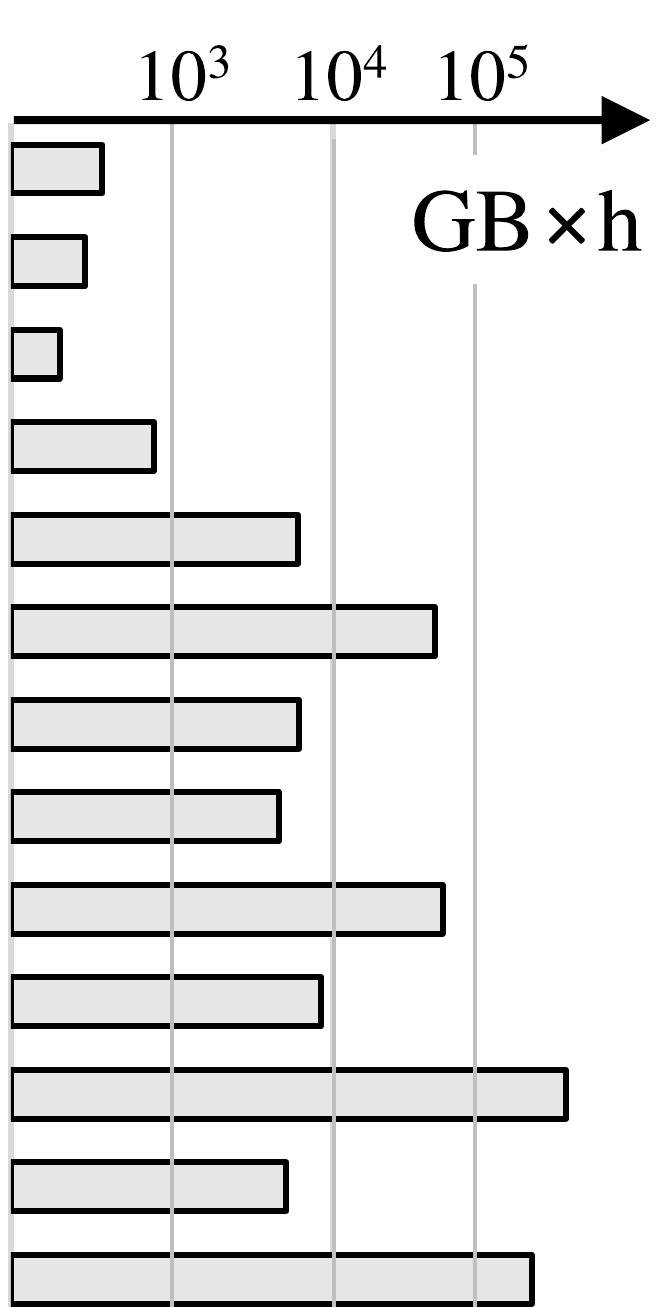}
	\end{minipage}
\end{table}

\begin{figure}
	\setlength\tabcolsep{1.0pt}
	\renewcommand{\arraystretch}{1.0}
	\footnotesize
	\begin{tabular}{cccccccccc}
	    & $t=1$ & $t=2$ & $t=4$ & $t=6$ & $t=8$ & $t=10$ & $t=12$ & $t=14$ & $t=16$ \\
	    [0.05cm]
		
		\raisebox{1.75\normalbaselineskip}[0pt][0pt]{\rotatebox[origin=c]{90}{\parbox{4cm}{\centering Future\\prediction}}} &
		\includegraphics[width=.100\linewidth]{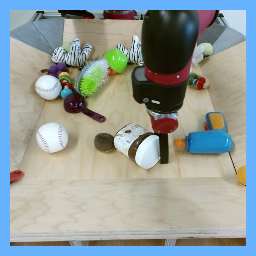} &
		\includegraphics[width=.100\linewidth]{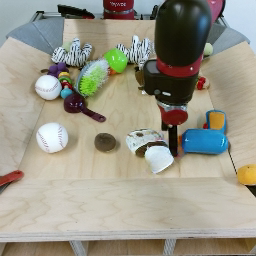} &
		\includegraphics[width=.100\linewidth]{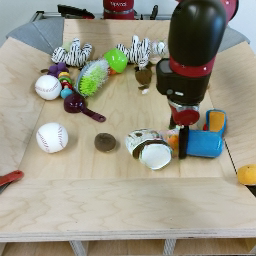} &
		\includegraphics[width=.100\linewidth]{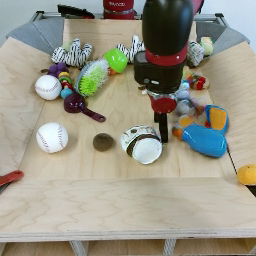} &
		\includegraphics[width=.100\linewidth]{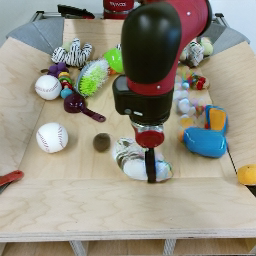} &
		\includegraphics[width=.100\linewidth]{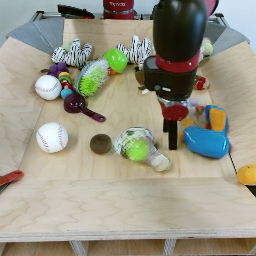} &
		\includegraphics[width=.100\linewidth]{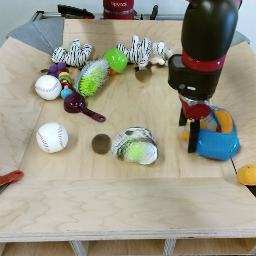} &
		\includegraphics[width=.100\linewidth]{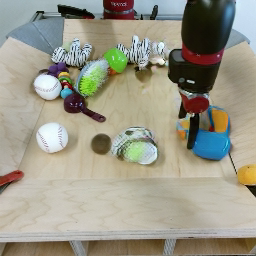} &
		\includegraphics[width=.100\linewidth]{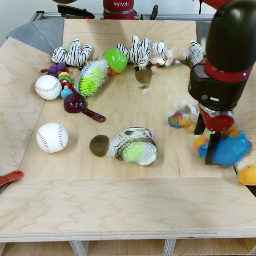} \\
		
		\raisebox{1.75\normalbaselineskip}[0pt][0pt]{\rotatebox[origin=c]{90}{\parbox{4cm}{\centering Point-to-\\point}}} &
		\includegraphics[width=.100\linewidth]{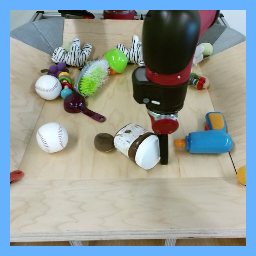} &
		\includegraphics[width=.100\linewidth]{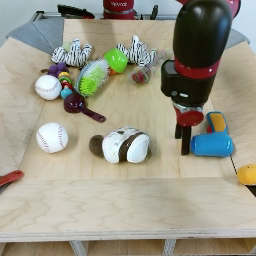} &
		\includegraphics[width=.100\linewidth]{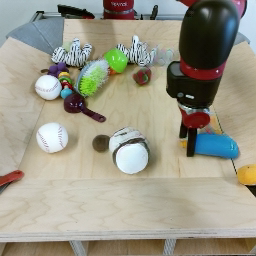} &
		\includegraphics[width=.100\linewidth]{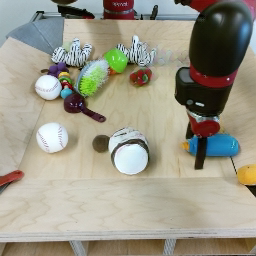} &
		\includegraphics[width=.100\linewidth]{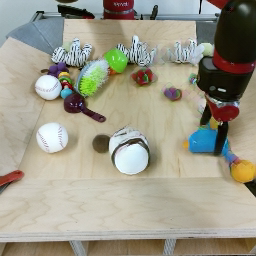} &
		\includegraphics[width=.100\linewidth]{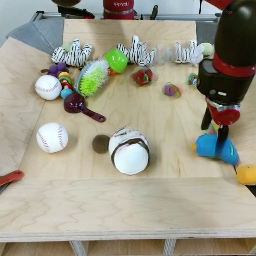} &
		\includegraphics[width=.100\linewidth]{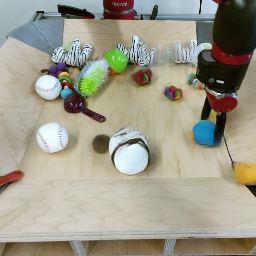} &
		\includegraphics[width=.100\linewidth]{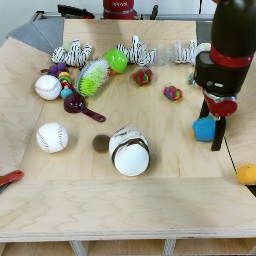} &
		\includegraphics[width=.100\linewidth]{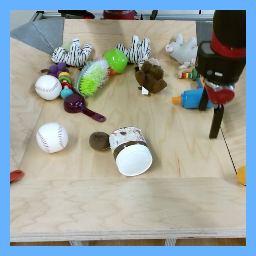} \\
		
		\raisebox{1.75\normalbaselineskip}[0pt][0pt]{\rotatebox[origin=c]{90}{\parbox{4cm}{\centering State-\\conditioned}}} &
		\includegraphics[width=.100\linewidth]{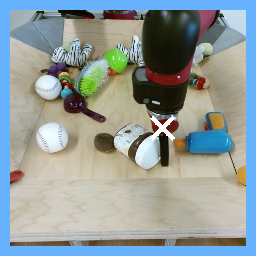} &
		\includegraphics[width=.100\linewidth]{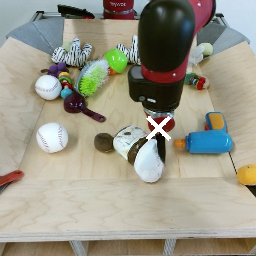} &
		\includegraphics[width=.100\linewidth]{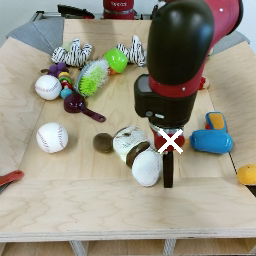} &
		\includegraphics[width=.100\linewidth]{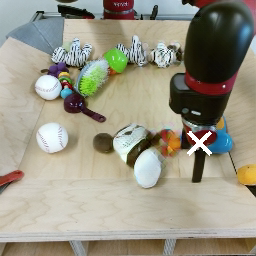} &
		\includegraphics[width=.100\linewidth]{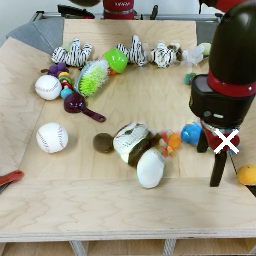} &
		\includegraphics[width=.100\linewidth]{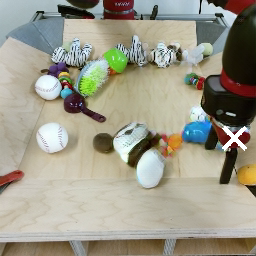} &
		\includegraphics[width=.100\linewidth]{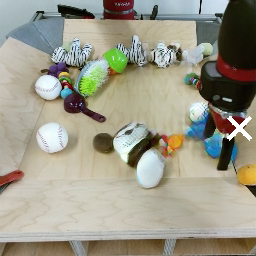} &
		\includegraphics[width=.100\linewidth]{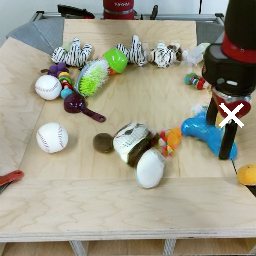} &
		\includegraphics[width=.100\linewidth]{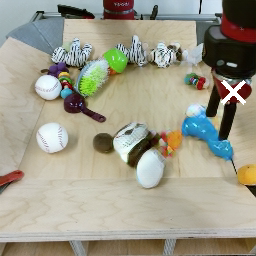} \\
		
		\raisebox{1.75\normalbaselineskip}[0pt][0pt]{\rotatebox[origin=c]{90}{\parbox{4cm}{\centering Real\\video}}} &
		\includegraphics[width=.100\linewidth]{neurips2021/figures/qualitative/real/vid_00074_1.png} &
		\includegraphics[width=.100\linewidth]{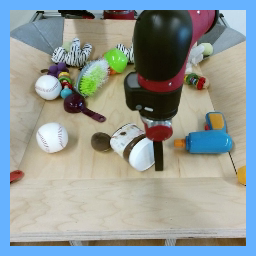} &
		\includegraphics[width=.100\linewidth]{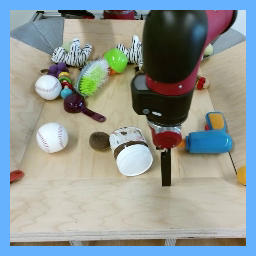} &
		\includegraphics[width=.100\linewidth]{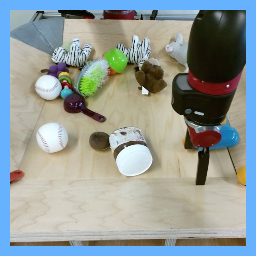} &
		\includegraphics[width=.100\linewidth]{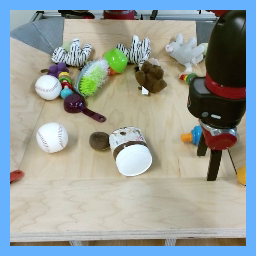} &
		\includegraphics[width=.100\linewidth]{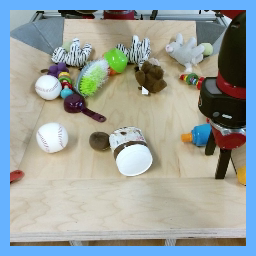} &
		\includegraphics[width=.100\linewidth]{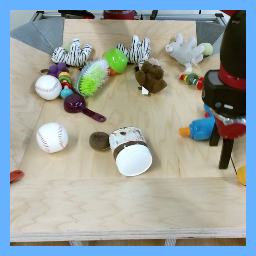} &
		\includegraphics[width=.100\linewidth]{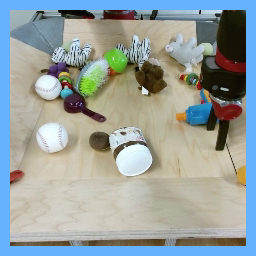} &
		\includegraphics[width=.100\linewidth]{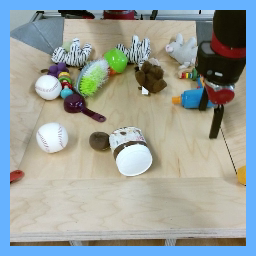} \\
	\end{tabular}
	\caption{Qualitative samples for different types of control on BAIR ($256 \times 256$). Zoom in for details.}
	\label{fig:bair-qualitative-256}
\end{figure}

\textbf{BAIR.} For future video prediction on BAIR (Table~\ref{tab:bair-prediction-64}), \method{} trained at $64 \stimes 64$ resolution (\textit{low res.}) is on par with the best method (L-size version of~\citep{weissenborn2020scaling}), but requires much less computing resources, and outperforms~\citep{weissenborn2020scaling} under similar resources. We also propose \textit{high res.} \method{} which is not strictly comparable to the prior arts as we use $256 \stimes 256$ image resolution for training and test, and resize the synthesized frames to $64 \stimes 64$ for computing FVD. However, using this variant demonstrates the performance gains that can arise by scaling \method{}. We additionally address point-to-point synthesis (with the end frame as a video-level annotation) and state-conditioned synthesis (with the estimated 2D position of the arm as a frame-level annotation). Point-to-point synthesis is more difficult than video prediction: Not only does it requires realistic video continuations, but also ones which explain the end position of all visible objects. Hence, FVD score is constant despite the additional input. Still, this yields better SSIM for mid-point and one-before-last frames. State-conditioned synthesis improves on FVD and mid-synthesis SSIM as motion becomes near-deterministic. Some synthetic frames for these tasks are shown in Figure~\ref{fig:bair-qualitative-256}. \method{} creates plausible high-quality videos in various settings, and true interactions with objects compared to previous attempts~\citep{menapace2021playable} at the same resolution. \revision{}

\begin{figure}\CenterFloatBoxes
\begin{floatrow}

\ttabbox{
 \centering
	\setlength\heavyrulewidth{0.25ex}
	\aboverulesep=0ex
    \belowrulesep=0.3ex
	\caption{Future video prediction on Kinetics ($64\stimes64$) of 16-frame videos from 5 consecutive input frames.}
	\label{tab:kinetics-prediction-64}
	\footnotesize
	\begin{tabular}{@{}L{3.40cm}@{}C{1.30cm}@{}C{2.60cm}@{}}
    	\toprule
        Method                                           & FVD $\downarrow$ & GPU/TPU Mem.\\
        \midrule
        LVT~\citep{rakhimov2020latent}                   & 225              & 128GB, 48h \\
        Video Transformer~\citep{weissenborn2020scaling} & $170_{\pm5}$     & 2TB, 336h$^*$     \\
        DVD-GAN-FP~\citep{clark2019adversarial}          & $69_{\pm1}$      & 2TB, 144h$^*$    \\
        \method{} (\emph{ours})                                       & $55_{\pm1}$               & 128GB, 300h      \\
        TriVD-GAN-FP~\citep{luc2020transformation}       & $26_{\pm1}$      & 16TB, 160h$^*$     \\
        \bottomrule
        \multicolumn{3}{c}{\scriptsize \hspace{2ex} $^*$: value confirmed by authors.}
    \end{tabular}
}{}

\ttabbox{
 \centering
	\setlength\heavyrulewidth{0.25ex}
	\aboverulesep=0ex
    \belowrulesep=0.3ex
	\caption{Codebook size on Kinetics.}
	\label{tab:kinetics-codebook-64}
	\footnotesize
	\begin{tabular}{@{}L{1.10cm}@{}
	C{1.10cm}@{}C{1.10cm}@{}C{1.10cm}@{}C{1.10cm}@{}}
    	\toprule
        \multirow{2}{*}{Size} & \multicolumn{3}{@{}c@{}}{Reconstruction} & Pred. \\
        \cmidrule(r){2-4} \cmidrule(l){5-5}
        & {\scriptsize FVD \hspace{-0.6ex} $\downarrow$} & {\scriptsize SSIM \hspace{-0.6ex} $\uparrow$} & {\scriptsize PSNR \hspace{-0.6ex} $\uparrow$} & {\scriptsize FVD \hspace{-0.6ex} $\downarrow$}\\
        \midrule
        $1024$ & $58_{\pm1}$ & $0.907$ & $31.2$ & $66_{\pm2}$ \\
        $4096$ & $54_{\pm1}$ & $0.917$ & $31.6$ & $64_{\pm1}$ \\
        $16384$ & $49_{\pm1}$ & $0.923$ & $32.0$ & $55_{\pm1}$ \\
        $65536$ & $45_{\pm1}$ & $0.928$ & $32.2$ & $61_{\pm1}$\\
        \hdashline
        $\infty$ & $12_{\pm1}$ & $0.963$ & $34.5$ & $229_{\pm1}$\\
        \bottomrule
        \multicolumn{5}{c}{~}
    \end{tabular}
}{}

\end{floatrow}
\end{figure}
\begin{figure}[h]
	\setlength\tabcolsep{1pt}
	\footnotesize
	\begin{tabular}{cccc}
	    $t=4$ & $t=8$ & $t=12$ & $t=16$ \\
	    
		\includegraphics[width=.100\linewidth]{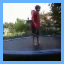} &
		\includegraphics[width=.100\linewidth]{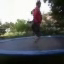} &
		\includegraphics[width=.100\linewidth]{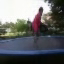} &
		\includegraphics[width=.100\linewidth]{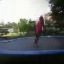}\\
		\includegraphics[width=.100\linewidth]{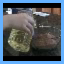} &
		\includegraphics[width=.100\linewidth]{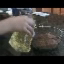} &
		\includegraphics[width=.100\linewidth]{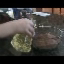} &
		\includegraphics[width=.100\linewidth]{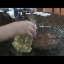}\\
		\includegraphics[width=.100\linewidth]{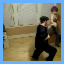} &
		\includegraphics[width=.100\linewidth]{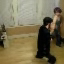} &
		\includegraphics[width=.100\linewidth]{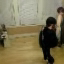} &
		\includegraphics[width=.100\linewidth]{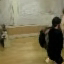}\\
	\end{tabular}
	\hspace{1cm}
	\begin{tabular}{cccc}
	    $t=4$ & $t=8$ & $t=12$ & $t=16$ \\
		\includegraphics[width=.100\linewidth]{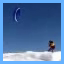} &
		\includegraphics[width=.100\linewidth]{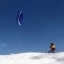} &
		\includegraphics[width=.100\linewidth]{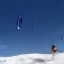} &
		\includegraphics[width=.100\linewidth]{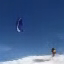}\\
		\includegraphics[width=.100\linewidth]{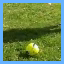} &
		\includegraphics[width=.100\linewidth]{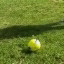} &
		\includegraphics[width=.100\linewidth]{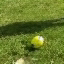} &
		\includegraphics[width=.100\linewidth]{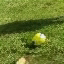}\\
		\includegraphics[width=.100\linewidth]{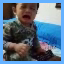} &
		\includegraphics[width=.100\linewidth]{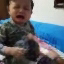} &
		\includegraphics[width=.100\linewidth]{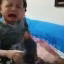} &
		\includegraphics[width=.100\linewidth]{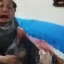}\\
	\end{tabular}
	\caption{Qualitative samples on future video prediction on Kinetics ($64 \times 64$). Zoom in for details.}
	\label{fig:kinetics-qualitative-64}
\end{figure}
\begin{table}
    \setlength\heavyrulewidth{0.25ex}
	\aboverulesep=0ex
    \belowrulesep=0.3ex
	\centering
	\caption{Sound-conditional video synthesis on AudioSet-Drums ($64 \stimes 64$).}
	\footnotesize
	\begin{tabular}{@{}L{2.80cm}@{}C{0.95cm}@{}C{0.95cm}@{}C{1.70cm}@{}C{1.70cm}@{}C{1.70cm}@{}C{1.30cm}@{}C{1.30cm}@{}C{1.30cm}@{}}
    	\toprule
        \multirow{2}{*}{Method} & \multirow{2}{*}{Cond.} & \multirow{2}{*}{Audio} & \multicolumn{3}{@{}c@{}}{SSIM \hspace{-1.0ex} $\uparrow$} & \multicolumn{3}{@{}c@{}}{PSNR \hspace{-1.0ex} $\uparrow$} \\
        \cmidrule(lr){4-6} \cmidrule(l){7-9}
        & & & {\scriptsize $t=16$} & {\scriptsize $t=30$} & {\scriptsize $t=45$} & {\scriptsize $t=16$} & {\scriptsize $t=30$} & {\scriptsize $t=45$} \\
        \midrule
        SVG-LP~\citep{denton2018stochastic} & 15 &  & $0.971_{\pm0.017}$ & $0.661_{\pm0.010}$ & $0.510_{\pm0.008}$ & $30.0_{\pm1.1}$ & $16.6_{\pm0.3}$ & $13.5_{\pm0.1}$ \\
        Vougioukas \textit{et al.}~\citep{vougioukas2018end} & 15 & \checkmark & $0.940_{\pm0.017}$ & $0.904_{\pm0.007}$ & $0.896_{\pm0.015}$ & $26.2_{\pm1.0}$ & $23.8_{\pm0.2}$ & $23.3_{\pm0.3}$\\
        Sound2Sight~\citep{cherian2020sound} & 15 & \checkmark & $0.984_{\pm0.009}$ & $0.954_{\pm0.007}$ & $\mathbf{0.947}_{\pm0.007}$ & $33.2_{\pm0.1}$ & $27.9_{\pm0.5}$ & $27.0_{\pm0.3}$ \\
        \method{} (\emph{ours})& 15 & \checkmark & $\mathbf{0.987}_{\pm0.001}$ & $\mathbf{0.956}_{\pm0.006}$ & $0.945_{\pm0.008}$ & $\mathbf{33.7}_{\pm0.4}$ & $\mathbf{28.4}_{\pm0.6}$ & $\mathbf{27.3}_{\pm0.5}$ \\
        \bottomrule
    \end{tabular}
	\label{tab:drum-sound-64}
\end{table}
\begin{figure}
	\setlength\tabcolsep{2.0pt}
	\renewcommand{\arraystretch}{1.}
	\footnotesize
	\begin{tabular}{ccccccccc}
	    & \multirow{2}{*}{DIV$^*$ $\uparrow$} & Input & Seed $1$ & Seed $2$ & Seed $3$ & Seed $4$ & Seed $5$ & Seed $6$ \\
	    & & $t=1$ & $t=15$ & $t=15$ & $t=15$ & $t=15$ & $t=15$ & $t=15$ \\
	    \cmidrule{2-9}
		\raisebox{-1.5\normalbaselineskip}[0pt][0pt]{\rotatebox[origin=c]{90}{BAIR}} & Whole frames &
		\multirow{4}{*}{\includegraphics[width=.100\linewidth]{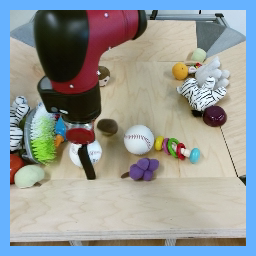}} &
		\multirow{4}{*}{\includegraphics[width=.100\linewidth]{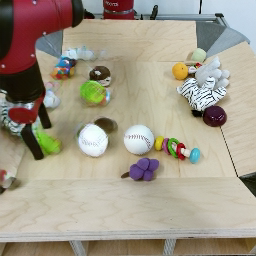}} &
		\multirow{4}{*}{\includegraphics[width=.100\linewidth]{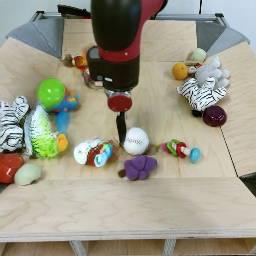}} &
		\multirow{4}{*}{\includegraphics[width=.100\linewidth]{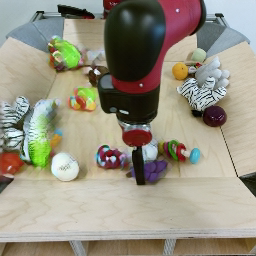}} &
		\multirow{4}{*}{\includegraphics[width=.100\linewidth]{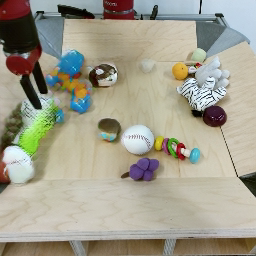}} &
		\multirow{4}{*}{\includegraphics[width=.100\linewidth]{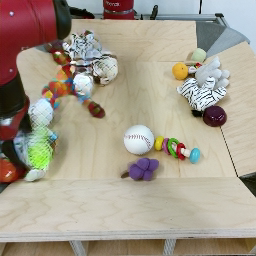}} &
		\multirow{4}{*}{\includegraphics[width=.100\linewidth]{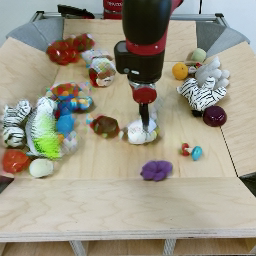}} \\
		& $19.53_{\pm2.60}$ &&&&&&& \\
		& Moving parts &&&&&&& \\
		& $136.88_{\pm12.27}$ &&&&&&& \\
		
		\cmidrule{2-9}
		\raisebox{-1.5\normalbaselineskip}[0pt][0pt]{\rotatebox[origin=c]{90}{Kinetics}} & Whole frames &
		\multirow{4}{*}{\includegraphics[width=.100\linewidth]{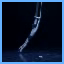}} &
		\multirow{4}{*}{\includegraphics[width=.100\linewidth]{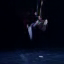}} &
		\multirow{4}{*}{\includegraphics[width=.100\linewidth]{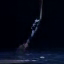}} &
		\multirow{4}{*}{\includegraphics[width=.100\linewidth]{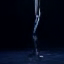}} &
		\multirow{4}{*}{\includegraphics[width=.100\linewidth]{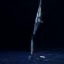}} &
		\multirow{4}{*}{\includegraphics[width=.100\linewidth]{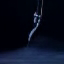}} &
		\multirow{4}{*}{\includegraphics[width=.100\linewidth]{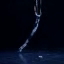}} \\
		& $20.85_{\pm2.90}$ &&&&&&& \\
		& Moving parts &&&&&&& \\
		& $48.16_{\pm13.83}$ &&&&&&& \\
		
		\cmidrule{2-9}
		\raisebox{-1.5\normalbaselineskip}[0pt][0pt]{\rotatebox[origin=c]{90}{\parbox{4cm}{\centering AudioSet-\\Drums}}} & Whole frames &
		\multirow{4}{*}{\includegraphics[width=.100\linewidth]{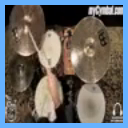}} &
		\multirow{4}{*}{\includegraphics[width=.100\linewidth]{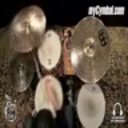}} &
		\multirow{4}{*}{\includegraphics[width=.100\linewidth]{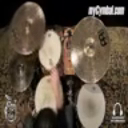}} &
		\multirow{4}{*}{\includegraphics[width=.100\linewidth]{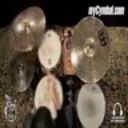}} &
		\multirow{4}{*}{\includegraphics[width=.100\linewidth]{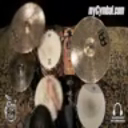}} &
		\multirow{4}{*}{\includegraphics[width=.100\linewidth]{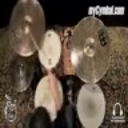}} &
		\multirow{4}{*}{\includegraphics[width=.100\linewidth]{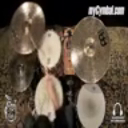}} \\
		& $2.25_{\pm0.26}$ &&&&&&& \\
		& Moving parts &&&&&&& \\
		& $65.45_{\pm6.00}$ &&&&&&& \\
		
		\multicolumn{9}{c}{\scriptsize$^*$: the pixel-wise distances are obtained by applying a $1.0\times10^{-3}$ factor (omitted for clarity).}

	\end{tabular}
	\caption{\revision{Diversity results for conditioned scenarios. The diversity metric (DIV) measures the mean pixel-wise distance among 10 15-frame synthetic trajectories conditioned on the same frame. We report the mean and std over 100 runs: on whole frames as in~\citep{yang2019diversity}, or moving parts only by masking static regions (optical flow magnitude between consecutive frames $<20$\% of the max magnitude) as in~\citep{villegas2017decomposing}. We show, for the same input, the $15$\textsuperscript{th} frame of various randomly seeded synthetic trajectories.}}
	\label{fig:div}
\end{figure}

\textbf{Kinetics.} \method{} ranks second on Kinetics video prediction benchmark. Kinetics contains more diversity than BAIR, and the reconstruction is thus more difficult. A solution~\citep{ramesh2021zero} is to increase the codebook size (Table~\ref{tab:kinetics-codebook-64}) but it stops translating into better prediction FVD at some point. We also try removing the quantization step (equivalent to an infinite codebook), and directly regressing latent features with $\transformer$ (instead of ranking the likelihood of possible tokens). It allows better reconstructions, yet the prediction FVD is high. \revision{Figure~\ref{fig:kinetics-qualitative-64} shows examples of synthetic continuations conditioned on 5-frame unseen test sequences. \method{} produces realistic and temporally coherent outputs which display various types of motion (\eg, body, hand, camera).}

\textbf{AudioSet-Drums.} \method{} achieves top performance on sound-conditioned video synthesis on AudioSet Drums (Table~\ref{tab:drum-sound-64}). \revision{Figure~\ref{fig:div} shows quantitative and qualitative insights on the diversity of the synthetic trajectories conditioned on the same input for the three datasets. The diversity metrics (DIV) computed on whole frames is lower on AudioSet Drums than on the other two datasets. This is explained by the fact that motion is quite repetitive and involves a limited portion of the frame. The same metric on moving parts only, and the end position of the drummer's hand and upper left cymbal in qualitative samples, show the diversity of synthetic trajectories. An ablation of \method{} with/without audio guidance as well as more qualitative results on diversity can be found in Appendix~\ref{sec:additional-results}.}

\section{Discussion}
\label{sec:discussion}

\method{} is on par or better than the state-of-the-art on standard benchmarks, uses less computational resources, and scales to high resolution. Training neural networks is environmentally costly, due to the carbon footprint to power processing hardware~\citep{strubell2019energy, dhar2020the}. Methods sparing GPU-hours like ours are crucial to make AI less polluting~\citep{strubell2019energy, lacoste2019quantifying}, and move from a ``Red'' to a ``Green'' AI~\citep{schwartz2020green}.
Future work will include exploring new codebook strategies and synthesis guided by textual information.

\paragraph{Limitations.} \revision{\method{} uses a complex architecture and a two-stage training strategy. Simplification of both is an interesting direction for improving the method. Moreover, \method{} lacks global regularization of motion (flow computed on pairs of timesteps), and its efficiency relies on recycling context information such that synthesizing content from scratch (\ie, no input frame given) remains difficult.}

\paragraph{Broader impact.} The increased accessibility and the many controls \method{} offers could accelerate the emergence of questionable applications, such as ``deepfakes'' (\eg, a video created from someone’s picture and an arbitrary audio) which could lead to harassment, defamation, or dissemination of fake news.  
On top of current efforts to automate their detection~\citep{nguyen2019deep}, it remains our responsibility to grow awareness of these possible misuses.
Despite these worrying aspects, our contribution has plenty of positive applications which outweigh the potential ethical harms. Our efficient compression scheme is a step in the direction of real-time solutions: \eg, enhancing human-robot interactions, or improving the safety of self-driving cars by predicting the trajectories of people and vehicles nearby. 
\section*{Acknowledgements}

This work was granted access to the HPC resources of IDRIS under the allocation 2020-AD011012227 made by GENCI. It was funded in part by the French government under management of Agence Nationale de la Recherche as part of the ``Investissements d’avenir'' program, reference ANR-19-P3IA-0001 (PRAIRIE 3IA Institute). JP was supported in part by the Louis Vuitton/ENS chair in artificial intelligence and the Inria/NYU collaboration. \revision{We thank the reviewers for useful comments.}

{\small
\bibliographystyle{plain}
\bibliography{neurips_2021}
}

\newpage
\appendix
\section{Detailed architecture design}
\label{sec:architecture}
\setcounter{table}{0}
\renewcommand{\thetable}{A\arabic{table}}
\setcounter{figure}{0}
\renewcommand{\thefigure}{A\arabic{figure}}

\begin{figure}[H]
    \centering
    \begin{tabular}{c@{}c}
        \multicolumn{1}{c}{\includegraphics[width=0.73\linewidth, valign=c]{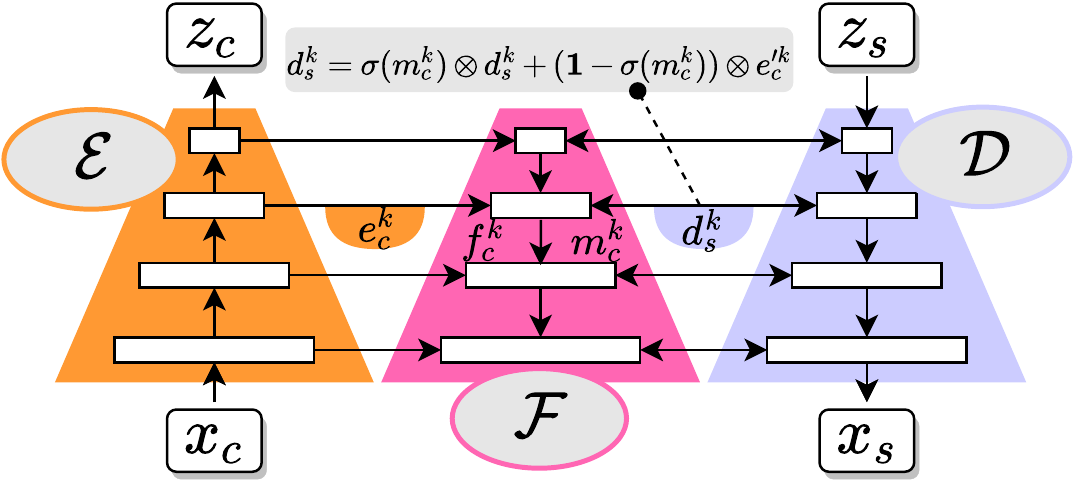}}
         & \includegraphics[width=0.21\linewidth, valign=c]{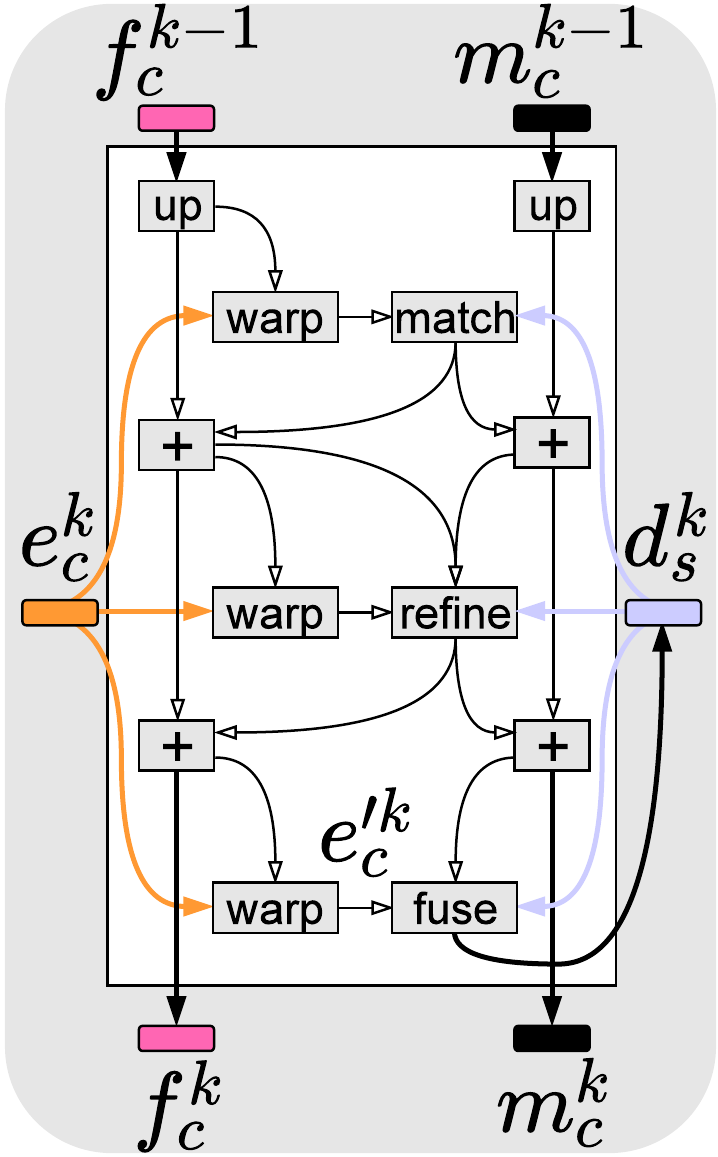} \\
         ~ \\ [-0.5em]
        \shortstack{(1) Conversion from image to feature space ($\encoder$) and vice versa ($\decoder$).\\ In the later case, context frames help producing new ones via $\flowestimator$.} & \shortstack{(2) Computations in $\flowestimator$ \\ at one resolution level.} \\
    \end{tabular}
    
    \hspace{0.2cm}
    
	\caption{In (1), illustration of the encoding ($\encoder$) and decoding ($\decoder$) architectures, and the interactions between the two thanks to the progressive optical flow and fusion mask estimation operator ($\flowestimator$). As opposed to $\encoder$, $\decoder$ processes features from low to high spatial resolution. To allow feature updates from context frame $x_c$ in the decoding phase of a latent embedding $z_s$, we first compute encoded features $e_c^k$ for all intermediate resolution levels $(r_k)_{k\in\llbracket 1, K\rrbracket}$ by applying $\encoder$ to $x_c$. We then decode $x_s=\decoder(z_s)$ from low ($k=1$) to high resolution ($k=K$) by iterating over the following steps: \textbf{(a)} apply one decoding sub-module to get $d_s^k$ from $d_s^{k-1}$, \textbf{(b)} use $d_s^k$ and $e_c^k$ (at corresponding levels) to estimate the flow $f_c^k$ from $e_c^k$ to $d_s^k$ and a fusion mask $m_c^k$, \textbf{(c)} use $f_c^k$ and $m_c^k$ to update $d_s^k$ with $e_c^k$ (see update rule), and \textbf{(d)} move to the next resolution level by going back to \textbf{(a)}. The residual computation of $f_c^k$ and $m_c^k$ from $e_c^k$, $d_s^k$, $f_c^{k-1}$ and $m_c^{k-1}$ is shown in (2) and detailed in the text.}
	\label{fig:skip-ae-sup}
\end{figure}

We describe the implementation choices for the different functional blocks which compose \method{}. Note that different parts of the framework may benefit from future advances in various research areas: autoregressive modeling, image synthesis, codebook learning and optical flow estimation for transformer $\transformer$, decoder $\decoder$, quantizer $\quantizer$ and flow module $\flowestimator$ respectively.

\paragraph{Encoder and decoder.}
Both the encoder $\encoder$ and the decoder $\decoder$, illustrated in Figure~\ref{fig:skip-ae-sup}-(1), use a sequence of $K$ modules handling information at different \revision{intermediate resolution levels $(r_k)_{k\in \llbracket 1;K \rrbracket}$} (with reverse ordering from $\encoder$ to $\decoder$). Each module is implemented by a residual block~\citep{he2016deep} which increases or decreases the spatial resolution by a factor $2$. In practice, we use the succession of two $3 \times 3$ 2D convolutions with LeakyReLU activations and a skip connection in the form of a $1 \times 1$ convolution.

\paragraph{Flow module.} 
The flow module $\flowestimator$, also illustrated in Figure~\ref{fig:skip-ae-sup}-(1), allows the sharing of information between the encoding and decoding stages across different timesteps. Computations for one resolution level $r_k$ are detailed in Figure~\ref{fig:skip-ae-sup}-(2): The ``match'' operation computes the correlation between features in $d_s^k$ and its closest spatial neighbours in $e_c^k$. \revision{We thus obtain a 3D cost volume of size $N \times h_k \times w_k$, where $N$ is the number of neighbours considered and $(h_k \times w_k)=r_k$ is the spatial resolution.} We apply 2D convolutions to this cost volume to deduce the residual flow and mask. The ``refine'' operation is also a residual update of $f_c^k$ and $m_c^k$. It is made of 2D convolutions and takes as input the concatenation of features, mask and flow. Finally, the ``fuse'' operation update features from $\decoder$ with information from the context in $\encoder$ according to the update rule given in Figure~\ref{fig:skip-ae-sup}-(1) and detailed in Section~\ref{sec:ae}. For these operations, we take our inspiration from LiteFlowNet~\citep{hui2018liteflownet} which we adapt to handle not only the optical flow but also the proposed fusion mask. An extension to multi-frame context for this temporal skip connection mechanism is proposed in Appendix~\ref{sec:extension}.

\paragraph{Image and temporal discriminators.} The architecture of the image discriminator $\imagedis$ is similar to that of StyleGAN2~\citep{karras2020analyzing}, and the temporal discriminator $\videodis$ is essentially the same, with 3D instead of 2D convolutions to account for the temporal dimension. To avoid the propagation of errors when decoding frames autoregressively during inference (see the ``shift-and-add'' module introduced for this purpose in Figure~\ref{fig:skip-ae}), we mimic this iterative process at train time: decode frames one by one, then encode them back to obtain context features for the next steps. The resulting synthetic videos are fed to $\videodis$. To avoid excessive memory consumption due to the recursive encoding / decoding operations, we only keep the gradients for the context features from the immediate preceding frame.

\paragraph{Quantizer.} The implementation of the quantizer $\quantizer$ is straightforward since it is just using a codebook of learnable embeddings. Yet, the non-differentiable nature of this operation prevents gradients from training losses (introduced in Section~\ref{sec:ae}) to backpropagate through $\mathcal{U}$ and $\quantizer$, and from $\decoder$ to $\encoder$ at the compression bottleneck. To solve this in practice, we use a straight-through gradient estimator~\citep{bengio2013estimating} which ignores the quantization step during backward-pass.

\paragraph{Transformer.} The transformer $\transformer$ is described in Section~\ref{sec:transformer}. Its input is obtained by transforming a sequence of tokens into a 2D matrix of size $L \times D$. For that, $\transformer$ associates with each possible token and each of the $L$ positions in the sequence a learnable embedding of size $D$.
To leverage the prior information that successive fixed-sized portions of the sequence represent annotated frames, we propose a spatio-temporal decomposition of the positional embeddings with a term indicating the timestep, and another one indicating the position within the frame-related portion. This is illustrated in Figure~\ref{fig:gpt}. The input matrix is computed by adding the token embedding and the positional embedding for each token of the sequence. $\transformer$ thus consists in the succession of multi-head self-attention layers, and position-wise fully connected layers.

\paragraph{Position estimator.}
In the experiments (Section~\ref{sec:experiments}), we mention that we manually annotated the $(x, y)$ location of the robotic arm (more precisely, its gripper) in 500 frames of the BAIR dataset. We use these to train a position estimator so that we can extract the trajectory of the arm from any new video (and use this trajectory to guide the state-conditioned synthesis of new videos). To facilitate training and accelerate inference, we design our position estimator to exploit latent features (obtained from images with $\encoder$) and to map them to the $(x,y)$ position of the arm. We use a simple architecture with a few downsampling convolutional layers and a fully connected layer to output the estimated position. The position estimator is trained by regressing the target 2D coordinates by minimizing the corresponding mean squared error.

\paragraph{Sound autoencoder.} To allow multimodal synthesis of video and sound, we first compute the short-time Fourier transform (STFT) of the raw audio to obtain a time-frequency representation of the sound. We then employ an encoding / quantization / decoding strategy similar to that used to construct $\encoder$, $\quantizer$ and $\decoder$ for video frames (without $\flowestimator$ in this case) to reduce dimension and obtain corresponding tokens for audio. These tokens are used as frame-level annotations to guide the synthesis, just as in state-conditioned synthesis. 

\section{Role of the flow module}
\label{sec:role}
\setcounter{table}{0}
\renewcommand{\thetable}{B\arabic{table}}
\setcounter{figure}{0}
\renewcommand{\thefigure}{B\arabic{figure}}
\setcounter{algocf}{0}
\renewcommand{\thealgocf}{B\arabic{algocf}}

\revision{Looking at Figure~\ref{fig:skip-ae-sup}, one may wonder whether the flow module $\flowestimator$ alone is sufficient for predicting features for future frames which would result in $z_s$ potentially being completely ignored. This would only apply if a frame was fully determined from the preceding one, yet, this is not the case due to the inherent stochastic nature of future prediction in the considered datasets. We have conducted a simple experiment to demonstrate that features $z_s$ are actually used and that they actively drive the dynamics of the scene: We have repeated the ``reconstruction'' experiment from Table~\ref{tab:ae-ablation} for which we compare the final model (last line) with an ablation with fixed $z_s$ (\ie, fixing $z_s=z_c$ for all predicted timesteps). Results are available in Table~\ref{tab:bair-fix-256}.}

\begin{table}
    \setlength\heavyrulewidth{0.25ex}
	\aboverulesep=0ex
    \belowrulesep=0.3ex
	\centering
	\caption{\revision{Reconstruction of 16-frame videos from compressed features ($z_s$) and the first frame on BAIR ($256 \stimes 256$). Features $z_s$ are either fixed (copied from the first timestep for subsequent timesteps) or dynamic (the true ones, obtained by applying $\encoder$ frame-by-frame to the video).}}
	\footnotesize
	\begin{tabular}{@{}L{2.00cm}@{}C{1.20cm}@{}C{1.20cm}@{}}
    	\toprule
    	$z_s$ & {FVD \hspace{-1.0ex} $\downarrow$} & {PSNR \hspace{-1.0ex} $\uparrow$} \\
    	\midrule
    	Fixed & $556_{\pm12}$ & $18.9$ \\
    	Dynamic & $45_{\pm1}$ & $26.8$ \\
    	\bottomrule
    \end{tabular}
	\label{tab:bair-fix-256}
\end{table}

\revision{We observe that the reconstruction performance is much poorer when we force $z_s$ to remain constant for all timesteps. Qualitatively, looking at the synthesized videos, we see that they do not exhibit any motion, as one would expect when fixing the encoded features. Moreover, thanks to the fusion masks estimated by the flow module $\flowestimator$ and used in Eq.~(\ref{eq:update}), it is possible to observe how much information comes from $z_s$ and from $\flowestimator$ respectively when generating $x_s$. Figure~\ref{fig:city-aug-256} shows some examples of estimated masks on Cityscapes. Those are white when the source is $z_s$, and black when it is context through $\flowestimator$. In practice, we see that they are white when the context is occluded and mostly grey in other regions. Thus, final videos are as much the result of context warping than direct decoding in non-occluded regions for this dataset. We observe similar behaviours on other datasets.}

\section{Flow inversion approximation}
\label{sec:inversion}
\setcounter{table}{0}
\renewcommand{\thetable}{C\arabic{table}}
\setcounter{figure}{0}
\renewcommand{\thefigure}{C\arabic{figure}}
\setcounter{algocf}{0}
\renewcommand{\thealgocf}{C\arabic{algocf}}

For affine transformations, the exact inverse flow can be determined analytically.
This is not the case for elastic deformations for which we propose to approximate this inverse flow $f \in \mathbb{R}^{2 \times H \times W}$ from an input flow $g \in \mathbb{R}^{2 \times H \times W}$ (where $H \times W$ is the image spatial resolution).
Here, $g$ maps cells from a grid $\llbracket 1; H \rrbracket \times \llbracket 1; W \rrbracket$ to cells from the same grid (we ignore cells for which $g$ points out of the grid since they do not help in computing $f$). Such a mapping is not surjective, that is, all the cells in the grid are not necessarily reached. We compute the pixel-accurate inversion in cells for which there exists a direct mapping, and approximate others by iterative interpolation. That is, we keep track of inverted cells with a completion mask $c$ which we update with the closest cells in cardinal directions at every step.
For the elastic deformations that we use when training \method{}, around 80\% of the grid can be directly inverted in average, which is enough to accurately reconstruct the missing cells by interpolation.

\begin{algorithm}[H]
\SetAlgoLined
\KwData{Flow $g \in \mathbb{R}^{2 \times H \times W}$}
\KwResult{Approximation of inverted flow $f \in \mathbb{R}^{2 \times H \times W}$ }
\tcc{initialization of flow and cell-completion mask}
 $f\,=\textbf{0} \in \mathbb{R}^{2 \times H \times W}$\;
 $c=\textbf{0} \in \{0; 1\}^{H \times W}$\;
 \tcc{invert flow in cells for which there exists a direct mapping}
 \For{$(h, w) \in \llbracket 1; H \rrbracket \times \llbracket 1; W \rrbracket$}{
 $dh\,=g_{0,h,w}$\;
 $dw=g_{1,h,w}$\;
 $h'\,\,=\round(h+dh)$\;
 $w'\,=\round(w+dw)$\;
 \If{$(h', w') \in \llbracket 1; H \rrbracket \times \llbracket 1; W \rrbracket$}{
 $f_{0, h', w'}=-dh$\;
 $f_{1, h', w'}=-dw$\;
 $c_{h', w'}\,\,=1$\;
 }
 }
 \tcc{fill empty cells iteratively by interpolating neighbours' flow}
 \While{$\exists (h, w) \in \llbracket 1; H \rrbracket \times \llbracket 1; W \rrbracket \mid \, \sim{c_{h,w}}$}{
  $c'=\dilate(c) \land \sim{c}$\;
  $i\,\,\,\,\,=\blur(f)$\;
  $w\,\,\,=\blur(c)$\;
  \For{$(h, w) \in \llbracket 1; H \rrbracket \times \llbracket 1; W \rrbracket \mid  c'_{h,w}$}{
  $f_{0, h, w}=i_{0, h, w} / w_{h, w}$\;
  $f_{1, h, w}=i_{1, h, w} / w_{h, w}$\;
  }
  
  $c=c \lor c'$\;
 }
\caption{Flow inversion approximation.}
\label{alg:flow-invert}
\end{algorithm}
The blurring function $\blur:\RR^{H \times W}\rightarrow\RR^{H \times W}$ is a 2D convolution with $3\times3$ Gaussian kernel (and $0$-valued $1$-sized padding to preserve the spatial resolution), and interpolates flow values from neighbouring cells weighted by the their proximity; Dilation function $\dilate:\{0; 1\}^{H \times W}\rightarrow\{0; 1\}^{H \times W}$ propagates $1$-valued cells in a 2D boolean mask according to the cardinal directions (up, down, left, right); $\land$, $\lor$ and $\sim$ are notations which designate logical \textit{AND}, \textit{OR} and \textit{NOT} respectively.

\revision{We note that an alternative to flow inversion is to use the flow module $\flowestimator$ to compute the backward flow. This requires an extra forward pass in the flow module to compute both forward and backward flows and would result in additional GPU computations. Not to slow down training we prefer to run flow inversion on parallelized CPU processes as part of data loading. While both alternatives take approximately the same amount of time (0.15s on a GPU for backward flow estimation for a batch of 16 images at resolution $256 \times 256$, 0.12s on parallelized CPU for flow inversion for the same input) flow inversion is advantageous because, in our setup, GPU consumption is the limiting factor for speed (fully utilized GPUs, data loading on CPUs can run in the background).}

\section{Extension of the flow module to multi-frame context}
\label{sec:multi}
\setcounter{table}{0}
\renewcommand{\thetable}{D\arabic{table}}
\setcounter{figure}{0}
\renewcommand{\thefigure}{D\arabic{figure}}
\label{sec:extension}

We extend the flow-module (introduced in Section~\ref{sec:ae}) which aims at reusing contextual information from single- to multi-frame context. We thus consider a context $(x_i)_{i \in \llbracket 1;c \rrbracket}$ consisting of $c$ frames. Just as before, for a given level $r_k$ and $i$ in $\llbracket 1;c \rrbracket$, we use $\flowestimator$ to compute the corresponding fusion mask $m_i^k$ and optical flow $f_i^k$ between intermediate encoded features $e_i^k$ (from context frame $x_i$) and the decoded features $d_s^k$ (that we wish to update). We recall that fusion masks $m_i^k$ handle occlusion by indicating for each spatial location the relevance of warped context features $e_i'^k=\warp(e_i^k, f_i^k)$, that is, whether features $e_i'^k$ correspond to $d_s^k$ at that location in terms of content, or not. Hence, we can derive from the fusion masks a confidence score $s_i^k$ defined as the context-wise normalization of features relevance, and use it to aggregate all the information that can be recovered from context by applying a weighted average:
\begin{minipage}{.5\linewidth}
\vspace{-0.2cm}
\begin{equation*}
    s_i^k=(\mathbf{1}-\sigma(m_i^k)) \oslash \left(\sum_{j=1}^{c}(\mathbf{1}-\sigma(m_j^k))\right),
\end{equation*}
\end{minipage}%
\begin{minipage}{.5\linewidth}
\begin{equation}
\begin{cases}
    m_{a}^k = \sum_{i=1}^{c}s_i^k \otimes m_i^k\\
    e_a'^k = \sum_{i=1}^{c}s_i^k \otimes e_i'^k
\end{cases}.
\end{equation}
\end{minipage}
where $\oslash$ represents the element-wise division. In other words, given a spatial location, $m_a^k$ is an estimate for the likelihood that the content of $d_s^k$ matches any of the $c$ context frames, and $e_a'^k$ is used to reconstruct this content faithfully by weighting proposals from the context by their relevance. This brings us back to the problem setup of Section~\ref{sec:ae}, so that the aggregated fusion mask $m_a^k$ and warped features $e_a'^k$ can be used to update $d_s^k$, just as in single-frame fusion.
The key benefits of this extension are the use of redundancy within context to better reconstruct the current frame, and the greater robustness to occlusion as different views are combined. This is demonstrated by the ablation study of the autoencoder in Table~\ref{tab:ae-ablation}.

\section{Additional results}
\label{sec:additional-results}
\setcounter{table}{0}
\renewcommand{\thetable}{D\arabic{table}}
\setcounter{figure}{0}
\renewcommand{\thefigure}{D\arabic{figure}}

\begin{table}[!h]
    \setlength\heavyrulewidth{0.25ex}
	\aboverulesep=0ex
    \belowrulesep=0.3ex
	\centering
	\caption{Unconditional video synthesis of 16-frame videos on UCF-101 ($128\stimes128$).}
	\footnotesize
	\begin{tabular}{@{}lccc@{}}
    	\toprule
        Method                                           & IS $\uparrow$ & FVD $\downarrow$\\
        \midrule
        StyleGAN2~\citep{karras2020analyzing} (repeat $\times16$) & $17.98_{\pm.12}$ & ${990_{\pm33}}$ \\
        Real frame (repeat $\times16$) & $28.41_{\pm.11}$ & ${838_{\pm27}}$ \\
        \hdashline
        VGAN~\citep{vondrick2016generating}              & $8.31_{\pm.09}$ & -  \\
        TGAN~\citep{saito2017temporal}              & $11.85_{\pm.07}$ & -  \\
        MoCoGAN~\citep{tulyakov2018mocogan}              & $12.42_{\pm.07}$ & -  \\
        ProgressiveVGAN~\citep{acharya2018towards}              & $14.56_{\pm.05}$ & -  \\
        LDVD-GAN~\citep{kahembwe2020lower}              & $22.91_{\pm.19}$ & -  \\
        TGANv2~\citep{saito2020train}              & $26.60_{\pm.47}$ & $1209_{\pm28}$  \\
        DVD-GAN~\citep{clark2019adversarial}              & $27.38_{\pm.53}$ & -  \\
        MoCoGAN-HD~\citep{tian2021good}              & $33.95_{\pm.25}$ & $700_{\pm24}$  \\
        StyleGAN2~\citep{karras2020analyzing} + \method{} (\emph{ours}) & $24.47_{\pm.13}$ & ${386_{\pm15}}$ \\ 
        \hdashline
        Real frame + \method{} (\emph{ours}) & $41.37_{\pm.39}$ & ${389_{\pm14}}$ \\
        \bottomrule
    \end{tabular}
	\label{tab:ucf101-unconditional-256}
\end{table}

\paragraph{Unconditional synthesis.} We apply \method{} to unconditional video synthesis (\ie, the production of new videos without prior information about their content) on UCF-101~\citep{khurram2012ucf101}, an action-recognition dataset of $101$ categories for a total of approximately $13$k video clips. Our approach leverages recent progress made in image synthesis by using StyleGAN2~\citep{karras2020analyzing} to produce the first frame. \method{} is able to improve over the state of the art on the FVD metric (computed on 2048 videos, under the same evaluation process as related approaches), but has an average performance ($24.47$) on the inception score (IS)~\citep{saito2020train} (computed on $10$k videos, again with the appropriate protocol). IS measures the coverage of the different categories, and whether one of these is clearly identifiable in each of the synthetic videos, through the lens of the Softmax scores of a C3D network~\citep{tran2015learning} trained for action recognition on the Sports-1M dataset~\citep{karpathy2014large} and fine-tuned on UCF-101. We also evaluate the performance of \method{} when considering a perfect image synthesizer (or one that has overfitted the training data) by using a real frame instead of a synthetic one to launch the prediction. This yields similar FVD but considerable improvements on IS ($41.37$). Note that \method{} favorably contribute to IS in this setting because videos made from the same real frames (by repeating them along the temporal axis) have much lower IS ($28.41$). A plausible explanation why MoCoGAN-HD~\citep{tian2021good} achieves a good IS score is due to it sampling multiple times from StyleGAN2~\citep{karras2020analyzing} (as opposed to once in our case, we use the same pretrained model). Indeed, MoCoGAN-HD synthesizes videos by finding a trajectory in the latent space of StyleGAN2. In this case, it may be easier to recognize a category because different modes can be interpolated (greater IS) but the outcome is less temporally realistic than ours (FVD near the one of still videos for MoCoGAN-HD). Future work will include trying to reconciliate both approaches.

\begin{figure}
	\setlength\tabcolsep{2.0pt}
	\renewcommand{\arraystretch}{1.0}
	\footnotesize
	\begin{tabular}{C{0.10cm}ccccc}
	    && $t=1$ & $t=10$ & $t=20$ & $t=30$ \\
	    [0.05cm]
		\raisebox{1.75\normalbaselineskip}[0pt][0pt]{\rotatebox[origin=c]{90}{Layout}} & 
		~ &
		\includegraphics[width=.230\linewidth]{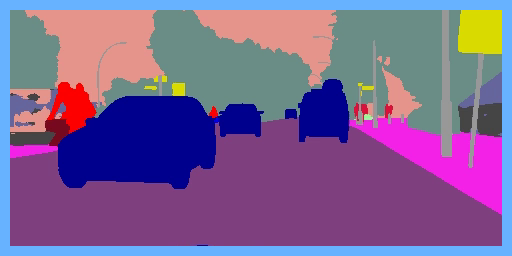} &
		\includegraphics[width=.230\linewidth]{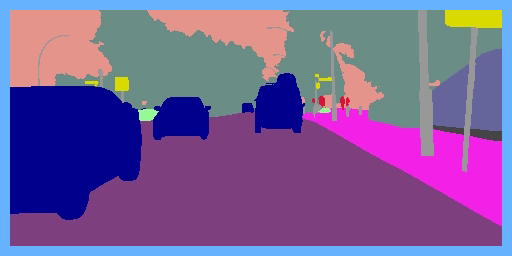} &
		\includegraphics[width=.230\linewidth]{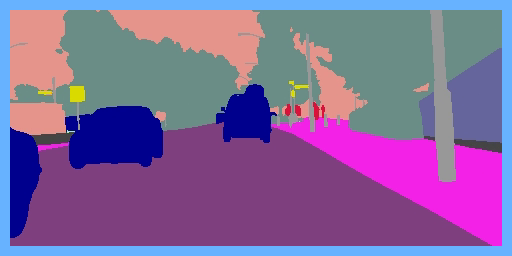} &
		\includegraphics[width=.230\linewidth]{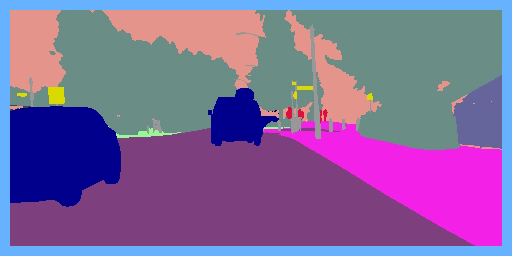} \\
		
		\raisebox{1.85\normalbaselineskip}[0pt][0pt]{\rotatebox[origin=c]{90}{Image}} & 
		~ &
		\includegraphics[width=.230\linewidth]{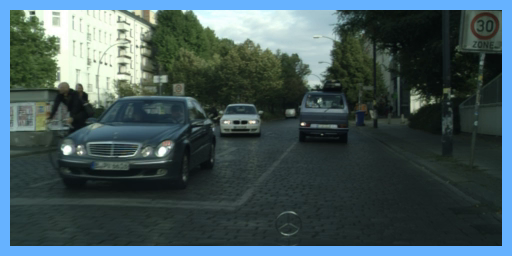} &
		\includegraphics[width=.230\linewidth]{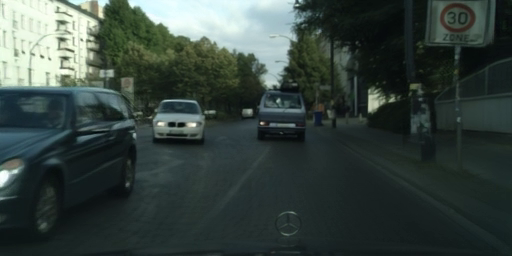} &
		\includegraphics[width=.230\linewidth]{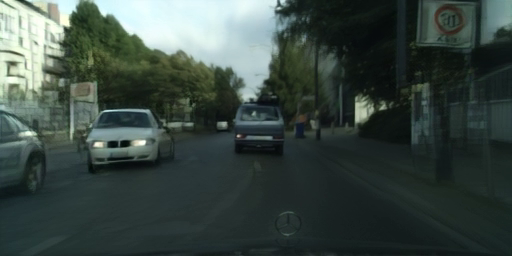} &
		\includegraphics[width=.230\linewidth]{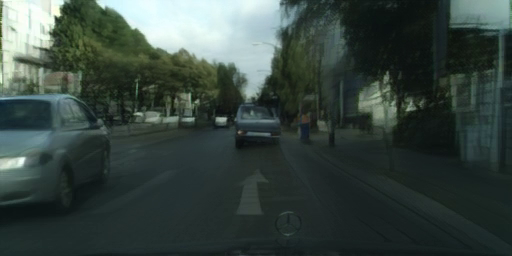} \\
		[0.3cm]
		\raisebox{1.75\normalbaselineskip}[0pt][0pt]{\rotatebox[origin=c]{90}{Layout}} & 
		~ &
		\includegraphics[width=.230\linewidth]{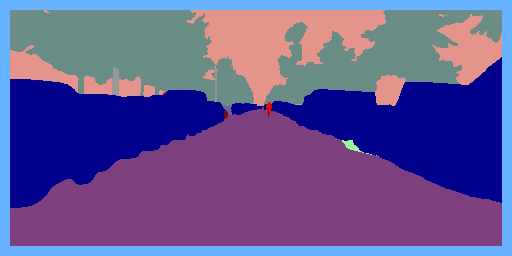} &
		\includegraphics[width=.230\linewidth]{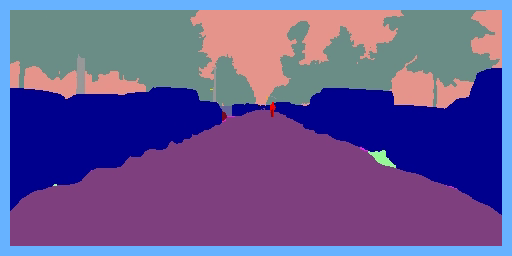} &
		\includegraphics[width=.230\linewidth]{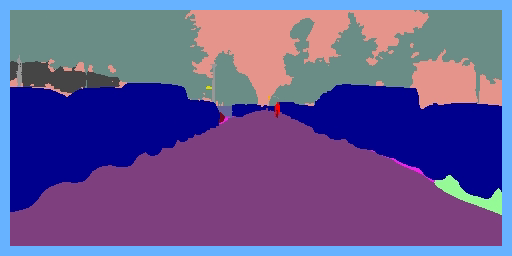} &
		\includegraphics[width=.230\linewidth]{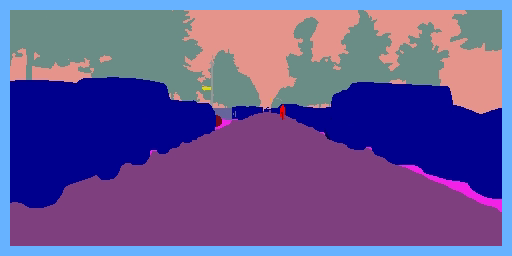} \\
		
		\raisebox{1.85\normalbaselineskip}[0pt][0pt]{\rotatebox[origin=c]{90}{Image}} & 
		~ &
		\includegraphics[width=.230\linewidth]{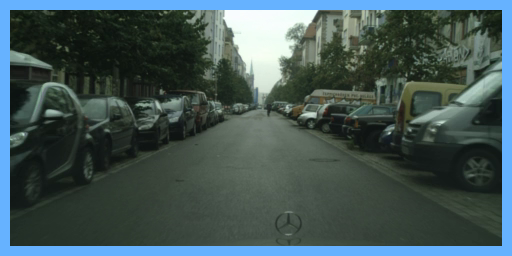} &
		\includegraphics[width=.230\linewidth]{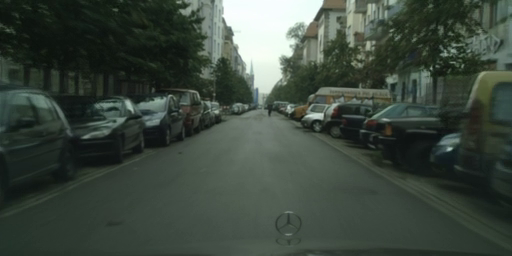} &
		\includegraphics[width=.230\linewidth]{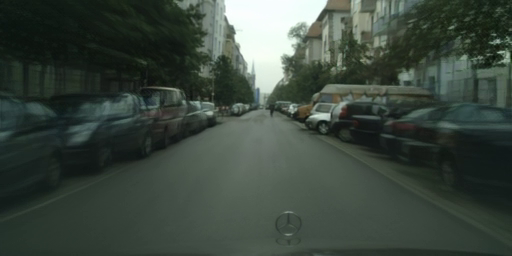} &
		\includegraphics[width=.230\linewidth]{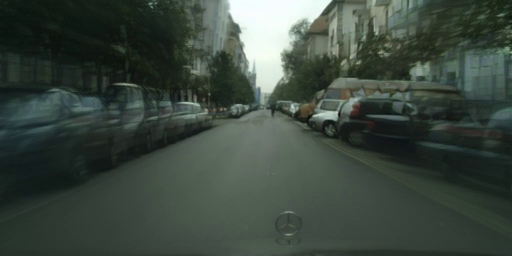} \\
		[0.3cm]
		\raisebox{1.75\normalbaselineskip}[0pt][0pt]{\rotatebox[origin=c]{90}{Layout}} & 
		~ &
		\includegraphics[width=.230\linewidth]{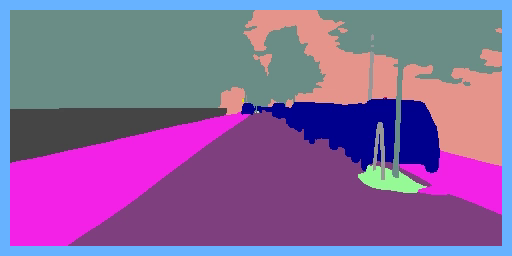} &
		\includegraphics[width=.230\linewidth]{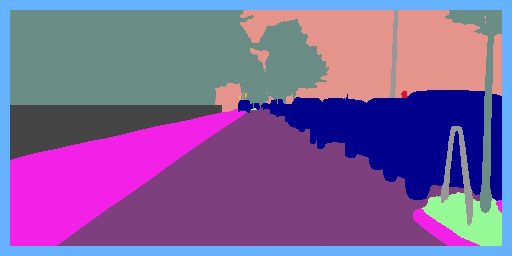} &
		\includegraphics[width=.230\linewidth]{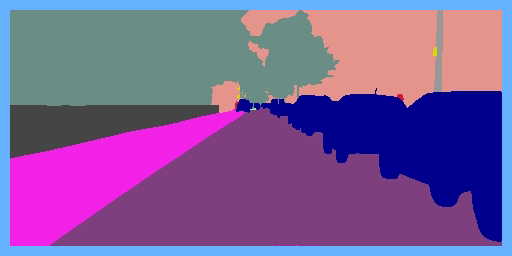} &
		\includegraphics[width=.230\linewidth]{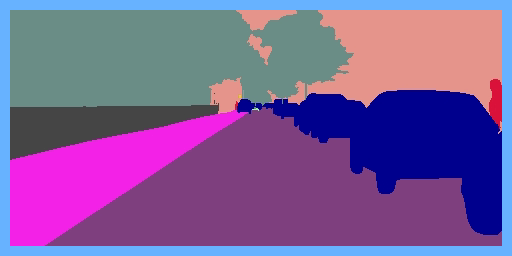} \\
		
		\raisebox{1.85\normalbaselineskip}[0pt][0pt]{\rotatebox[origin=c]{90}{Image}} & 
		~ &
		\includegraphics[width=.230\linewidth]{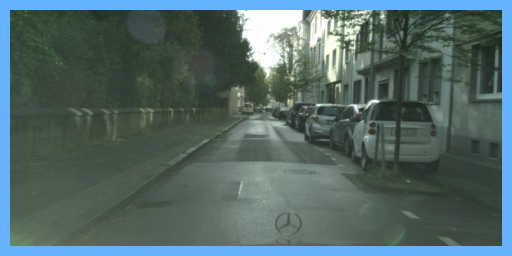} &
		\includegraphics[width=.230\linewidth]{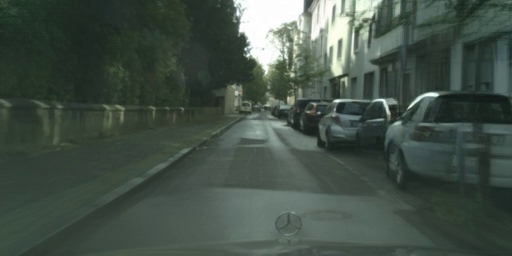} &
		\includegraphics[width=.230\linewidth]{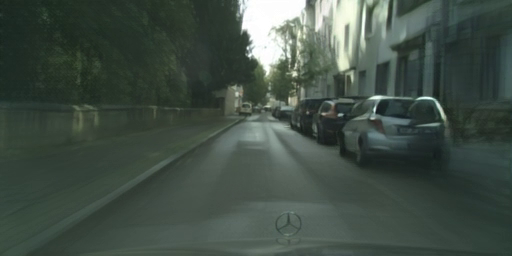} &
		\includegraphics[width=.230\linewidth]{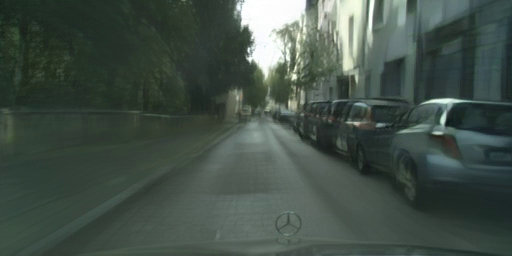} \\
		[0.3cm]
		\raisebox{1.75\normalbaselineskip}[0pt][0pt]{\rotatebox[origin=c]{90}{Layout}} & 
		~ &
		\includegraphics[width=.230\linewidth]{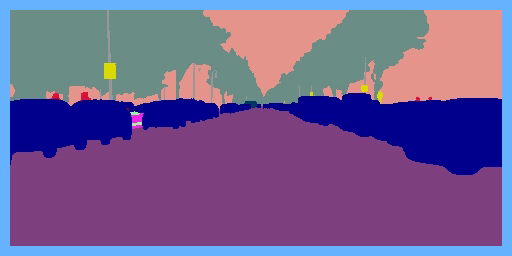} &
		\includegraphics[width=.230\linewidth]{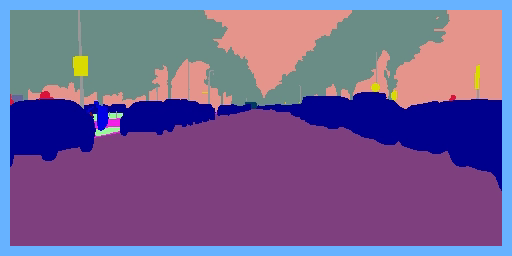} &
		\includegraphics[width=.230\linewidth]{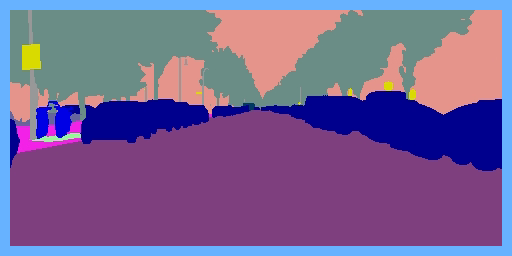} &
		\includegraphics[width=.230\linewidth]{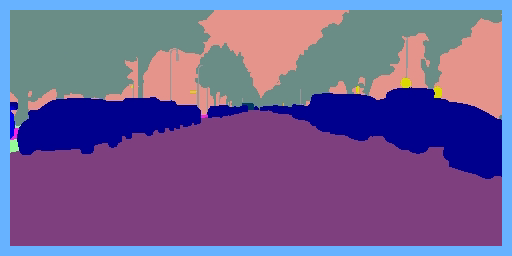} \\
		
		\raisebox{1.85\normalbaselineskip}[0pt][0pt]{\rotatebox[origin=c]{90}{Image}} & 
		~ &
		\includegraphics[width=.230\linewidth]{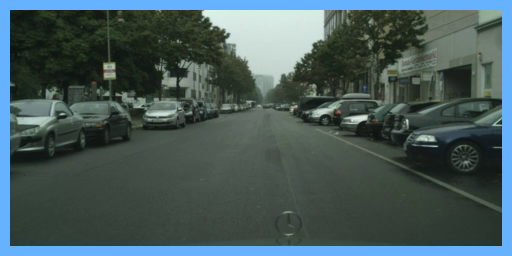} &
		\includegraphics[width=.230\linewidth]{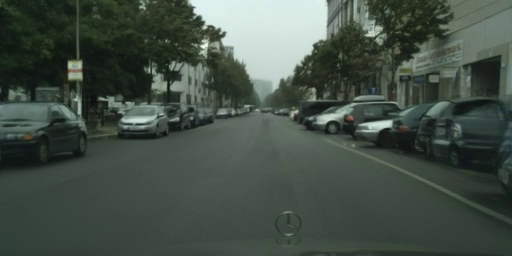} &
		\includegraphics[width=.230\linewidth]{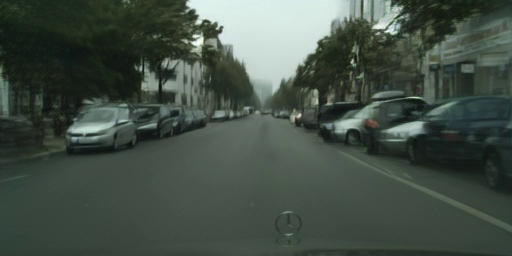} &
		\includegraphics[width=.230\linewidth]{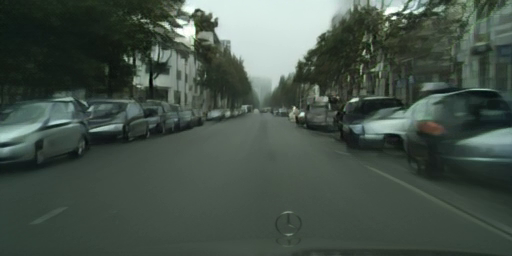} \\
		[0.3cm]
		\raisebox{1.75\normalbaselineskip}[0pt][0pt]{\rotatebox[origin=c]{90}{Layout}} & 
		~ &
		\includegraphics[width=.230\linewidth]{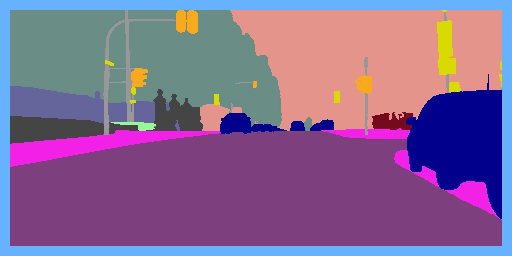} &
		\includegraphics[width=.230\linewidth]{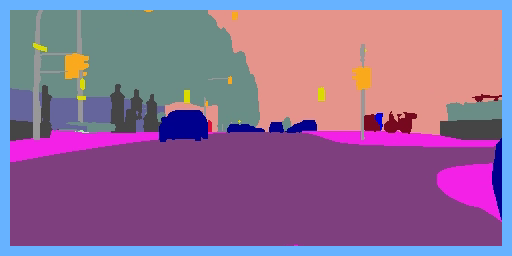} &
		\includegraphics[width=.230\linewidth]{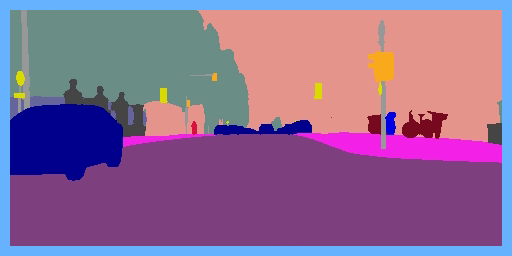} &
		\includegraphics[width=.230\linewidth]{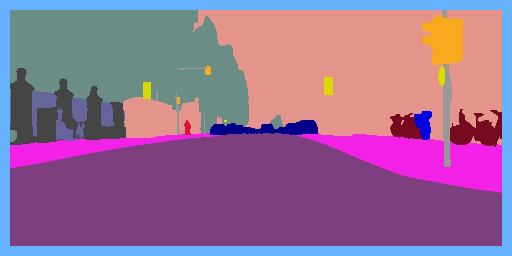} \\
		
		\raisebox{1.85\normalbaselineskip}[0pt][0pt]{\rotatebox[origin=c]{90}{Image}} & 
		~ &
		\includegraphics[width=.230\linewidth]{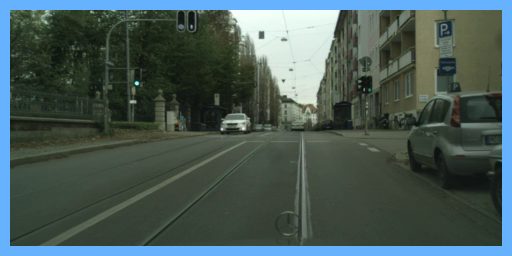} &
		\includegraphics[width=.230\linewidth]{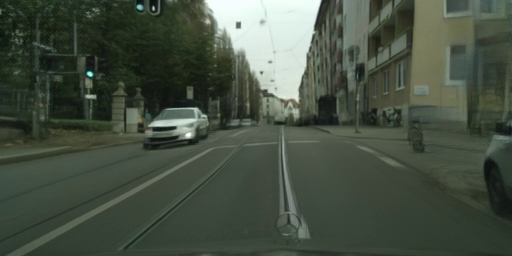} &
		\includegraphics[width=.230\linewidth]{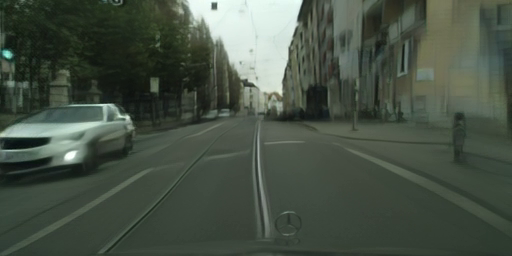} &
		\includegraphics[width=.230\linewidth]{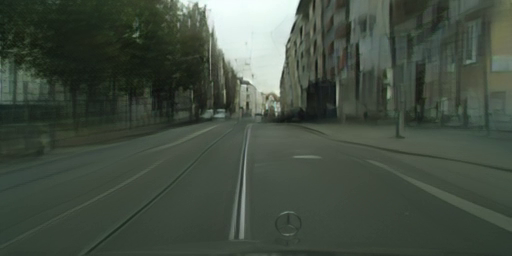} \\
		
	\end{tabular}
	\caption{Layout-conditioned synthesis of 30-frame videos given 3 starting frames and layouts for all timesteps on Cityscapes ($256 \times 512$). We show different samples of real layout and synthetic image pairs at intermediate timesteps. We see that synthesized frames follow both the target semantic structure specified by the layouts and the texture extracted from the conditioning frames.}
	\label{fig:city-layout-256}
\end{figure}

\begin{figure}
	\setlength\tabcolsep{2.0pt}
	\renewcommand{\arraystretch}{1.0}
	\footnotesize
	\begin{tabular}{C{0.10cm}ccccc}
	    && $t=1$ & $t=10$ & $t=20$ & $t=30$ \\
	    [0.05cm]
		\raisebox{1.75\normalbaselineskip}[0pt][0pt]{\rotatebox[origin=c]{90}{Layout}} & 
		~ &
		\includegraphics[width=.230\linewidth]{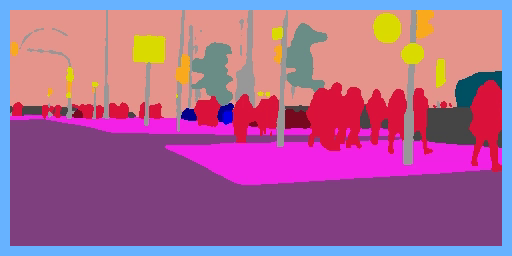} &
		\includegraphics[width=.230\linewidth]{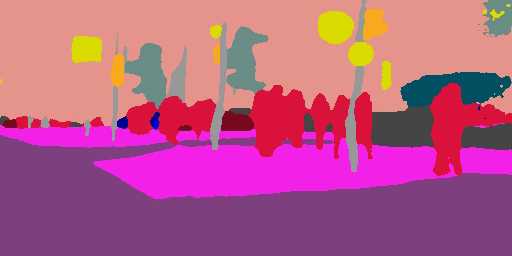} &
		\includegraphics[width=.230\linewidth]{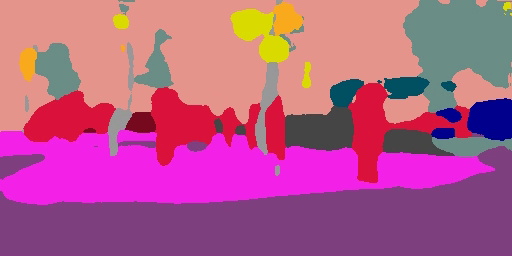} &
		\includegraphics[width=.230\linewidth]{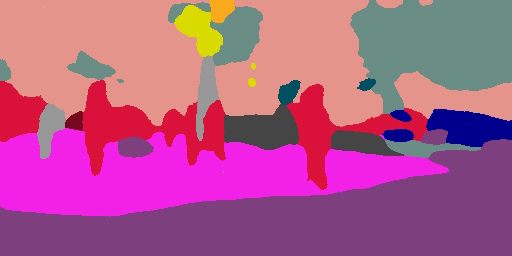} \\
		
		\raisebox{1.85\normalbaselineskip}[0pt][0pt]{\rotatebox[origin=c]{90}{Image}} & 
		~ &
		\includegraphics[width=.230\linewidth]{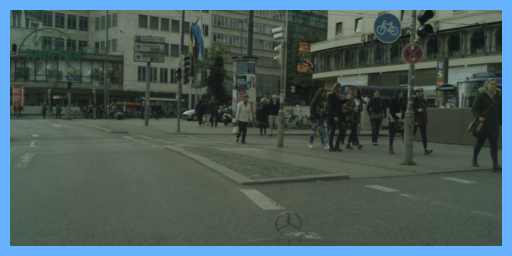} &
		\includegraphics[width=.230\linewidth]{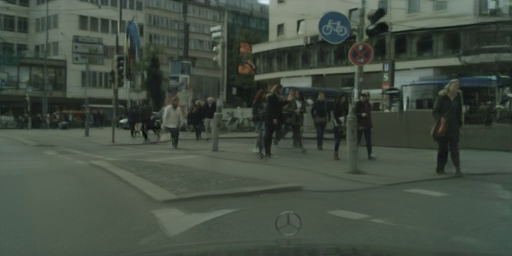} &
		\includegraphics[width=.230\linewidth]{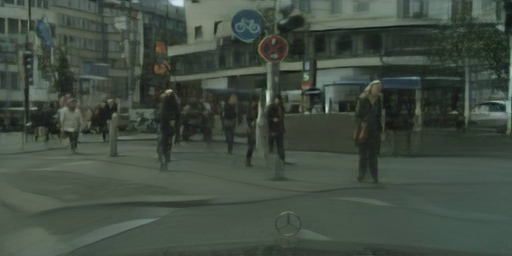} &
		\includegraphics[width=.230\linewidth]{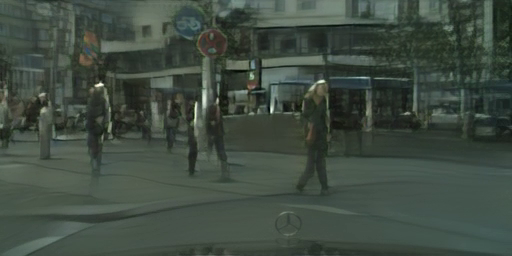} \\
		[0.3cm]
		\raisebox{1.75\normalbaselineskip}[0pt][0pt]{\rotatebox[origin=c]{90}{Layout}} & 
		~ &
		\includegraphics[width=.230\linewidth]{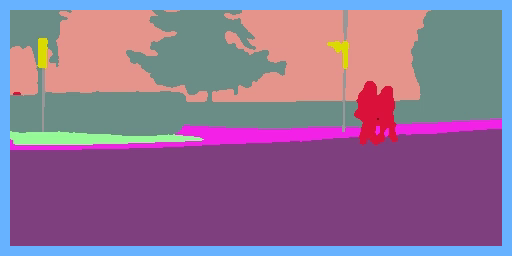} &
		\includegraphics[width=.230\linewidth]{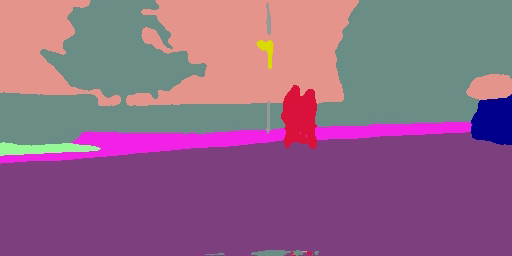} &
		\includegraphics[width=.230\linewidth]{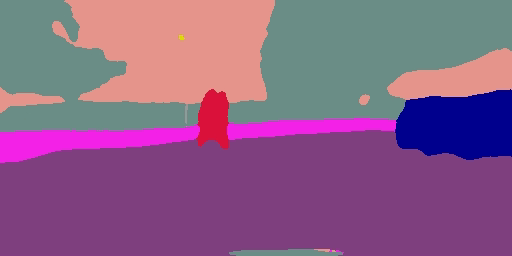} &
		\includegraphics[width=.230\linewidth]{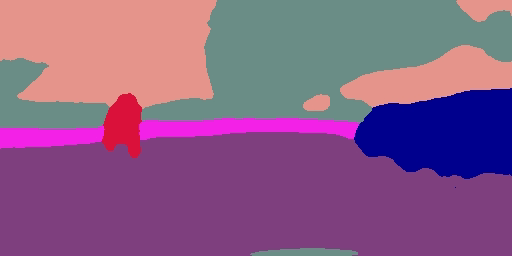} \\
		
		\raisebox{1.85\normalbaselineskip}[0pt][0pt]{\rotatebox[origin=c]{90}{Image}} & 
		~ &
		\includegraphics[width=.230\linewidth]{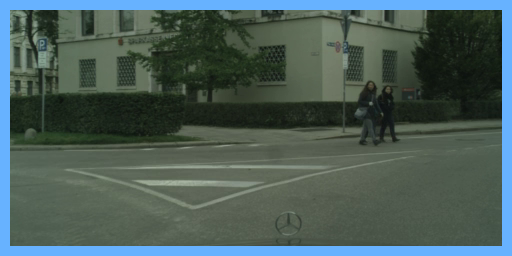} &
		\includegraphics[width=.230\linewidth]{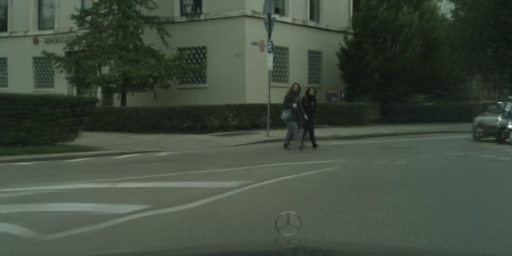} &
		\includegraphics[width=.230\linewidth]{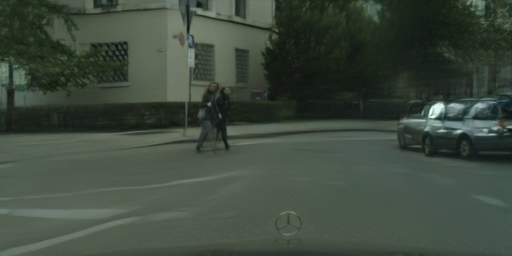} &
		\includegraphics[width=.230\linewidth]{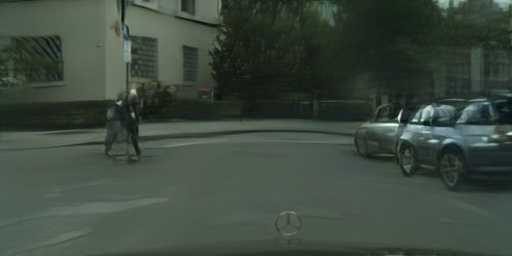} \\
		[0.3cm]
		\raisebox{1.75\normalbaselineskip}[0pt][0pt]{\rotatebox[origin=c]{90}{Layout}} & 
		~ &
		\includegraphics[width=.230\linewidth]{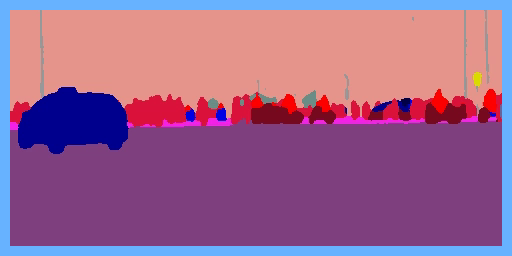} &
		\includegraphics[width=.230\linewidth]{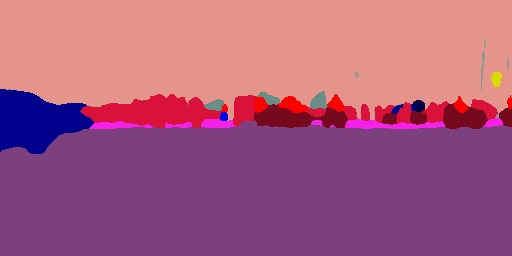} &
		\includegraphics[width=.230\linewidth]{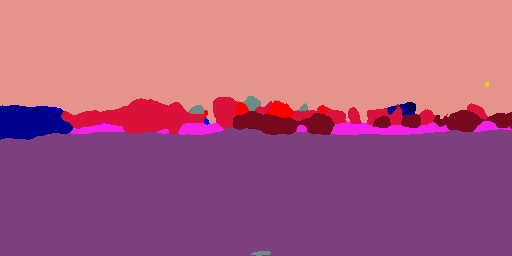} &
		\includegraphics[width=.230\linewidth]{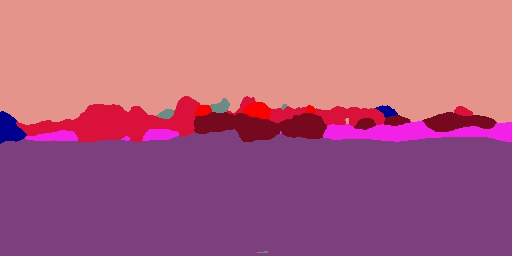} \\
		
		\raisebox{1.85\normalbaselineskip}[0pt][0pt]{\rotatebox[origin=c]{90}{Image}} & 
		~ &
		\includegraphics[width=.230\linewidth]{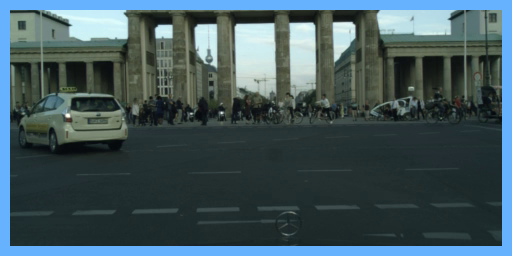} &
		\includegraphics[width=.230\linewidth]{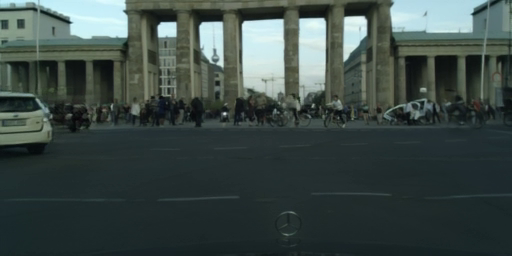} &
		\includegraphics[width=.230\linewidth]{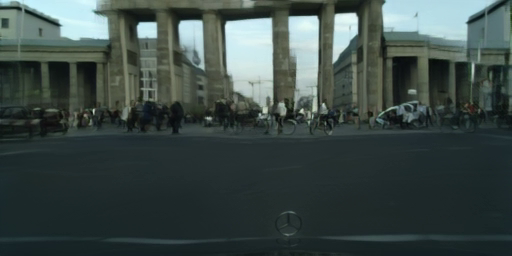} &
		\includegraphics[width=.230\linewidth]{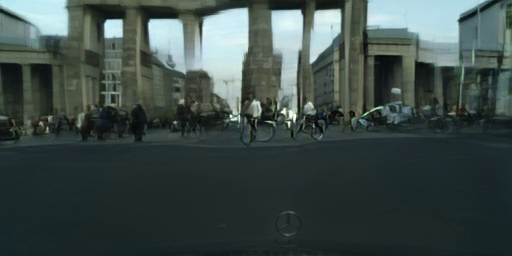} \\
		[0.3cm]
		\raisebox{1.75\normalbaselineskip}[0pt][0pt]{\rotatebox[origin=c]{90}{Layout}} & 
		~ &
		\includegraphics[width=.230\linewidth]{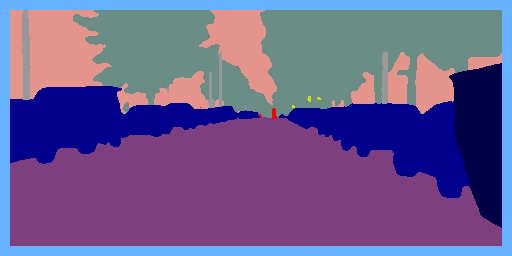} &
		\includegraphics[width=.230\linewidth]{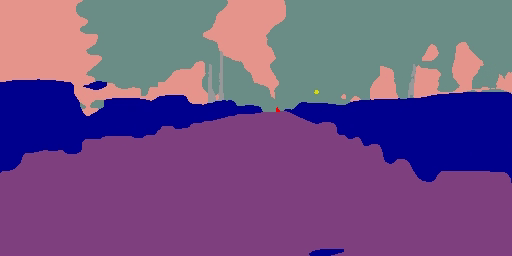} &
		\includegraphics[width=.230\linewidth]{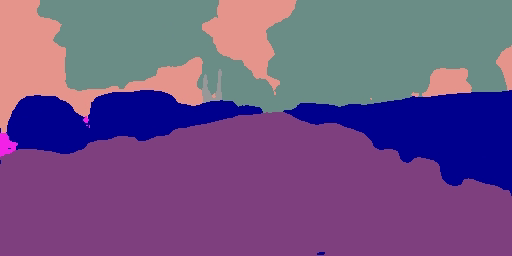} &
		\includegraphics[width=.230\linewidth]{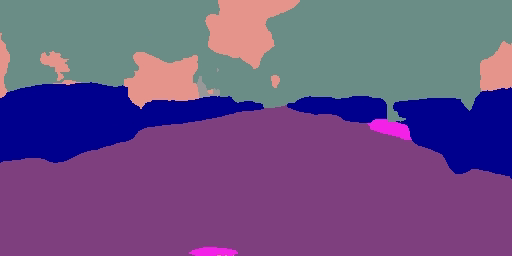} \\
		
		\raisebox{1.85\normalbaselineskip}[0pt][0pt]{\rotatebox[origin=c]{90}{Image}} & 
		~ &
		\includegraphics[width=.230\linewidth]{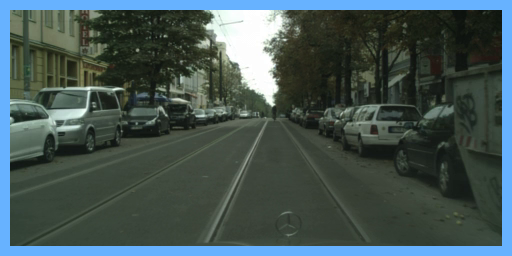} &
		\includegraphics[width=.230\linewidth]{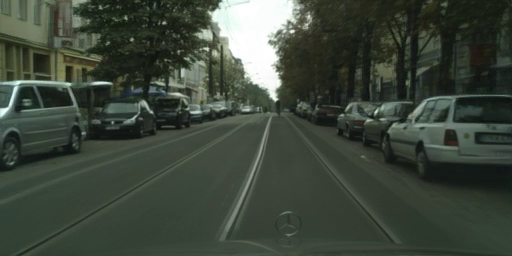} &
		\includegraphics[width=.230\linewidth]{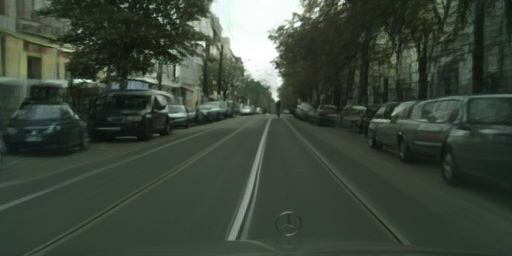} &
		\includegraphics[width=.230\linewidth]{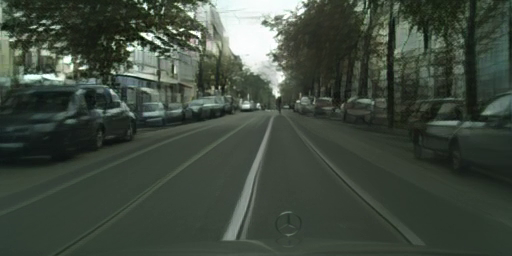} \\
		[0.3cm]
		\raisebox{1.75\normalbaselineskip}[0pt][0pt]{\rotatebox[origin=c]{90}{Layout}} & 
		~ &
		\includegraphics[width=.230\linewidth]{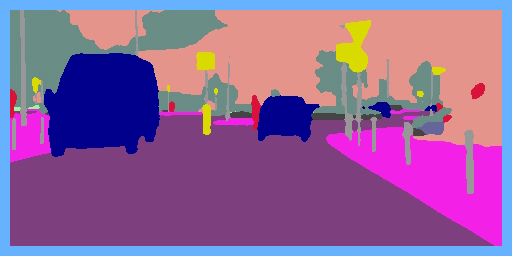} &
		\includegraphics[width=.230\linewidth]{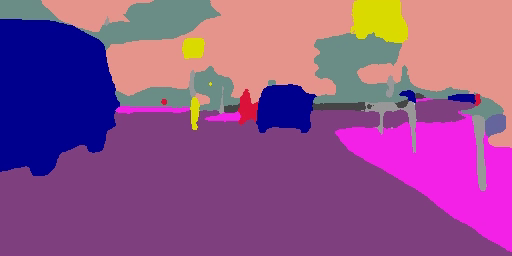} &
		\includegraphics[width=.230\linewidth]{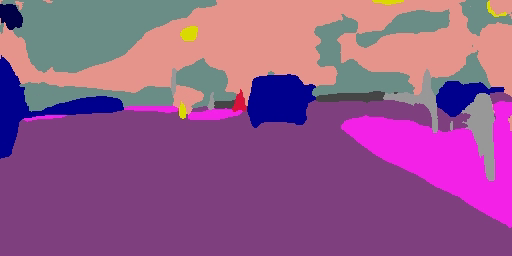} &
		\includegraphics[width=.230\linewidth]{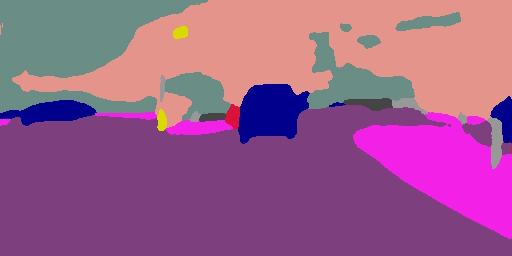} \\
		
		\raisebox{1.85\normalbaselineskip}[0pt][0pt]{\rotatebox[origin=c]{90}{Image}} & 
		~ &
		\includegraphics[width=.230\linewidth]{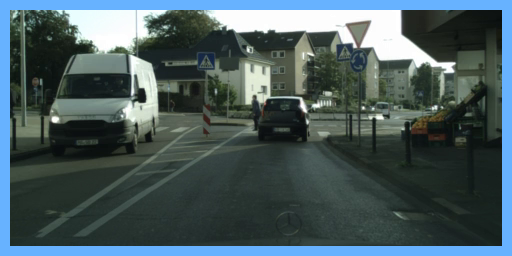} &
		\includegraphics[width=.230\linewidth]{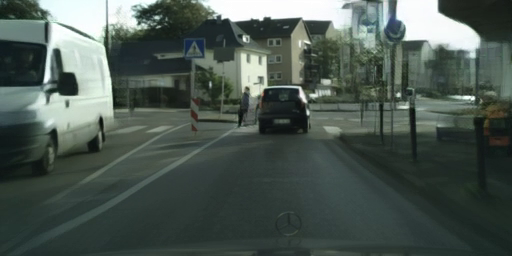} &
		\includegraphics[width=.230\linewidth]{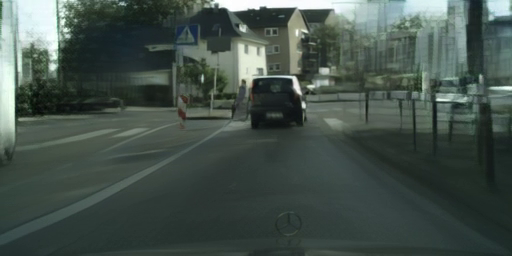} &
		\includegraphics[width=.230\linewidth]{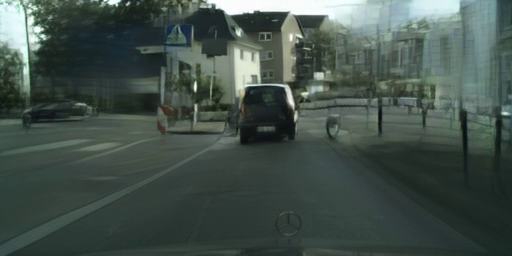} \\
		
	\end{tabular}
	\caption{Future video synthesis of 30-frame videos given 3 starting layout-frame pairs on Cityscapes ($256 \times 512$). We show different samples of synthetic layout and image pairs at intermediate timesteps. In this case, layouts are a byproduct of the synthesis process, \ie, they are synthesized alongside predicted frames.}
	\label{fig:city-layout-off-256}
\end{figure}

\begin{figure}
	\setlength\tabcolsep{2.0pt}
	\renewcommand{\arraystretch}{1.0}
	\footnotesize
	\begin{tabular}{C{0.10cm}ccccc}
	    && \method{} & \method{} + \citep{park2019semantic} & Real image & Real layout \\
	    [0.05cm]
		\raisebox{1.75\normalbaselineskip}[0pt][0pt]{\rotatebox[origin=c]{90}{Full}} & 
		~ &
		\includegraphics[width=.230\linewidth]{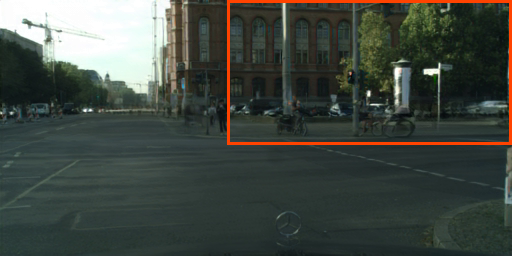} &
		\includegraphics[width=.230\linewidth]{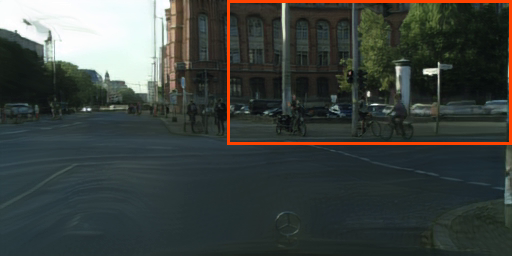} &
		\includegraphics[width=.230\linewidth]{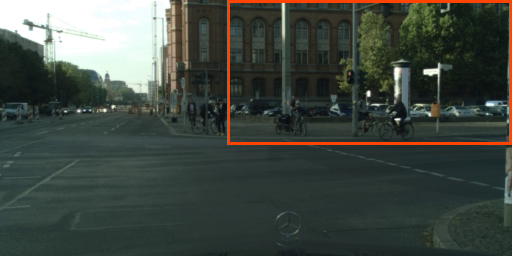} &
		\includegraphics[width=.230\linewidth]{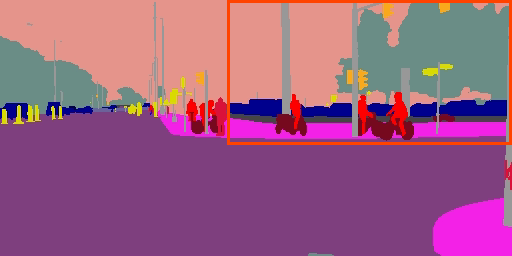} \\
		\raisebox{1.75\normalbaselineskip}[0pt][0pt]{\rotatebox[origin=c]{90}{Zoom}} & 
		~ &
		\includegraphics[width=.230\linewidth]{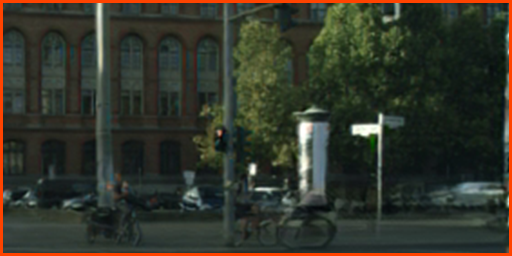} &
		\includegraphics[width=.230\linewidth]{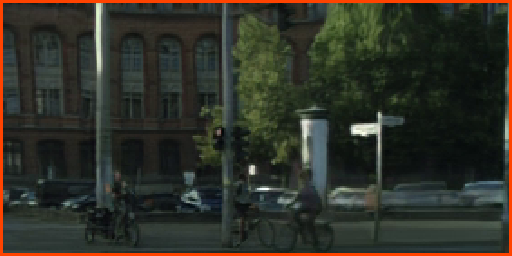} &
		\includegraphics[width=.230\linewidth]{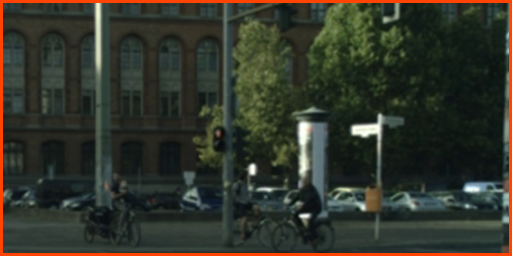} &
		\includegraphics[width=.230\linewidth]{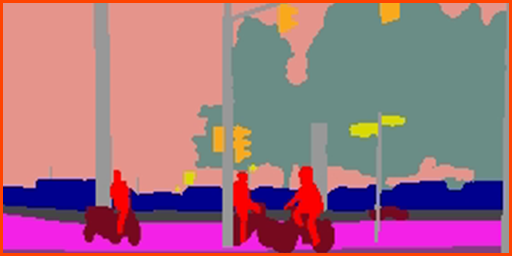} \\
		[0.3cm]
		\raisebox{1.75\normalbaselineskip}[0pt][0pt]{\rotatebox[origin=c]{90}{Full}} & 
		~ &
		\includegraphics[width=.230\linewidth]{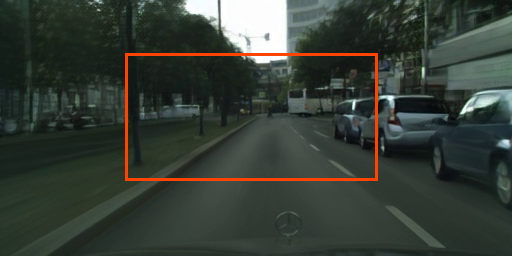} &
		\includegraphics[width=.230\linewidth]{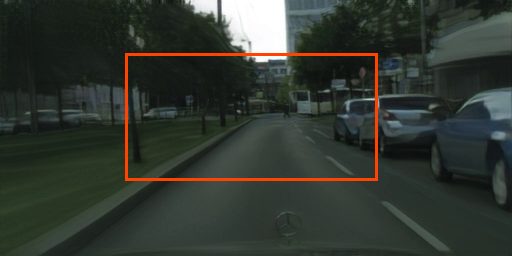} &
		\includegraphics[width=.230\linewidth]{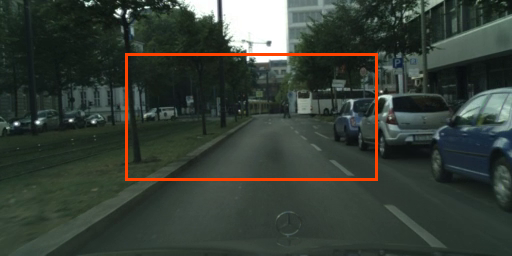} &
		\includegraphics[width=.230\linewidth]{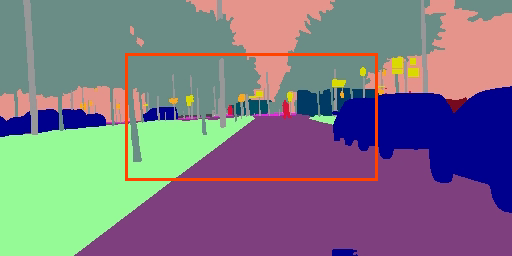} \\
		\raisebox{1.75\normalbaselineskip}[0pt][0pt]{\rotatebox[origin=c]{90}{Zoom}} & 
		~ &
		\includegraphics[width=.230\linewidth]{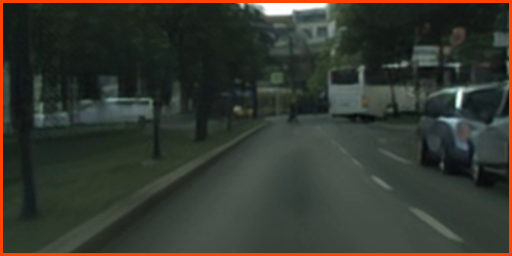} &
		\includegraphics[width=.230\linewidth]{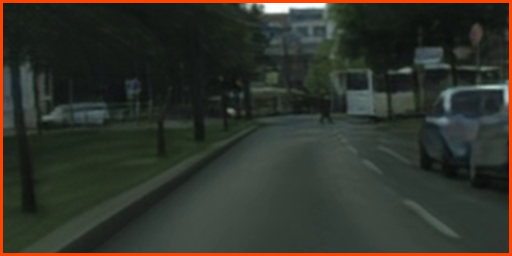} &
		\includegraphics[width=.230\linewidth]{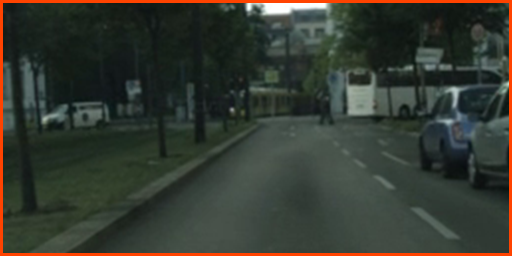} &
		\includegraphics[width=.230\linewidth]{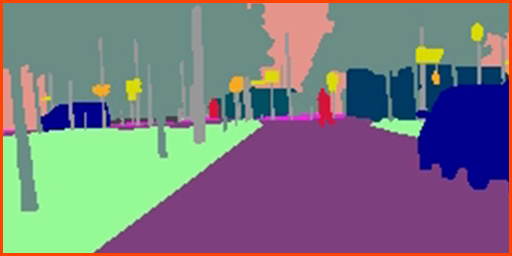} \\
		[0.3cm]
		\raisebox{1.75\normalbaselineskip}[0pt][0pt]{\rotatebox[origin=c]{90}{Full}} & 
		~ &
		\includegraphics[width=.230\linewidth]{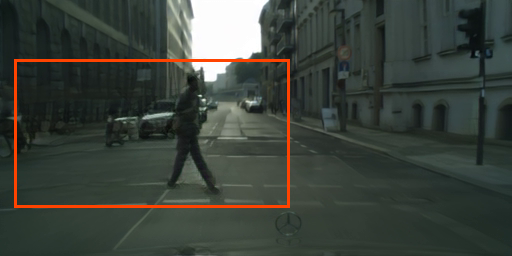} &
		\includegraphics[width=.230\linewidth]{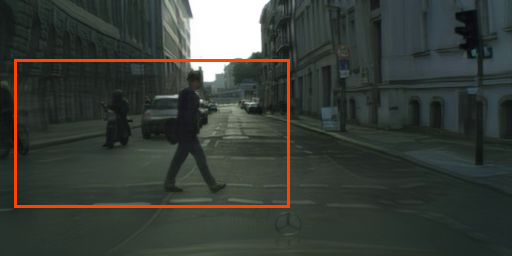} &
		\includegraphics[width=.230\linewidth]{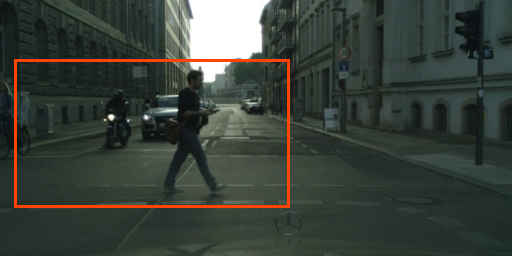} &
		\includegraphics[width=.230\linewidth]{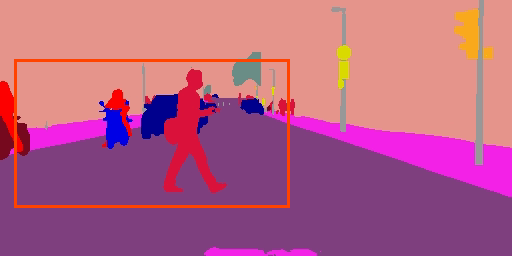} \\
		\raisebox{1.75\normalbaselineskip}[0pt][0pt]{\rotatebox[origin=c]{90}{Zoom}} & 
		~ &
		\includegraphics[width=.230\linewidth]{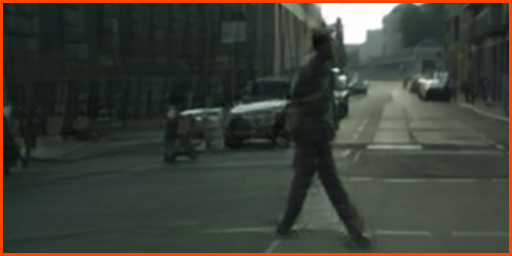} &
		\includegraphics[width=.230\linewidth]{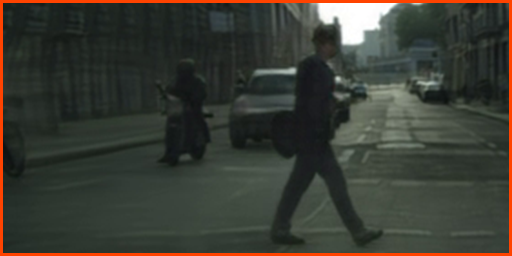} &
		\includegraphics[width=.230\linewidth]{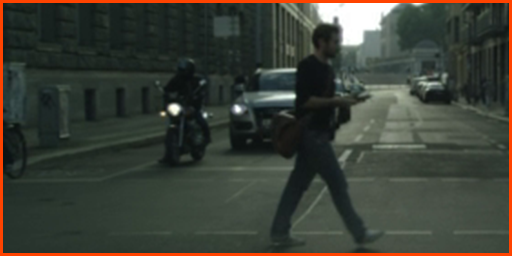} &
		\includegraphics[width=.230\linewidth]{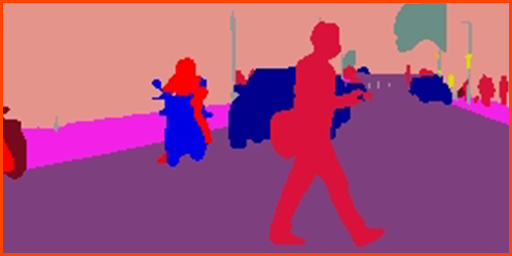} \\
		[0.3cm]
		\raisebox{1.75\normalbaselineskip}[0pt][0pt]{\rotatebox[origin=c]{90}{Full}} & 
		~ &
		\includegraphics[width=.230\linewidth]{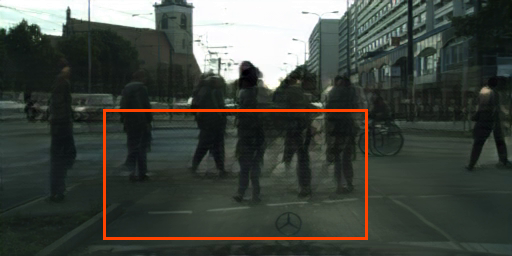} &
		\includegraphics[width=.230\linewidth]{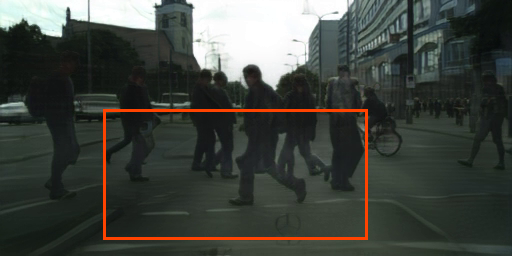} &
		\includegraphics[width=.230\linewidth]{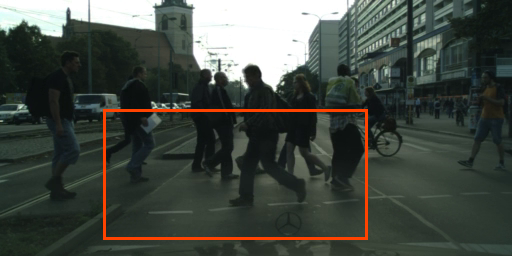} &
		\includegraphics[width=.230\linewidth]{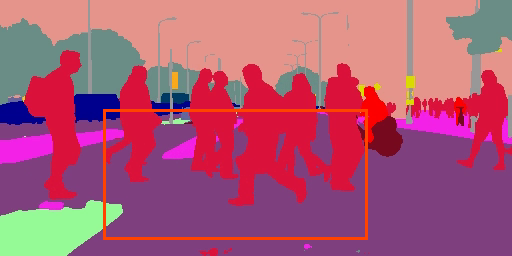} \\
		\raisebox{1.75\normalbaselineskip}[0pt][0pt]{\rotatebox[origin=c]{90}{Zoom}} & 
		~ &
		\includegraphics[width=.230\linewidth]{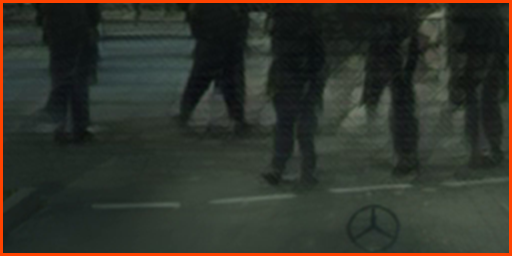} &
		\includegraphics[width=.230\linewidth]{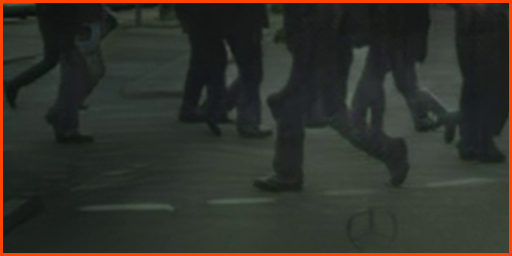} &
		\includegraphics[width=.230\linewidth]{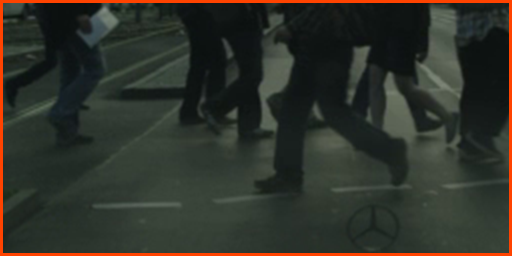} &
		\includegraphics[width=.230\linewidth]{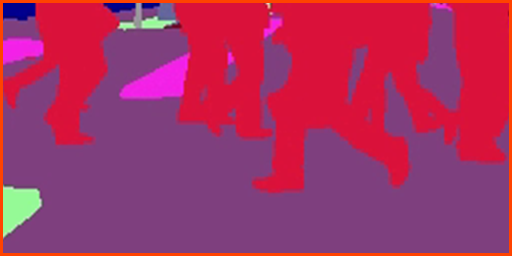} \\
		[0.3cm]
		\raisebox{1.75\normalbaselineskip}[0pt][0pt]{\rotatebox[origin=c]{90}{Full}} & 
		~ &
		\includegraphics[width=.230\linewidth]{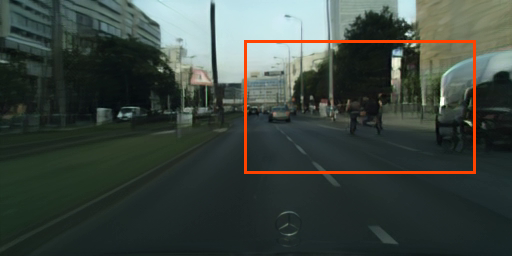} &
		\includegraphics[width=.230\linewidth]{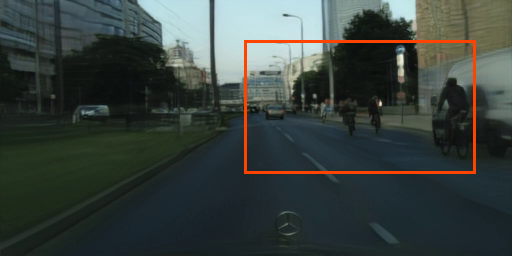} &
		\includegraphics[width=.230\linewidth]{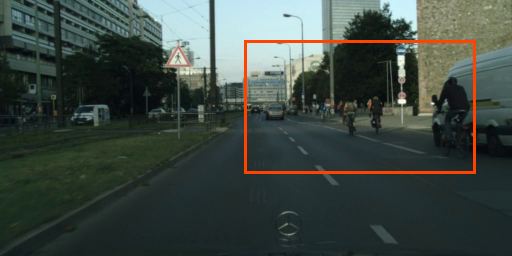} &
		\includegraphics[width=.230\linewidth]{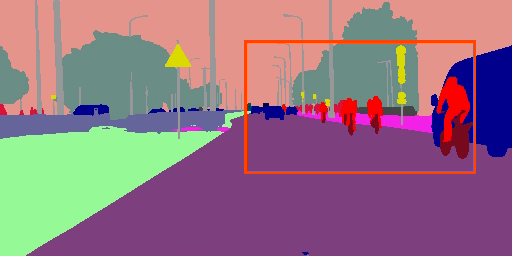} \\
		\raisebox{1.75\normalbaselineskip}[0pt][0pt]{\rotatebox[origin=c]{90}{Zoom}} & 
		~ &
		\includegraphics[width=.230\linewidth]{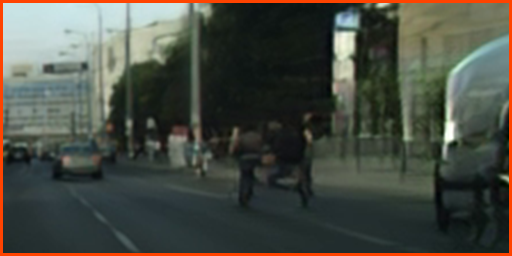} &
		\includegraphics[width=.230\linewidth]{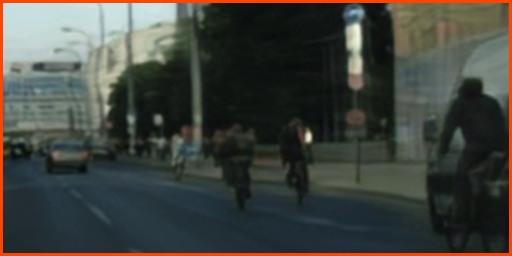} &
		\includegraphics[width=.230\linewidth]{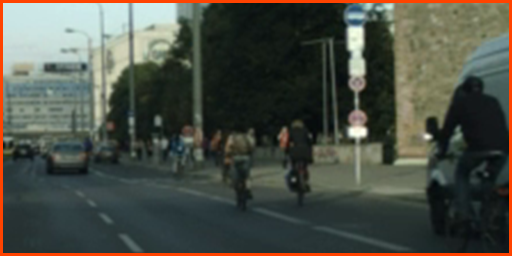} &
		\includegraphics[width=.230\linewidth]{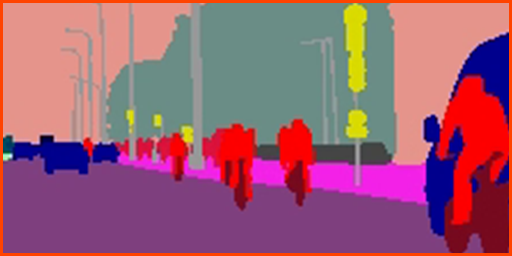} \\
		
	\end{tabular}
	\caption{\revision{Reconstruction of the 30\textsuperscript{th} frame given the real 1\textsuperscript{st} frame and both compressed features and layouts for subsequent timesteps. We compare our method with and without SPADE decoding blocks~\citep{park2019semantic} to the expected real image and layout at the same 30\textsuperscript{th} timestep. We zoom in specific areas to facilitate comparisons. We observe better alignment and more precise details for objects displaying complex motion when SPADE is used. }}
	\label{fig:city-layout-spade-256}
\end{figure}

\paragraph{Layout-conditioned synthesis.} We explore here layout-conditioned synthesis where one controls the semantic structure of new frames by attributing a class to each of their spatial locations. To this end, we apply \method{} on image and layout sequence pairs on the Cityscapes dataset~\citep{cordts2016cityscapes} which contains 3475 and 1525 30-frame sequences for train and test respectively. A single layout is annotated for each sequence, and we use a pretrained segmentation model~\citep{tao2020hierarchical} to obtain the remaining ones. Layouts are used as frame-level annotations for processing in $\transformer$. They are encoded, quantized, and decoded just as it is done for images. To save memory consumption during the decoding stage and ensure greater consistency between the two, frame and layout are decoded simultaneously with a single decoder $\decoder$ (we just supplement $\decoder$ with another prediction head to output layout masks). We show some qualitative results of \method{} on the synthesis of 30-frame videos given 3 starting frames and layouts for all timesteps in Figure~\ref{fig:city-layout-256}. Synthetic videos closely follow the target semantic structure, remain temporally consistent, and can handle complex motions. We repeat this synthesis process in Figure~\ref{fig:city-layout-off-256}, this time without layouts at subsequent timesteps. \method{} forecasts the semantic evolution of the scene on its own, predict corresponding layouts, and translate these into high-quality images. Both approaches (with or without layout at subsequent timesteps) are compared in Table~\ref{tab:city-layout-ablation-256}. Having this additional information allows to produce videos which are closer to the real ones. \revision{
A common architectural choice in layout-conditioned synthesis is to replace ResNet blocks~\citep{he2016deep} by SPADE blocks~\citep{park2019semantic} to strengthen the layout guidance in the decoding stage~\citep{mallya2020world}. Quantitatively, we see that it significantly improves the fidelity of the synthesized videos to the real ones in terms of SSIM and PSNR. Qualitatively, side-by-side reconstructions with and without SPADE in Figure~\ref{fig:city-layout-spade-256} show that the flow module allows temporally consistent textures (similar to the 3D world model used in~\citep{mallya2020world} but without being limited to static objects here) while SPADE enhances the compliance to the semantic structure, especially for small fast-moving objects.}

\begin{table}
    \setlength\heavyrulewidth{0.25ex}
	\aboverulesep=0ex
    \belowrulesep=0.3ex
	\centering
	\caption{Layout-conditioned video synthesis on Cityscapes ($256 \stimes 512$).}
	\footnotesize
	\begin{tabular}{@{}L{2.51cm}@{}C{0.95cm}@{}C{1.00cm}@{}C{1.70cm}@{}C{1.70cm}@{}C{1.70cm}@{}C{1.36cm}@{}C{1.36cm}@{}C{1.36cm}@{}}
    	\toprule
        \multirow{2}{*}{Method} & \multirow{2}{*}{Cond.} & \multirow{2}{*}{Layout} & \multicolumn{3}{@{}c@{}}{SSIM \hspace{-1.0ex} $\uparrow$} & \multicolumn{3}{@{}c@{}}{PSNR \hspace{-1.0ex} $\uparrow$}\\
        \cmidrule(r){4-6} \cmidrule{7-9}
        & & & {\scriptsize$t=10$} & {\scriptsize$t=20$} & {\scriptsize$t=30$} & {\scriptsize$t=10$} & {\scriptsize$t=20$} & {\scriptsize$t=30$} \\
        \midrule
        \method{} (\emph{ours}) & 3 &  & $0.750_{\pm0.002}$ & $0.639_{\pm0.003}$ & $0.582_{\pm0.005}$ & $20.7_{\pm0.1}$ & $18.8_{\pm0.1}$ & $17.9_{\pm0.1}$ \\
        \method{} (\emph{ours})& 3 & \checkmark & $\mathbf{0.783}_{\pm0.002}$ & ${0.690}_{\pm0.003}$ & ${0.640}_{\pm0.005}$ & ${21.5}_{\pm0.1}$ & ${19.6}_{\pm0.1}$ & ${18.7}_{\pm0.1}$ \\
        \method{} + \citep{park2019semantic} (\emph{ours})& 3 & \checkmark & ${0.776}_{\pm0.002}$ & $\mathbf{0.742}_{\pm0.002}$ & $\mathbf{0.723}_{\pm0.005}$ & $\mathbf{22.1}_{\pm0.1}$ & $\mathbf{20.6}_{\pm0.1}$ & $\mathbf{19.9}_{\pm0.1}$ \\
        \bottomrule
    \end{tabular}
	\label{tab:city-layout-ablation-256}
\end{table}

\paragraph{Sound ablation study.} A simple ablation of \method{} on AudioSet Drum~\citep{gemmeke2017audio} is presented in Table~\ref{tab:drum-sound-ablation-64}. It shows that, as time goes by, the performance gap between unimodal and multimodal synthesis (when considering audio) increases. However, the improvement on account of the audio information seems relatively small compared to the overall performance. We suspect this is due to a minor part of the scene being animated (the drummer upper body) and to relatively fast motion (it is difficult to strictly minimize the image-to-image error between real and synthetic videos in this case).

\begin{table}
    \setlength\heavyrulewidth{0.25ex}
	\aboverulesep=0ex
    \belowrulesep=0.3ex
	\centering
	\caption{Sound-conditioned video synthesis on AudioSet-Drums ($64 \stimes 64$).}
	\footnotesize
	\begin{tabular}{@{}L{2.51cm}@{}C{0.95cm}@{}C{1.00cm}@{}C{1.70cm}@{}C{1.70cm}@{}C{1.70cm}@{}C{1.36cm}@{}C{1.36cm}@{}C{1.36cm}@{}}
    	\toprule
        \multirow{2}{*}{Method} & \multirow{2}{*}{Cond.} & \multirow{2}{*}{Audio} & \multicolumn{3}{@{}c@{}}{SSIM \hspace{-1.0ex} $\uparrow$} & \multicolumn{3}{@{}c@{}}{PSNR \hspace{-1.0ex} $\uparrow$}\\
        \cmidrule(r){4-6} \cmidrule{7-9}
        & & & {\scriptsize$t=16$} & {\scriptsize$t=30$} & {\scriptsize$t=45$} & {\scriptsize$t=16$} & {\scriptsize$t=30$} & {\scriptsize$t=45$} \\
        \midrule
        \method{} (\emph{ours}) & 1 &  & $0.952_{\pm0.005}$ & $0.942_{\pm0.009}$ & $0.925_{\pm0.002}$ & $28.1_{\pm0.5}$ & $26.8_{\pm0.5}$ & $25.8_{\pm0.6}$ \\
        \method{} (\emph{ours})& 1 & \checkmark & $\mathbf{0.955}_{\pm0.006}$ & $\mathbf{0.943}_{\pm0.008}$ & $\mathbf{0.931}_{\pm0.010}$ & $\mathbf{28.2}_{\pm0.5}$ & $\mathbf{27.1}_{\pm0.4}$ & $\mathbf{26.3}_{\pm0.4}$ \\
        \bottomrule
    \end{tabular}
	\label{tab:drum-sound-ablation-64}
\end{table}

\begin{figure}
	\setlength\tabcolsep{2.0pt}
	\renewcommand{\arraystretch}{1.0}
	\footnotesize
	\begin{tabular}{cccccccc}
	    & $t=1$ & $t=5$ & $t=10$ & $t=15$ & $t=20$ & $t=25$ & $t=30$ \\
	    [0.05cm]
		
		\raisebox{2.3\normalbaselineskip}[0pt][0pt]{\rotatebox[origin=c]{90}{$L=20$}} &
		\includegraphics[width=.125\linewidth]{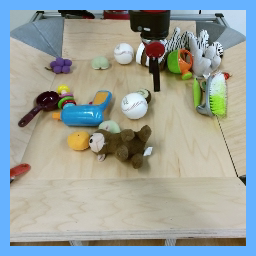} &
		\includegraphics[width=.125\linewidth]{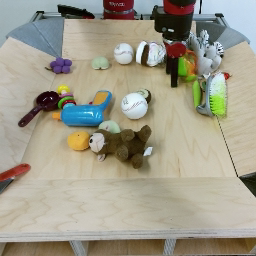} &
		\includegraphics[width=.125\linewidth]{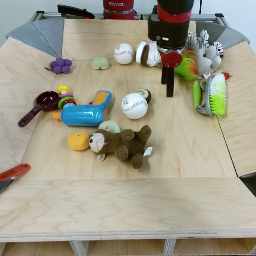} &
		\includegraphics[width=.125\linewidth]{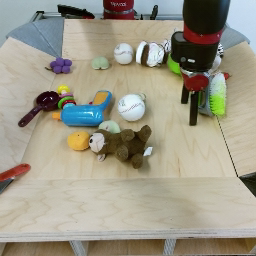} &
		\includegraphics[width=.125\linewidth]{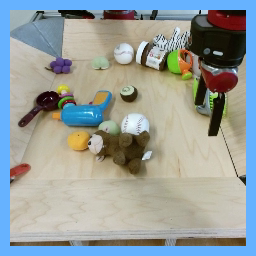} &
		&
		\\
		
		\raisebox{2.3\normalbaselineskip}[0pt][0pt]{\rotatebox[origin=c]{90}{$L=25$}} &
		\includegraphics[width=.125\linewidth]{neurips2021/figures/sup/p2p/30/vid_00006_1.png} &
		\includegraphics[width=.125\linewidth]{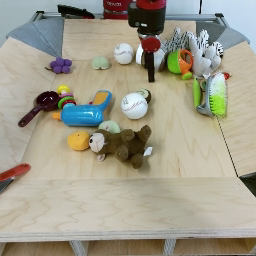} &
		\includegraphics[width=.125\linewidth]{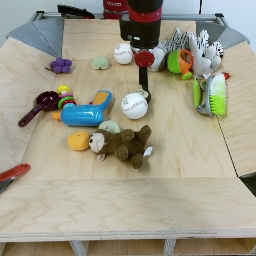} &
		\includegraphics[width=.125\linewidth]{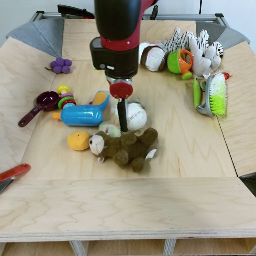} &
		\includegraphics[width=.125\linewidth]{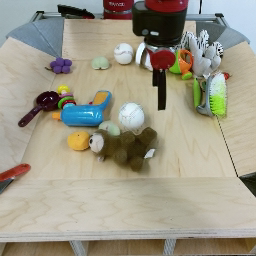} &
		\includegraphics[width=.125\linewidth]{neurips2021/figures/sup/p2p/30/vid_00006_30.png} &
		\\
		
		\raisebox{2.3\normalbaselineskip}[0pt][0pt]{\rotatebox[origin=c]{90}{$L=30$}} &
		\includegraphics[width=.125\linewidth]{neurips2021/figures/sup/p2p/30/vid_00006_1.png} &
		\includegraphics[width=.125\linewidth]{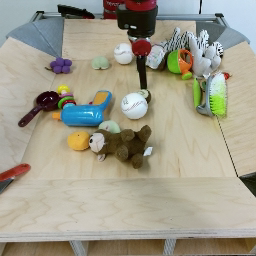} &
		\includegraphics[width=.125\linewidth]{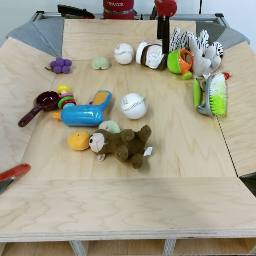} &
		\includegraphics[width=.125\linewidth]{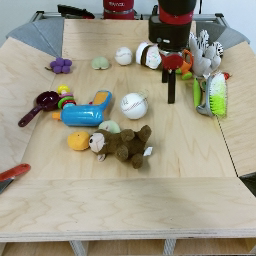} &
		\includegraphics[width=.125\linewidth]{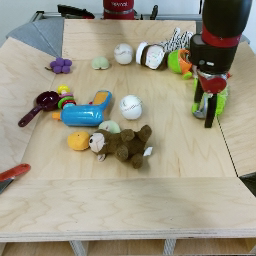} &
		\includegraphics[width=.125\linewidth]{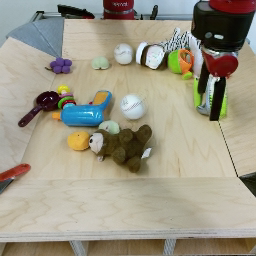} &
		\includegraphics[width=.125\linewidth]{neurips2021/figures/sup/p2p/30/vid_00006_30.png} \\
		
		\\
		
		\raisebox{2.3\normalbaselineskip}[0pt][0pt]{\rotatebox[origin=c]{90}{$L=20$}} &
		\includegraphics[width=.125\linewidth]{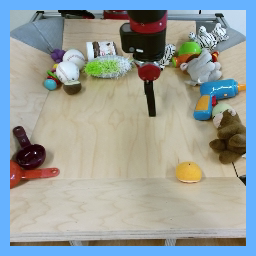} &
		\includegraphics[width=.125\linewidth]{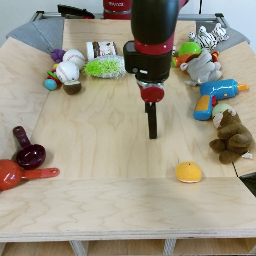} &
		\includegraphics[width=.125\linewidth]{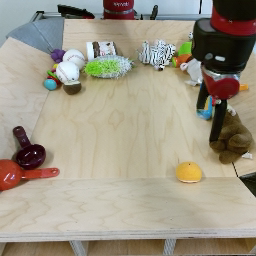} &
		\includegraphics[width=.125\linewidth]{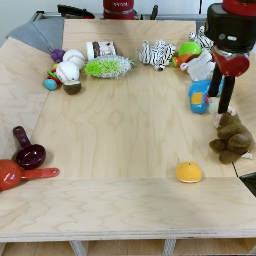} &
		\includegraphics[width=.125\linewidth]{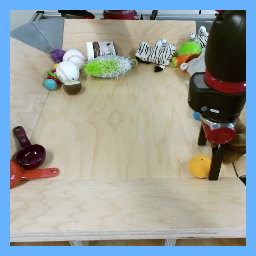} &
		&
		\\
		
		\raisebox{2.3\normalbaselineskip}[0pt][0pt]{\rotatebox[origin=c]{90}{$L=25$}} &
		\includegraphics[width=.125\linewidth]{neurips2021/figures/sup/p2p/30/vid_00034_1.png} &
		\includegraphics[width=.125\linewidth]{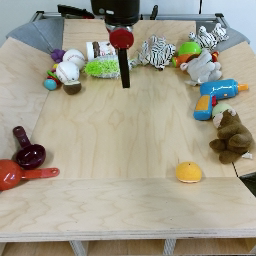} &
		\includegraphics[width=.125\linewidth]{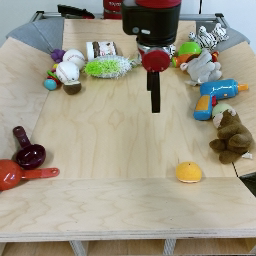} &
		\includegraphics[width=.125\linewidth]{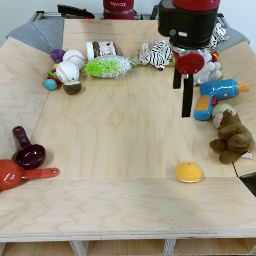} &
		\includegraphics[width=.125\linewidth]{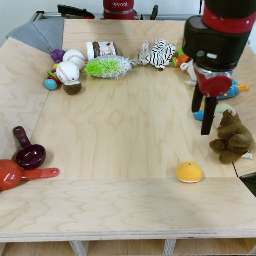} &
		\includegraphics[width=.125\linewidth]{neurips2021/figures/sup/p2p/30/vid_00034_30.png} &
		\\
		
		\raisebox{2.3\normalbaselineskip}[0pt][0pt]{\rotatebox[origin=c]{90}{$L=30$}} &
		\includegraphics[width=.125\linewidth]{neurips2021/figures/sup/p2p/30/vid_00034_1.png} &
		\includegraphics[width=.125\linewidth]{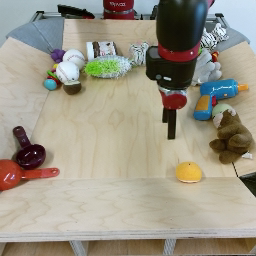} &
		\includegraphics[width=.125\linewidth]{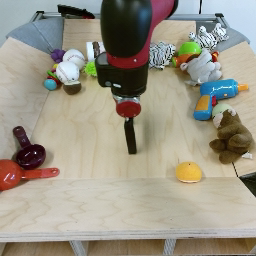} &
		\includegraphics[width=.125\linewidth]{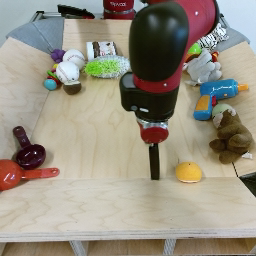} &
		\includegraphics[width=.125\linewidth]{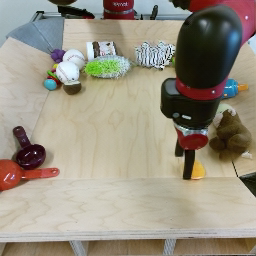} &
		\includegraphics[width=.125\linewidth]{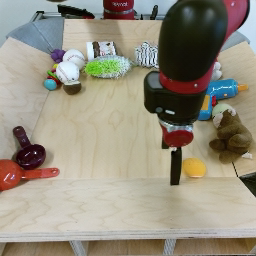} &
		\includegraphics[width=.125\linewidth]{neurips2021/figures/sup/p2p/30/vid_00034_30.png} \\
		
		\\
		
		\raisebox{2.3\normalbaselineskip}[0pt][0pt]{\rotatebox[origin=c]{90}{$L=20$}} &
		\includegraphics[width=.125\linewidth]{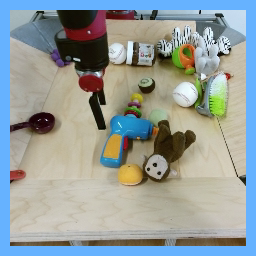} &
		\includegraphics[width=.125\linewidth]{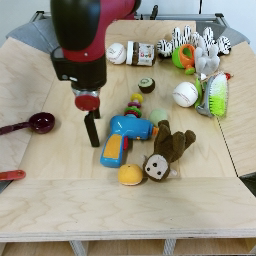} &
		\includegraphics[width=.125\linewidth]{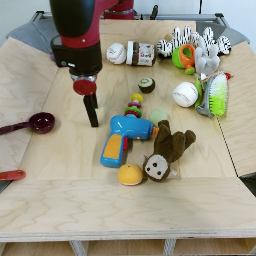} &
		\includegraphics[width=.125\linewidth]{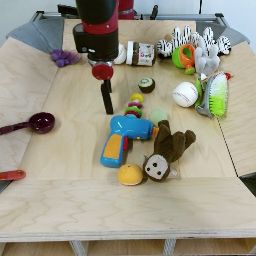} &
		\includegraphics[width=.125\linewidth]{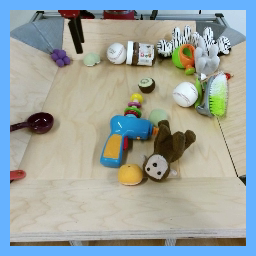} &
		&
		\\
		
		\raisebox{2.3\normalbaselineskip}[0pt][0pt]{\rotatebox[origin=c]{90}{$L=25$}} &
		\includegraphics[width=.125\linewidth]{neurips2021/figures/sup/p2p/30/vid_00018_1.png} &
		\includegraphics[width=.125\linewidth]{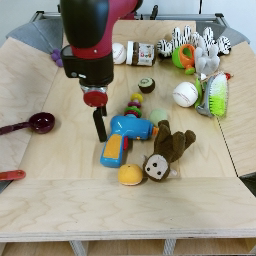} &
		\includegraphics[width=.125\linewidth]{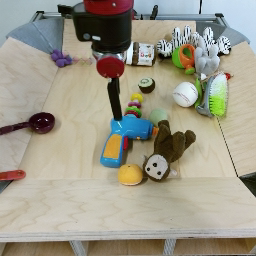} &
		\includegraphics[width=.125\linewidth]{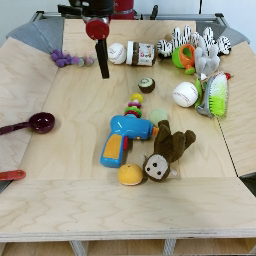} &
		\includegraphics[width=.125\linewidth]{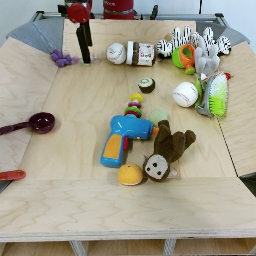} &
		\includegraphics[width=.125\linewidth]{neurips2021/figures/sup/p2p/30/vid_00018_30.png} &
		\\
		
		\raisebox{2.3\normalbaselineskip}[0pt][0pt]{\rotatebox[origin=c]{90}{$L=30$}} &
		\includegraphics[width=.125\linewidth]{neurips2021/figures/sup/p2p/30/vid_00018_1.png} &
		\includegraphics[width=.125\linewidth]{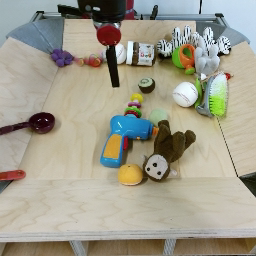} &
		\includegraphics[width=.125\linewidth]{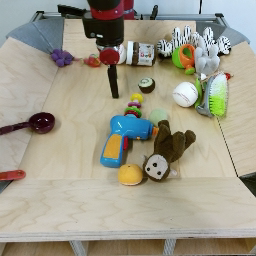} &
		\includegraphics[width=.125\linewidth]{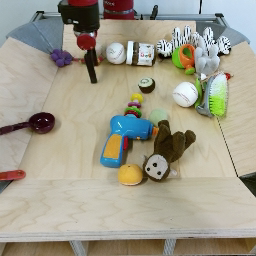} &
		\includegraphics[width=.125\linewidth]{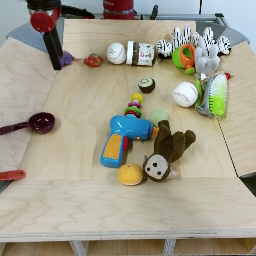} &
		\includegraphics[width=.125\linewidth]{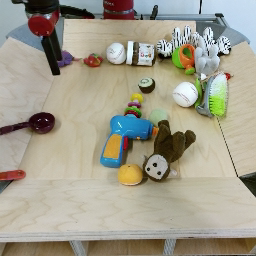} &
		\includegraphics[width=.125\linewidth]{neurips2021/figures/sup/p2p/30/vid_00018_30.png} \\
		
	\end{tabular}
	\caption{Videos of various lengths ($L \in \{20, 25, 30\}$) synthesized with our point-to-point method given a pair of start and end frames on BAIR ($256 \times 256$). Note that the same model is used to synthesize videos of different lengths. Our approach generalizes to arbitrary lengths between start and end frames by taking the length into account in the positional embedding of the end frame.}
	\label{fig:bair-p2p-length-256}
\end{figure}

\paragraph{Point-to-point synthesis.} We explore further the performance of \method{} on the point-to-point synthesis task which we introduced in Section~\ref{sec:experiments} of the article. We show some quantitative results against different baselines in Table~\ref{tab:bair-p2p-64}. \method{} is on par or significantly outperforms the state of the art on $3$ out of $4$ metrics. It produces more faithful reconstructions when latent features are known. \method{} is more likely to synthesize the true outcome among $100$ predicted samples compared to baselines. It has a good control point consistency (\ie, the last synthetic frame is close to the corresponding ground-truth), but lower diversity than prior arts. We note, however, that SSIM diversity may capture the capacity of producing different trajectories between start and end points, but may also increase due to visual quality degradations. Like previous works~\citep{wang2019point}, \method{} is able to produce videos of arbitrary length between the target start and end frames. We recall that the end frame is used as a video-level annotation in the transformer model $\transformer$ (Figure~\ref{fig:gpt}) so that it is taken into account when predicting frames at earlier timesteps. To allow any timestep for the end frame (possibly greater than the capacity of $\transformer$), we make its positional embedding reflect the gap with current predicted frames. Namely, the positional embedding of the end frame is updated as the sliding window moves forward in time and gets closer to the end timestep. We show qualitative results for different video lengths in Figure~\ref{fig:bair-p2p-length-256}.

\begin{table}
    \setlength\heavyrulewidth{0.25ex}
	\aboverulesep=0ex
    \belowrulesep=0.3ex
	\centering
	\caption{Point-to-point video synthesis on BAIR ($64 \stimes 64$). We follow~\citep{wang2019point} and synthesize 30-frame videos conditioned on the start and end frames. For each of the real test videos we produce synthetic ones (100 predicted samples) and compute the pairwise SSIM between real and synthetic frames. As a quality assessment, we check if the real video is represented among synthetic ones by reporting the best SSIM among samples (``Best''). We estimate the diversity of the predictions by computing the variance of the SSIM across samples with the true video as reference (``Div.''). We evaluate the control point consistency by measuring the SSIM between the last synthesized frame and the corresponding real one for all samples (``C.P.C.''). Finally, we assess the reconstruction quality for the different methods by using latent features extracted from real frames at intermediate timesteps instead of predicting them (``Rec.''). Metrics are computed in a $95\%$ confidence interval like in~\citep{wang2019point}.}
	\footnotesize
	\begin{tabular}{@{}L{2.00cm}@{}C{1.20cm}@{}C{2.00cm}@{}C{2.00cm}@{}C{2.00cm}@{}C{2.00cm}@{}}
    	\toprule
        \multirow{2}{*}{Method} & \multirow{2}{*}{Cond.} & \multicolumn{4}{@{}c@{}}{SSIM \hspace{-1.0ex} $\uparrow$}\\
        \cmidrule(){3-6}
        & & {\scriptsize Best} & {\scriptsize Div. (1E-3)} & {\scriptsize C.P.C.} & {\scriptsize Rec.} \\
        \midrule
        SVG~\citep{denton2018stochastic} & 2 & $0.845_{\pm0.006}$ & $\mathbf{0.716}_{\pm0.166}$ & $0.775_{\pm0.008}$ & $0.926_{\pm0.003}$ \\
        SV2P~\citep{babaeizadeh2018stochastic} & 2 & $0.841_{\pm0.010}$ & $0.186_{\pm0.021}$ & $0.770_{\pm0.009}$ & $0.847_{\pm0.004}$ \\
        P2PVG~\citep{wang2019point} & 2 & $0.847_{\pm0.004}$ & $0.664_{\pm0.049}$ & $\mathbf{0.824}_{\pm0.015}$ & $0.907_{\pm0.006}$ \\
        \method{} (\emph{ours}) & 2 & $\mathbf{0.857}_{\pm0.006}$ & $0.454_{\pm0.041}$ & $\mathbf{0.824}_{\pm0.007}$ & $\mathbf{0.960}_{\pm0.005}$\\
        \bottomrule
    \end{tabular}
	\label{tab:bair-p2p-64}
\end{table}

\begin{figure}
	\setlength\tabcolsep{2.0pt}
	\renewcommand{\arraystretch}{1.0}
	\footnotesize
	\begin{tabular}{cccccccccc}
	    & $t=1$ & $t=8$ & $t=16$ & $t=24$ & $t=32$ & $t=40$ & $t=48$ & $t=56$ & $t=64$ \\
	    [0.05cm]
		
		\raisebox{1.7\normalbaselineskip}[0pt][0pt]{\rotatebox[origin=c]{90}{walk}}
		&\includegraphics[width=.095\linewidth]{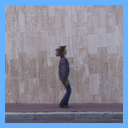} 
		&\includegraphics[width=.095\linewidth]{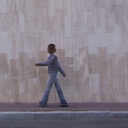}
		&\includegraphics[width=.095\linewidth]{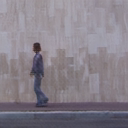} &\includegraphics[width=.095\linewidth]{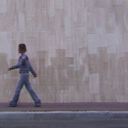} &\includegraphics[width=.095\linewidth]{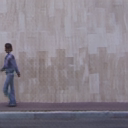} &\includegraphics[width=.095\linewidth]{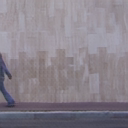} &\includegraphics[width=.095\linewidth]{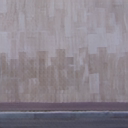} &\includegraphics[width=.095\linewidth]{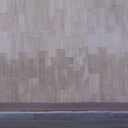} 
		&\includegraphics[width=.095\linewidth]{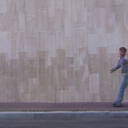} \\
		
		\raisebox{1.7\normalbaselineskip}[0pt][0pt]{\rotatebox[origin=c]{90}{run}}
		&\includegraphics[width=.095\linewidth]{neurips2021/figures/sup/weiz/from_img/side.png} 
		&\includegraphics[width=.095\linewidth]{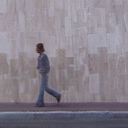}
		&\includegraphics[width=.095\linewidth]{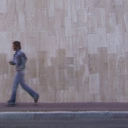} &\includegraphics[width=.095\linewidth]{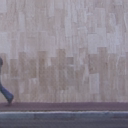} &\includegraphics[width=.095\linewidth]{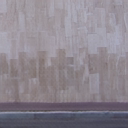} &\includegraphics[width=.095\linewidth]{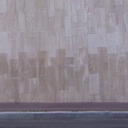} &\includegraphics[width=.095\linewidth]{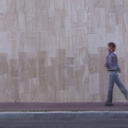} &\includegraphics[width=.095\linewidth]{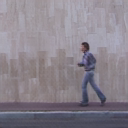} 
		&\includegraphics[width=.095\linewidth]{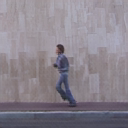} \\
		
		\raisebox{1.7\normalbaselineskip}[0pt][0pt]{\rotatebox[origin=c]{90}{jump}}
		&\includegraphics[width=.095\linewidth]{neurips2021/figures/sup/weiz/from_img/side.png} 
		&\includegraphics[width=.095\linewidth]{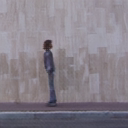}
		&\includegraphics[width=.095\linewidth]{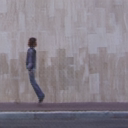} &\includegraphics[width=.095\linewidth]{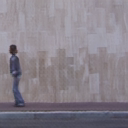} &\includegraphics[width=.095\linewidth]{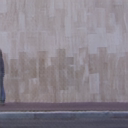} &\includegraphics[width=.095\linewidth]{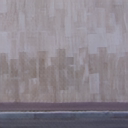} &\includegraphics[width=.095\linewidth]{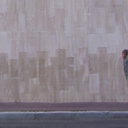} &\includegraphics[width=.095\linewidth]{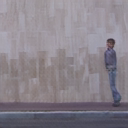} 
		&\includegraphics[width=.095\linewidth]{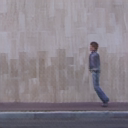} \\
		
		\raisebox{1.7\normalbaselineskip}[0pt][0pt]{\rotatebox[origin=c]{90}{bend}}
		&\includegraphics[width=.095\linewidth]{neurips2021/figures/sup/weiz/from_img/side.png} 
		&\includegraphics[width=.095\linewidth]{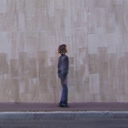}
		&\includegraphics[width=.095\linewidth]{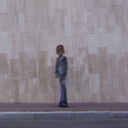} &\includegraphics[width=.095\linewidth]{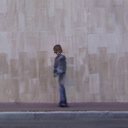} &\includegraphics[width=.095\linewidth]{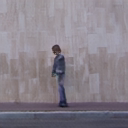} &\includegraphics[width=.095\linewidth]{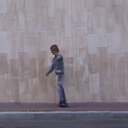} &\includegraphics[width=.095\linewidth]{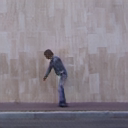} &\includegraphics[width=.095\linewidth]{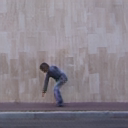} 
		&\includegraphics[width=.095\linewidth]{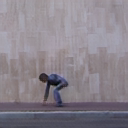} \\
		
		\raisebox{1.7\normalbaselineskip}[0pt][0pt]{\rotatebox[origin=c]{90}{skip}}
		&\includegraphics[width=.095\linewidth]{neurips2021/figures/sup/weiz/from_img/side.png} 
		&\includegraphics[width=.095\linewidth]{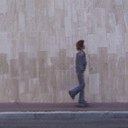}
		&\includegraphics[width=.095\linewidth]{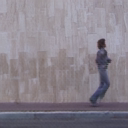} &\includegraphics[width=.095\linewidth]{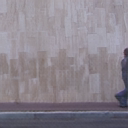} &\includegraphics[width=.095\linewidth]{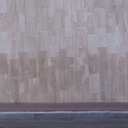} &\includegraphics[width=.095\linewidth]{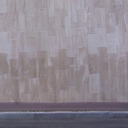} &\includegraphics[width=.095\linewidth]{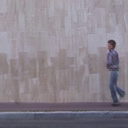} &\includegraphics[width=.095\linewidth]{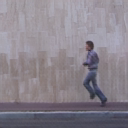} 
		&\includegraphics[width=.095\linewidth]{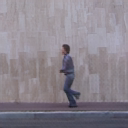} \\
		
		\raisebox{1.7\normalbaselineskip}[0pt][0pt]{\rotatebox[origin=c]{90}{side}}
		&\includegraphics[width=.095\linewidth]{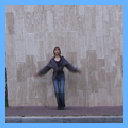} 
		&\includegraphics[width=.095\linewidth]{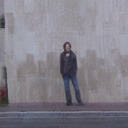}
		&\includegraphics[width=.095\linewidth]{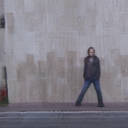} &\includegraphics[width=.095\linewidth]{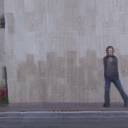} &\includegraphics[width=.095\linewidth]{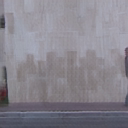} &\includegraphics[width=.095\linewidth]{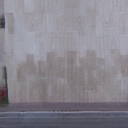} &\includegraphics[width=.095\linewidth]{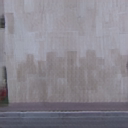} &\includegraphics[width=.095\linewidth]{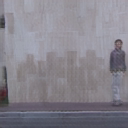} 
		&\includegraphics[width=.095\linewidth]{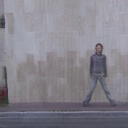} \\
		
		\raisebox{1.7\normalbaselineskip}[0pt][0pt]{\rotatebox[origin=c]{90}{wave1}}
		&\includegraphics[width=.095\linewidth]{neurips2021/figures/sup/weiz/from_img/front.png} 
		&\includegraphics[width=.095\linewidth]{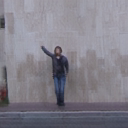}
		&\includegraphics[width=.095\linewidth]{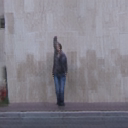} &\includegraphics[width=.095\linewidth]{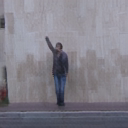} &\includegraphics[width=.095\linewidth]{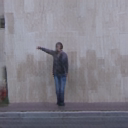} &\includegraphics[width=.095\linewidth]{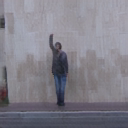} &\includegraphics[width=.095\linewidth]{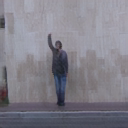} &\includegraphics[width=.095\linewidth]{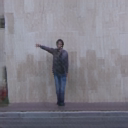} 
		&\includegraphics[width=.095\linewidth]{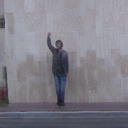} \\
		
		\raisebox{1.7\normalbaselineskip}[0pt][0pt]{\rotatebox[origin=c]{90}{wave2}}
		&\includegraphics[width=.095\linewidth]{neurips2021/figures/sup/weiz/from_img/front.png} 
		&\includegraphics[width=.095\linewidth]{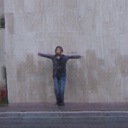}
		&\includegraphics[width=.095\linewidth]{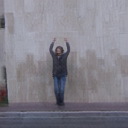} &\includegraphics[width=.095\linewidth]{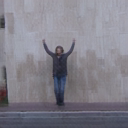} &\includegraphics[width=.095\linewidth]{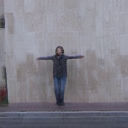} &\includegraphics[width=.095\linewidth]{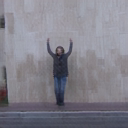} &\includegraphics[width=.095\linewidth]{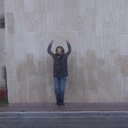} &\includegraphics[width=.095\linewidth]{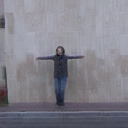} 
		&\includegraphics[width=.095\linewidth]{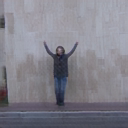} \\
		
		\raisebox{1.7\normalbaselineskip}[0pt][0pt]{\rotatebox[origin=c]{90}{pjump}}
		&\includegraphics[width=.095\linewidth]{neurips2021/figures/sup/weiz/from_img/front.png} 
		&\includegraphics[width=.095\linewidth]{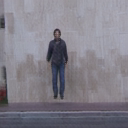}
		&\includegraphics[width=.095\linewidth]{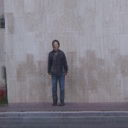} &\includegraphics[width=.095\linewidth]{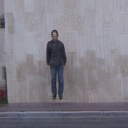} &\includegraphics[width=.095\linewidth]{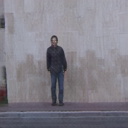} &\includegraphics[width=.095\linewidth]{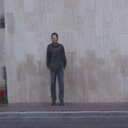} &\includegraphics[width=.095\linewidth]{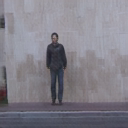} &\includegraphics[width=.095\linewidth]{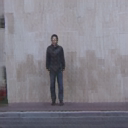} 
		&\includegraphics[width=.095\linewidth]{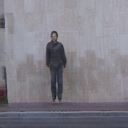} \\
		
		\raisebox{1.7\normalbaselineskip}[0pt][0pt]{\rotatebox[origin=c]{90}{jack}}
		&\includegraphics[width=.095\linewidth]{neurips2021/figures/sup/weiz/from_img/front.png} 
		&\includegraphics[width=.095\linewidth]{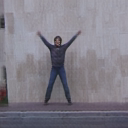}
		&\includegraphics[width=.095\linewidth]{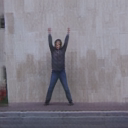} &\includegraphics[width=.095\linewidth]{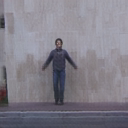} &\includegraphics[width=.095\linewidth]{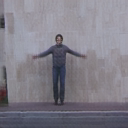} &\includegraphics[width=.095\linewidth]{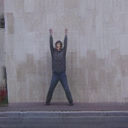} &\includegraphics[width=.095\linewidth]{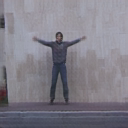} &\includegraphics[width=.095\linewidth]{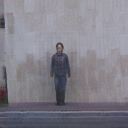} 
		&\includegraphics[width=.095\linewidth]{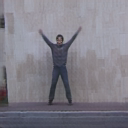} \\

	\end{tabular}
	\caption{Qualitative samples of class-conditioned synthesis from a single image on Weizmann ($128 \stimes 128$). We use one conditioning frame from a held-out subject (not seen during training), that is, a side view for lateral actions (``walk'', ``run'', ``jump'', ``bend'' and ``skip'') and a front view for frontal actions (``side'', ``wave1'', ``wave2'', ``pjump'' and ``jack'').}
	\label{fig:weizmann-class-128}
\end{figure}

\begin{figure}
	\setlength\tabcolsep{2.0pt}
	\renewcommand{\arraystretch}{1.0}
	\footnotesize
	\begin{tabular}{cccccccccc}
	    & $t=1$ & $t=8$ & $t=16$ & $t=24$ & $t=32$ & $t=40$ & $t=48$ & $t=56$ & $t=64$ \\
	    [0.05cm]
		
		\raisebox{1.7\normalbaselineskip}[0pt][0pt]{\rotatebox[origin=c]{90}{walk}}
		&\includegraphics[width=.095\linewidth]{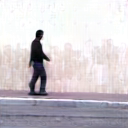} 
		&\includegraphics[width=.095\linewidth]{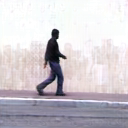}
		&\includegraphics[width=.095\linewidth]{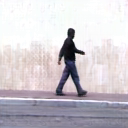} &\includegraphics[width=.095\linewidth]{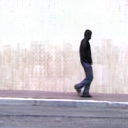} &\includegraphics[width=.095\linewidth]{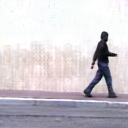} &\includegraphics[width=.095\linewidth]{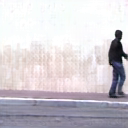} &\includegraphics[width=.095\linewidth]{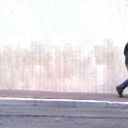} &\includegraphics[width=.095\linewidth]{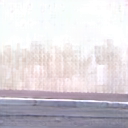} 
		&\includegraphics[width=.095\linewidth]{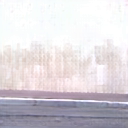} \\
		
		\raisebox{1.7\normalbaselineskip}[0pt][0pt]{\rotatebox[origin=c]{90}{run}}
		&\includegraphics[width=.095\linewidth]{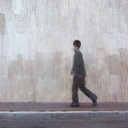}
		&\includegraphics[width=.095\linewidth]{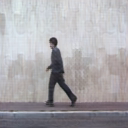}
		&\includegraphics[width=.095\linewidth]{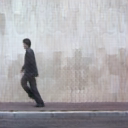} &\includegraphics[width=.095\linewidth]{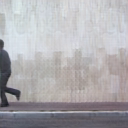} &\includegraphics[width=.095\linewidth]{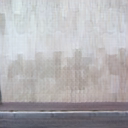} &\includegraphics[width=.095\linewidth]{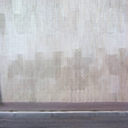} &\includegraphics[width=.095\linewidth]{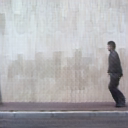} &\includegraphics[width=.095\linewidth]{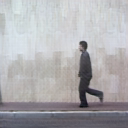} 
		&\includegraphics[width=.095\linewidth]{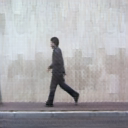} \\
		
		\raisebox{1.7\normalbaselineskip}[0pt][0pt]{\rotatebox[origin=c]{90}{jump}}
		&\includegraphics[width=.095\linewidth]{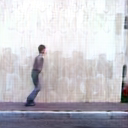}
		&\includegraphics[width=.095\linewidth]{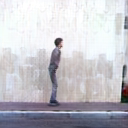}
		&\includegraphics[width=.095\linewidth]{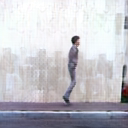} &\includegraphics[width=.095\linewidth]{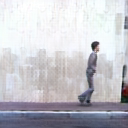} &\includegraphics[width=.095\linewidth]{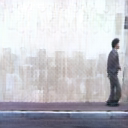} &\includegraphics[width=.095\linewidth]{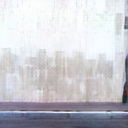} &\includegraphics[width=.095\linewidth]{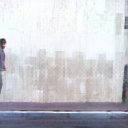} &\includegraphics[width=.095\linewidth]{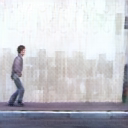} 
		&\includegraphics[width=.095\linewidth]{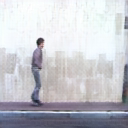} \\
		
		\raisebox{1.7\normalbaselineskip}[0pt][0pt]{\rotatebox[origin=c]{90}{bend}}
		&\includegraphics[width=.095\linewidth]{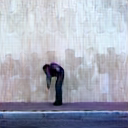}
		&\includegraphics[width=.095\linewidth]{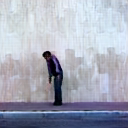}
		&\includegraphics[width=.095\linewidth]{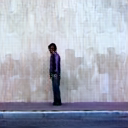} &\includegraphics[width=.095\linewidth]{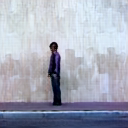} &\includegraphics[width=.095\linewidth]{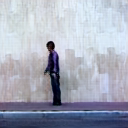} &\includegraphics[width=.095\linewidth]{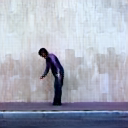} &\includegraphics[width=.095\linewidth]{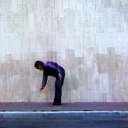} &\includegraphics[width=.095\linewidth]{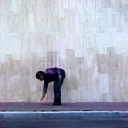} 
		&\includegraphics[width=.095\linewidth]{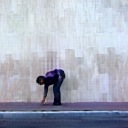} \\
		
		\raisebox{1.7\normalbaselineskip}[0pt][0pt]{\rotatebox[origin=c]{90}{skip}}
		&\includegraphics[width=.095\linewidth]{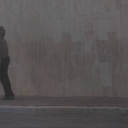}
		&\includegraphics[width=.095\linewidth]{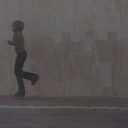}
		&\includegraphics[width=.095\linewidth]{neurips2021/figures/sup/weiz/without_img/more/vid_00000_16.png} &\includegraphics[width=.095\linewidth]{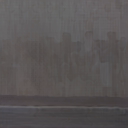} &\includegraphics[width=.095\linewidth]{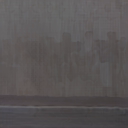} &\includegraphics[width=.095\linewidth]{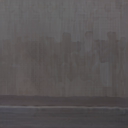} &\includegraphics[width=.095\linewidth]{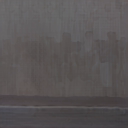} &\includegraphics[width=.095\linewidth]{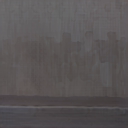} 
		&\includegraphics[width=.095\linewidth]{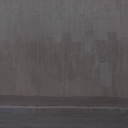} \\
		
		\raisebox{1.7\normalbaselineskip}[0pt][0pt]{\rotatebox[origin=c]{90}{side}}
		&\includegraphics[width=.095\linewidth]{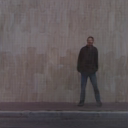}
		&\includegraphics[width=.095\linewidth]{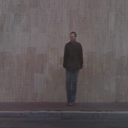}
		&\includegraphics[width=.095\linewidth]{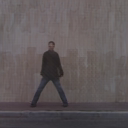} &\includegraphics[width=.095\linewidth]{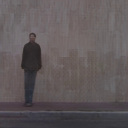} &\includegraphics[width=.095\linewidth]{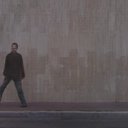} &\includegraphics[width=.095\linewidth]{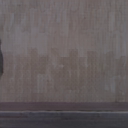} &\includegraphics[width=.095\linewidth]{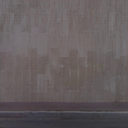} &\includegraphics[width=.095\linewidth]{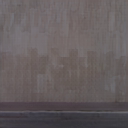} 
		&\includegraphics[width=.095\linewidth]{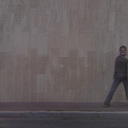} \\
		
		\raisebox{1.7\normalbaselineskip}[0pt][0pt]{\rotatebox[origin=c]{90}{wave1}}
		&\includegraphics[width=.095\linewidth]{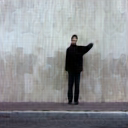}
		&\includegraphics[width=.095\linewidth]{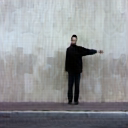}
		&\includegraphics[width=.095\linewidth]{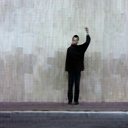} &\includegraphics[width=.095\linewidth]{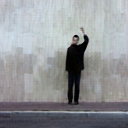} &\includegraphics[width=.095\linewidth]{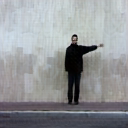} &\includegraphics[width=.095\linewidth]{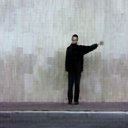} &\includegraphics[width=.095\linewidth]{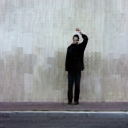} &\includegraphics[width=.095\linewidth]{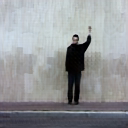} 
		&\includegraphics[width=.095\linewidth]{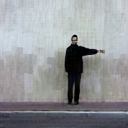} \\
		
		\raisebox{1.7\normalbaselineskip}[0pt][0pt]{\rotatebox[origin=c]{90}{wave2}}
		&\includegraphics[width=.095\linewidth]{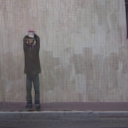}
		&\includegraphics[width=.095\linewidth]{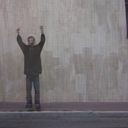}
		&\includegraphics[width=.095\linewidth]{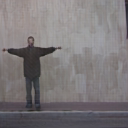} &\includegraphics[width=.095\linewidth]{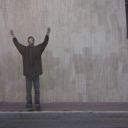} &\includegraphics[width=.095\linewidth]{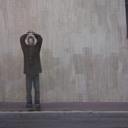} &\includegraphics[width=.095\linewidth]{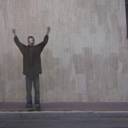} &\includegraphics[width=.095\linewidth]{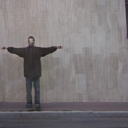} &\includegraphics[width=.095\linewidth]{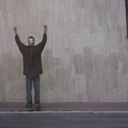} 
		&\includegraphics[width=.095\linewidth]{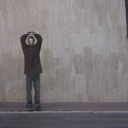} \\
		
		\raisebox{1.7\normalbaselineskip}[0pt][0pt]{\rotatebox[origin=c]{90}{pjump}}
		&\includegraphics[width=.095\linewidth]{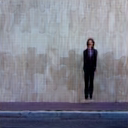}
		&\includegraphics[width=.095\linewidth]{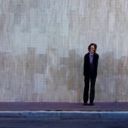}
		&\includegraphics[width=.095\linewidth]{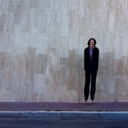} &\includegraphics[width=.095\linewidth]{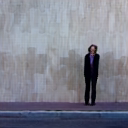} &\includegraphics[width=.095\linewidth]{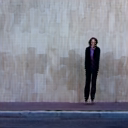} &\includegraphics[width=.095\linewidth]{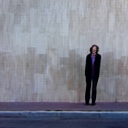} &\includegraphics[width=.095\linewidth]{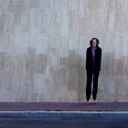} &\includegraphics[width=.095\linewidth]{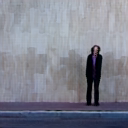} 
		&\includegraphics[width=.095\linewidth]{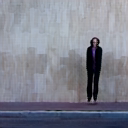} \\
		
		\raisebox{1.7\normalbaselineskip}[0pt][0pt]{\rotatebox[origin=c]{90}{jack}}
		&\includegraphics[width=.095\linewidth]{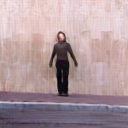}
		&\includegraphics[width=.095\linewidth]{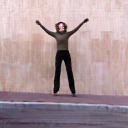}
		&\includegraphics[width=.095\linewidth]{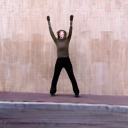} &\includegraphics[width=.095\linewidth]{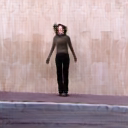} &\includegraphics[width=.095\linewidth]{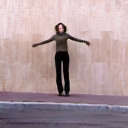} &\includegraphics[width=.095\linewidth]{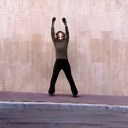} &\includegraphics[width=.095\linewidth]{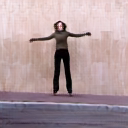} &\includegraphics[width=.095\linewidth]{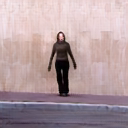} 
		&\includegraphics[width=.095\linewidth]{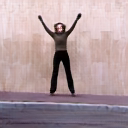} \\

	\end{tabular}
	\caption{Qualitative samples of class-conditioned synthesis without input image on Weizmann ($128 \stimes 128$). Videos are synthesized from the class label alone. The diversity of brightness, contrast and saturation levels reflects the augmentations we use during training. This is necessary to avoid severe overfitting due to the small size of the dataset.}
	\label{fig:weizmann-class-nof-128}
\end{figure}

\paragraph{Class-conditioned synthesis.} Next, we conduct a qualitative study of video synthesis conditioned on a class label on the Weizmann dataset~\citep{gorelick2007actions}, which consists of short clips captured from a fixed viewpoint from $9$ subjects, each performing the same $10$ actions. We use $8$ subjects for training and leave $1$ out for testing. \method{} tackles this task by specifying the class label as a video-level annotation in $\transformer$ (Figure~\ref{fig:gpt}). We can make the held-out subject perform the learned actions accurately given a single input image (Table~\ref{fig:weizmann-class-128}). Or we can synthesize new videos without any prior knowledge other than the desired class label (Table~\ref{fig:weizmann-class-nof-128}).


\paragraph{Qualitative ablation.} We repeat the ablation study of the autoencoder conducted in Table~\ref{tab:ae-ablation} on BAIR~\citep{ebert2017self}, by giving some qualitative samples of the reconstruction this time. The reconstruction results of 30-frame videos given the first frame and the compressed features at subsequent timesteps can be found in Table~\ref{tab:bair-ae-qualitative-ablation-256}. To best compare different ablations of \method{}, we use the same input data for all. A model trained with the $L_1$ loss in RGB space reconstructs the static background and the robotic arm well, while other objects look very blurry. When using the same loss in the feature space of a $\VGG$ net~\citep{simonyan2015very}, objects display higher-frequency details, but look transparent. Having the image discriminator $\imagedis$ allows much more realistic outcomes, although these are still not temporally consistent, nor do they faithfully correspond to the ground-truth video. Having the temporal discriminator $\videodis$ seems to improve over the temporal consistency issue. Thanks to the flow module $\flowestimator$, reconstructed videos are closer to the ground-truth ones, in particular at early timesteps. However, we see that flow imperfections cause synthesis artefacts which add up as time goes by, and result in quality degradation. The self-supervised technique used to learn $\flowestimator$ mitigates this shortcoming. Finally, larger context windows and longer training times help to better recover details (see the more realistic texture and shape of manipulated objects for instance). The final proposal creates high-quality videos almost indistinguishable from real ones, and shows true interactions with objects (a ball and a colorful gun in this example). We see that in case of disocclusion (around $t=20$ the upper right corner is revealed), our model is able to synthesize plausible arrangements of objects despite lossy compression. These new objects remain consistent as the synthesis process unfolds (see $t=30$) thanks to the autoregressive nature of our approach.
\begin{table}
 \setlength\heavyrulewidth{0.25ex}
 \setlength\tabcolsep{2.0pt}
 \aboverulesep=0ex
 \belowrulesep=0.3ex
 \centering
 \caption{Qualitative ablation study of the autoencoder on BAIR ($256\stimes256$). Reconstruction of 30-frame videos given the first frame and the compressed features at subsequent timesteps.}
 \label{tab:bair-ae-qualitative-ablation-256}
 \footnotesize
 \begin{tabular}{@{}M{0.80cm}@{}M{0.80cm}@{}M{0.80cm}@{}M{0.80cm}@{}M{0.80cm}@{}M{0.80cm}@{}M{0.80cm}@{}M{1.90cm}M{1.90cm}M{1.90cm}M{1.90cm}@{}}
 \toprule
 \multicolumn{4}{c}{\footnotesize Self-recovery} & \multicolumn{3}{c}{\footnotesize Ctxt.-recovery} & \multicolumn{4}{c}{\footnotesize Reconstructed frames} \\
 \cmidrule(r){1-4} \cmidrule(lr){5-7} \cmidrule(lr){8-11}
 {\scriptsize RGB} & {\scriptsize VGG} & {\scriptsize $\imagedis$} & {\scriptsize $\videodis$} & {\scriptsize $\flowestimator$} & {\scriptsize Sup.} & {\scriptsize Ctxt.} & {\scriptsize $t=2$} & {\scriptsize $t=10$} & {\scriptsize $t=20$} & {\scriptsize $t=30$} \\
 \midrule
 
 \checkmark &   &   &   &   &   & 0 & \includegraphics[width=0.97\linewidth]{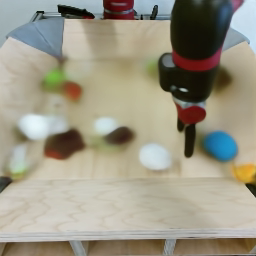} &
 \includegraphics[width=0.97\linewidth]{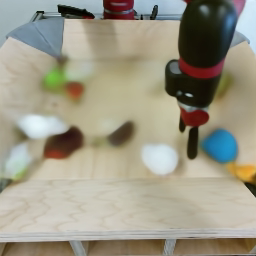} &
 \includegraphics[width=0.97\linewidth]{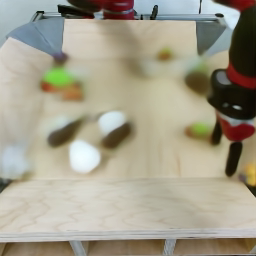} &
 \includegraphics[width=0.97\linewidth]{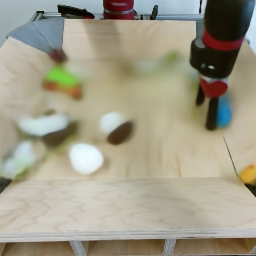}\\
 
 \rowcolor[HTML]{EFEFEF}[0pt][0pt]
 \checkmark & \checkmark &   &  &   &   & 0 &
 \includegraphics[width=0.97\linewidth]{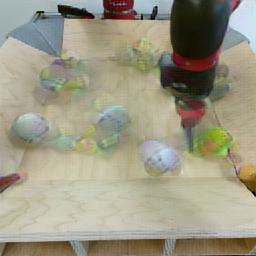} &
 \includegraphics[width=0.97\linewidth]{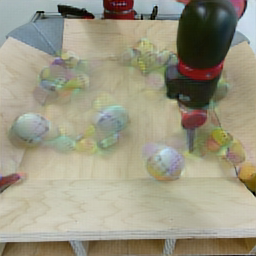} &
 \includegraphics[width=0.97\linewidth]{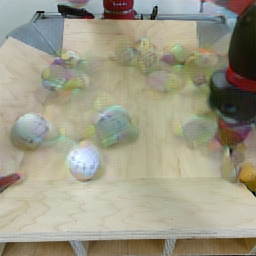} &
 \includegraphics[width=0.97\linewidth]{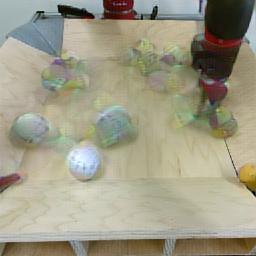}\\
 
 \checkmark & \checkmark & \checkmark   &   &   &   & 0 &
 \includegraphics[width=0.97\linewidth]{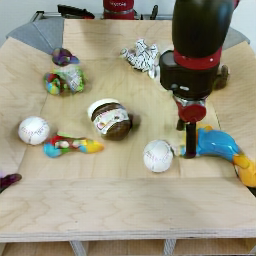} &
 \includegraphics[width=0.97\linewidth]{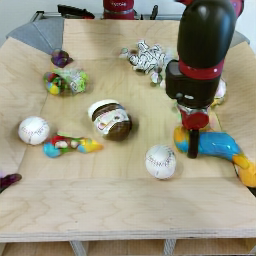} &
 \includegraphics[width=0.97\linewidth]{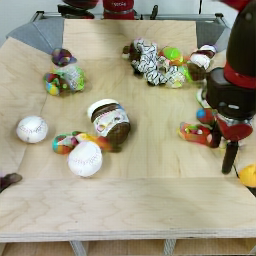} &
 \includegraphics[width=0.97\linewidth]{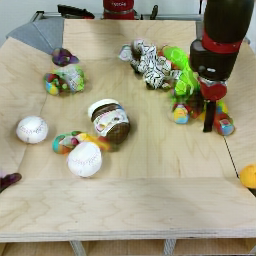}\\

 \rowcolor[HTML]{EFEFEF}[0pt][0pt]
 \checkmark & \checkmark & \checkmark & \checkmark   &   &   & 0 &
 \includegraphics[width=0.97\linewidth]{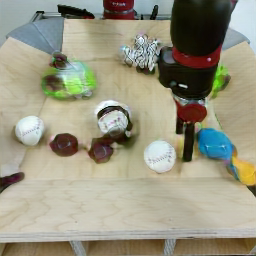} &
 \includegraphics[width=0.97\linewidth]{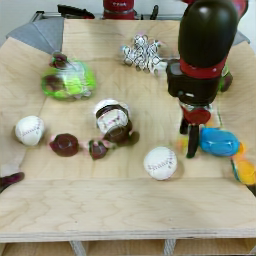} &
 \includegraphics[width=0.97\linewidth]{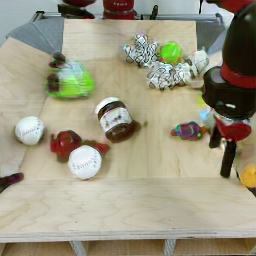} &
 \includegraphics[width=0.97\linewidth]{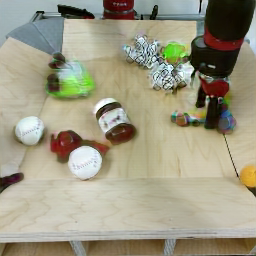}\\
 
 \checkmark & \checkmark & \checkmark & \checkmark & \checkmark &   & 1 &
 \includegraphics[width=0.97\linewidth]{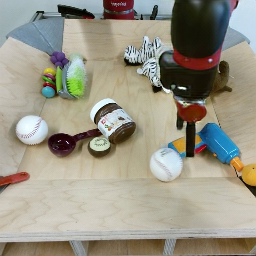} &
 \includegraphics[width=0.97\linewidth]{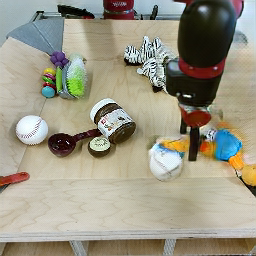} &
 \includegraphics[width=0.97\linewidth]{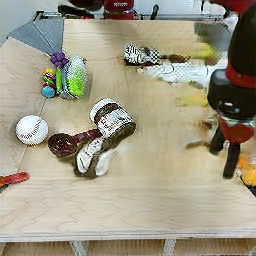} &
 \includegraphics[width=0.97\linewidth]{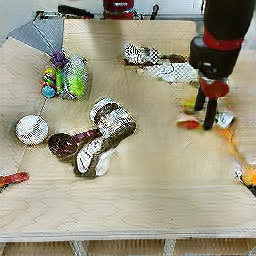}\\
 
 \rowcolor[HTML]{EFEFEF}[0pt][0pt]
 \checkmark & \checkmark & \checkmark & \checkmark & \checkmark & \checkmark   & 1 &
 \includegraphics[width=0.97\linewidth]{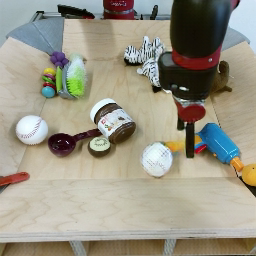} &
 \includegraphics[width=0.97\linewidth]{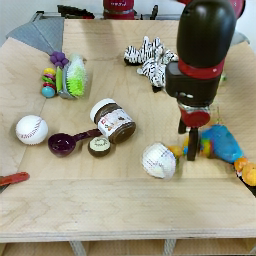} &
 \includegraphics[width=0.97\linewidth]{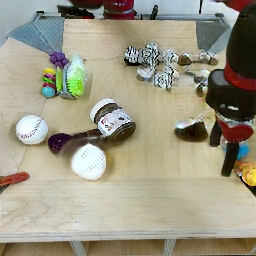} &
 \includegraphics[width=0.97\linewidth]{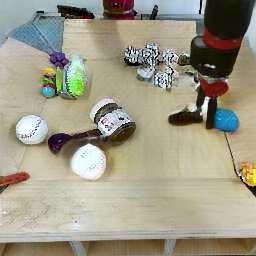}\\
 
 \checkmark & \checkmark & \checkmark & \checkmark & \checkmark & \checkmark   & 8 &
 \includegraphics[width=0.97\linewidth]{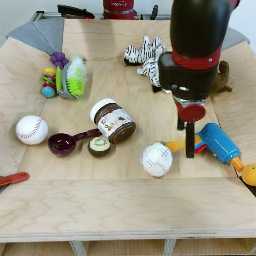} &
 \includegraphics[width=0.97\linewidth]{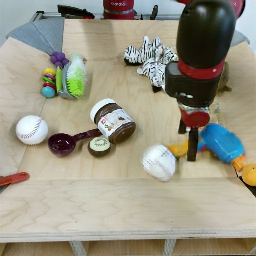} &
 \includegraphics[width=0.97\linewidth]{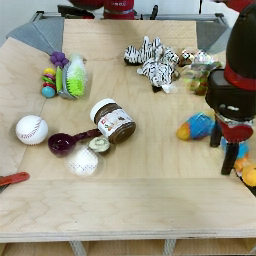} &
 \includegraphics[width=0.97\linewidth]{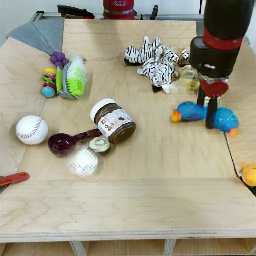}\\
 
 \rowcolor[HTML]{EFEFEF}[0pt][0pt]
 \checkmark & \checkmark & \checkmark & \checkmark & \checkmark & \checkmark   & 15 &
 \includegraphics[width=0.97\linewidth]{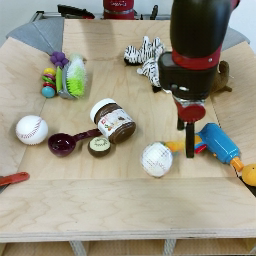} &
 \includegraphics[width=0.97\linewidth]{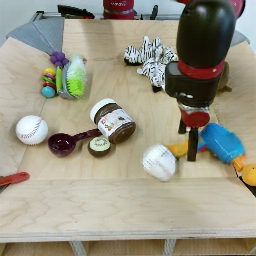} &
 \includegraphics[width=0.97\linewidth]{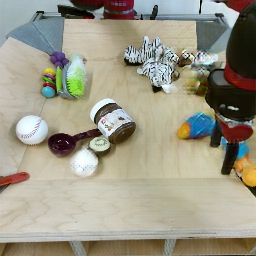} &
 \includegraphics[width=0.97\linewidth]{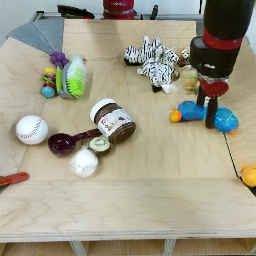}\\
 
 \multicolumn{7}{@{}c@{}}{\textit{Training longer ~~~~ (num. epochs $\stimes \, 3$)}} &
 \includegraphics[width=0.97\linewidth]{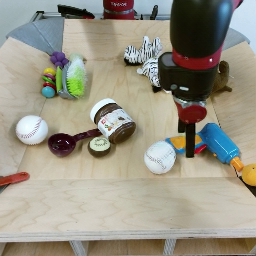} &
 \includegraphics[width=0.97\linewidth]{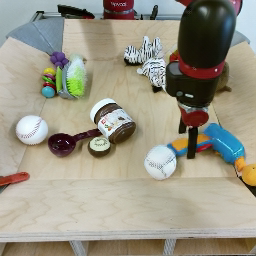} &
 \includegraphics[width=0.97\linewidth]{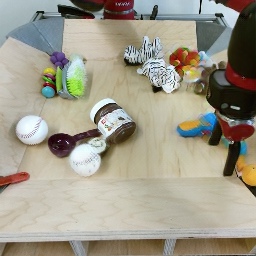} &
 \includegraphics[width=0.97\linewidth]{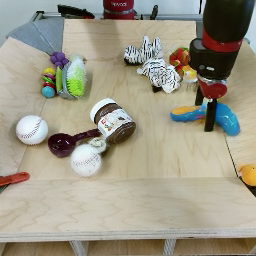}\\
 
 \midrule
 
 \multicolumn{7}{@{}c@{}}{\textit{Oracle}} &
 \includegraphics[width=0.97\linewidth]{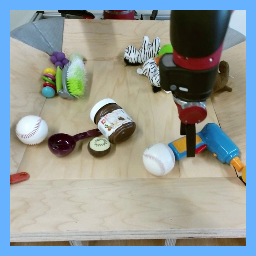} &
 \includegraphics[width=0.97\linewidth]{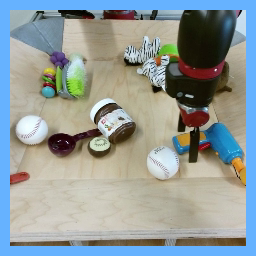} &
 \includegraphics[width=0.97\linewidth]{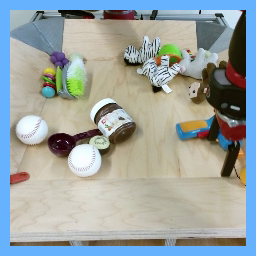} &
 \includegraphics[width=0.97\linewidth]{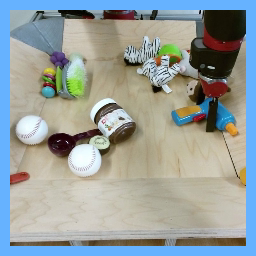}\\
 
 \bottomrule
 \multicolumn{11}{c}{\scriptsize ``Sup.'': self-supervision of $\flowestimator$; \hspace{2ex} ``Ctxt.'': number of context frames taken into account (in $\flowestimator$) when decoding current frame (in $\decoder$).} \\
 \end{tabular}
\end{table}

\begin{figure}
	\setlength\tabcolsep{1.0pt}
	\renewcommand{\arraystretch}{1.0}
	\footnotesize
	\begin{tabular}{ccccccccccc}
	    \multicolumn{11}{c}{\textbf{(a)} BAIR ($256 \times 256$).}\\
	    & $t=10$ & $t=20$ & $t=30$ & $t=40$ & $t=50$ & $t=60$ & $t=70$ & $t=80$ & $t=90$ \\
	    [0.05cm]
		\raisebox{1.75\normalbaselineskip}[0pt][0pt]{\rotatebox[origin=c]{90}{$\textrm{Ctxt.}=0$}} &
		\includegraphics[width=.100\linewidth]{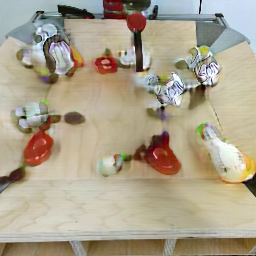} &
		\includegraphics[width=.100\linewidth]{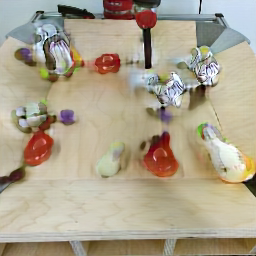} &
		\includegraphics[width=.100\linewidth]{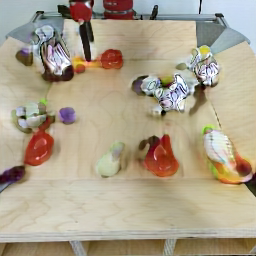} &
		\includegraphics[width=.100\linewidth]{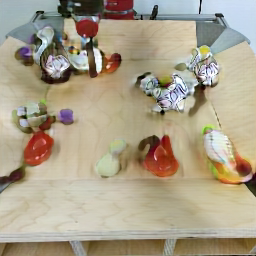} &
		\includegraphics[width=.100\linewidth]{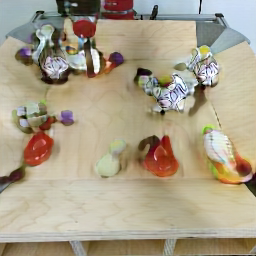} &
		\includegraphics[width=.100\linewidth]{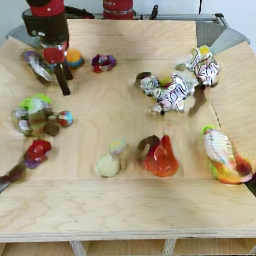} &
		\includegraphics[width=.100\linewidth]{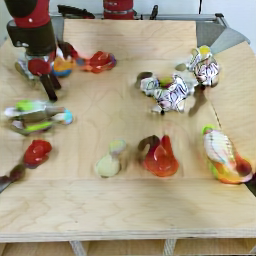} &
		\includegraphics[width=.100\linewidth]{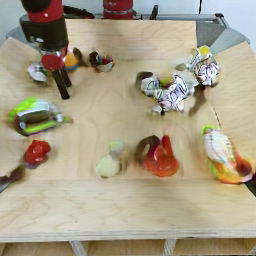} &
		\includegraphics[width=.100\linewidth]{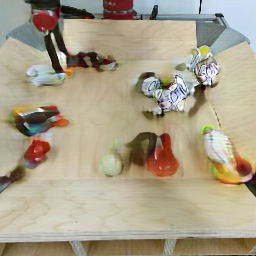} \\
		\raisebox{1.75\normalbaselineskip}[0pt][0pt]{\rotatebox[origin=c]{90}{$\textrm{Ctxt.}=1$}} &
		\includegraphics[width=.100\linewidth]{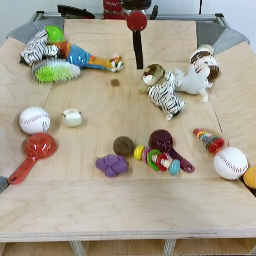} &
		\includegraphics[width=.100\linewidth]{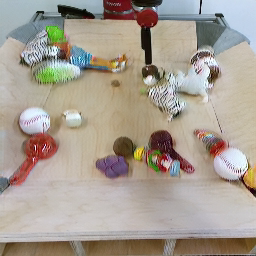} &
		\includegraphics[width=.100\linewidth]{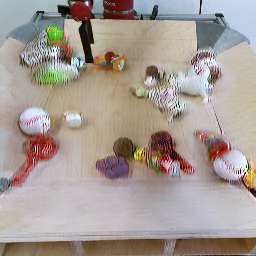} &
		\includegraphics[width=.100\linewidth]{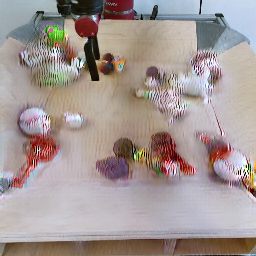} &
		\includegraphics[width=.100\linewidth]{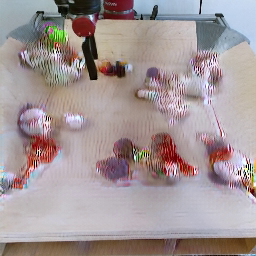} &
		\includegraphics[width=.100\linewidth]{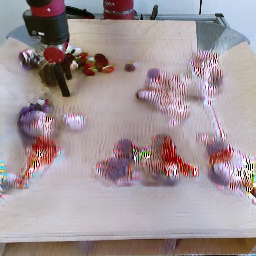} &
		\includegraphics[width=.100\linewidth]{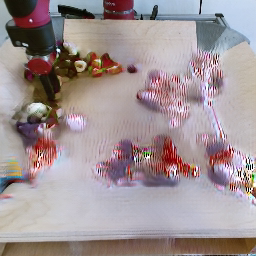} &
		\includegraphics[width=.100\linewidth]{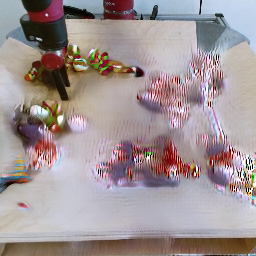} &
		\includegraphics[width=.100\linewidth]{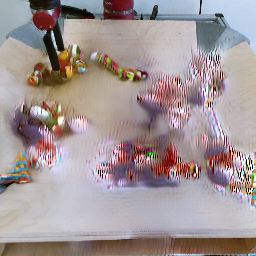} \\
		\raisebox{1.75\normalbaselineskip}[0pt][0pt]{\rotatebox[origin=c]{90}{$\textrm{Ctxt.}=4$}} &
		\includegraphics[width=.100\linewidth]{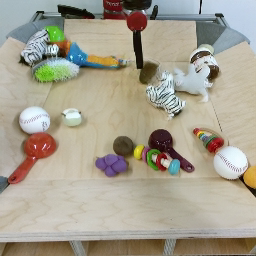} &
		\includegraphics[width=.100\linewidth]{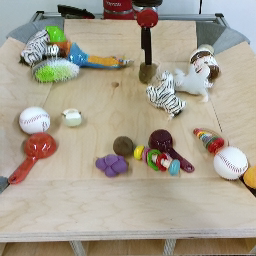} &
		\includegraphics[width=.100\linewidth]{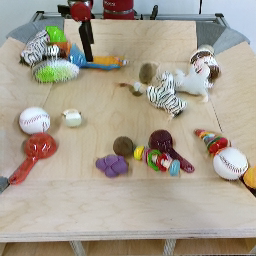} &
		\includegraphics[width=.100\linewidth]{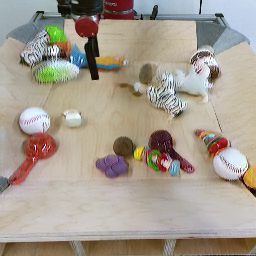} &
		\includegraphics[width=.100\linewidth]{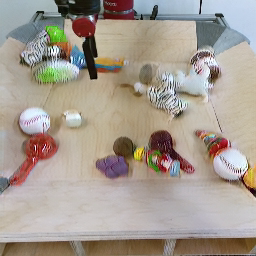} &
		\includegraphics[width=.100\linewidth]{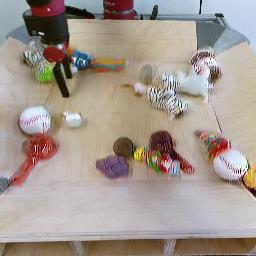} &
		\includegraphics[width=.100\linewidth]{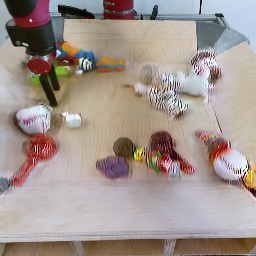} &
		\includegraphics[width=.100\linewidth]{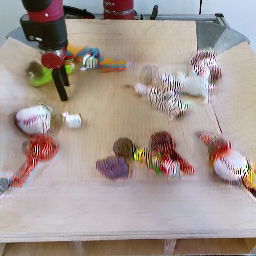} &
		\includegraphics[width=.100\linewidth]{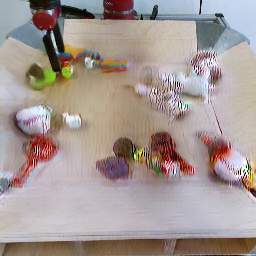} \\
		\raisebox{1.75\normalbaselineskip}[0pt][0pt]{\rotatebox[origin=c]{90}{$\textrm{Ctxt.}=16$}} &
		\includegraphics[width=.100\linewidth]{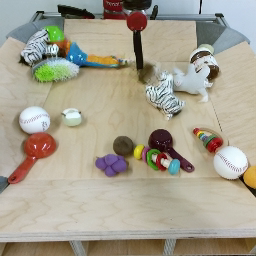} &
		\includegraphics[width=.100\linewidth]{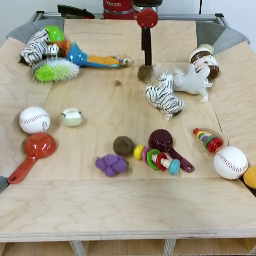} &
		\includegraphics[width=.100\linewidth]{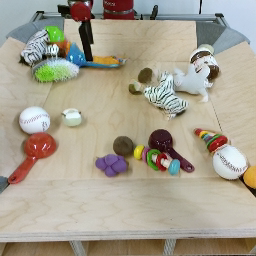} &
		\includegraphics[width=.100\linewidth]{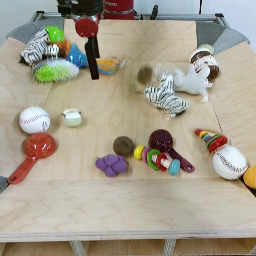} &
		\includegraphics[width=.100\linewidth]{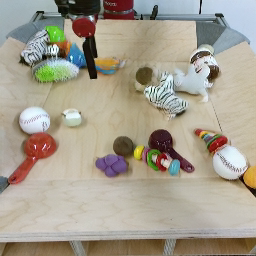} &
		\includegraphics[width=.100\linewidth]{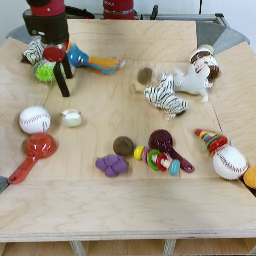} &
		\includegraphics[width=.100\linewidth]{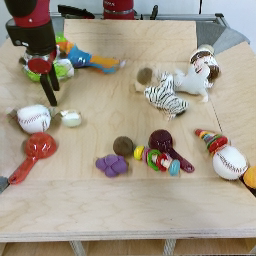} &
		\includegraphics[width=.100\linewidth]{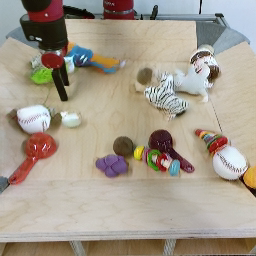} &
		\includegraphics[width=.100\linewidth]{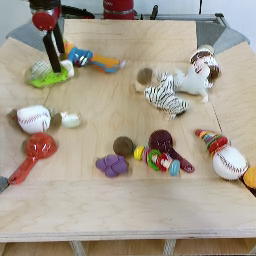} \\
		[0.3cm]
		
		\multicolumn{11}{c}{\textbf{(b)} Kinetics ($64 \times 64$).}\\
	    & $t=10$ & $t=20$ & $t=30$ & $t=40$ & $t=50$ & $t=60$ & $t=70$ & $t=80$ & $t=90$ \\
	    [0.05cm]
		\raisebox{1.75\normalbaselineskip}[0pt][0pt]{\rotatebox[origin=c]{90}{$\textrm{Ctxt.}=0$}} &
		\includegraphics[width=.100\linewidth]{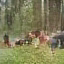} &
		\includegraphics[width=.100\linewidth]{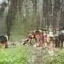} &
		\includegraphics[width=.100\linewidth]{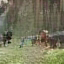} &
		\includegraphics[width=.100\linewidth]{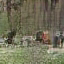} &
		\includegraphics[width=.100\linewidth]{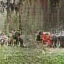} &
		\includegraphics[width=.100\linewidth]{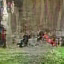} &
		\includegraphics[width=.100\linewidth]{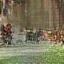} &
		\includegraphics[width=.100\linewidth]{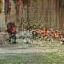} &
		\includegraphics[width=.100\linewidth]{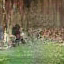} \\
		\raisebox{1.75\normalbaselineskip}[0pt][0pt]{\rotatebox[origin=c]{90}{$\textrm{Ctxt.}=1$}} &
		\includegraphics[width=.100\linewidth]{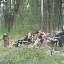} &
		\includegraphics[width=.100\linewidth]{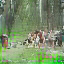} &
		\includegraphics[width=.100\linewidth]{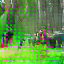} &
		\includegraphics[width=.100\linewidth]{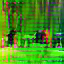} &
		\includegraphics[width=.100\linewidth]{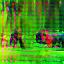} &
		\includegraphics[width=.100\linewidth]{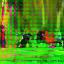} &
		\includegraphics[width=.100\linewidth]{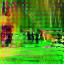} &
		\includegraphics[width=.100\linewidth]{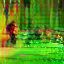} &
		\includegraphics[width=.100\linewidth]{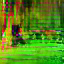} \\
		\raisebox{1.75\normalbaselineskip}[0pt][0pt]{\rotatebox[origin=c]{90}{$\textrm{Ctxt.}=4$}} &
		\includegraphics[width=.100\linewidth]{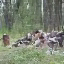} &
		\includegraphics[width=.100\linewidth]{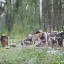} &
		\includegraphics[width=.100\linewidth]{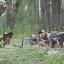} &
		\includegraphics[width=.100\linewidth]{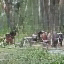} &
		\includegraphics[width=.100\linewidth]{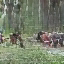} &
		\includegraphics[width=.100\linewidth]{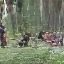} &
		\includegraphics[width=.100\linewidth]{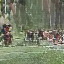} &
		\includegraphics[width=.100\linewidth]{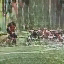} &
		\includegraphics[width=.100\linewidth]{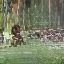} \\
		\raisebox{1.75\normalbaselineskip}[0pt][0pt]{\rotatebox[origin=c]{90}{$\textrm{Ctxt.}=16$}} &
		\includegraphics[width=.100\linewidth]{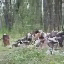} &
		\includegraphics[width=.100\linewidth]{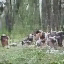} &
		\includegraphics[width=.100\linewidth]{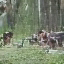} &
		\includegraphics[width=.100\linewidth]{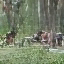} &
		\includegraphics[width=.100\linewidth]{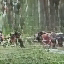} &
		\includegraphics[width=.100\linewidth]{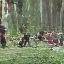} &
		\includegraphics[width=.100\linewidth]{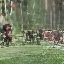} &
		\includegraphics[width=.100\linewidth]{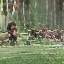} &
		\includegraphics[width=.100\linewidth]{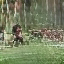} \\
		[0.3cm]
		
		\multicolumn{11}{c}{\textbf{(c)} AudioSet-Drums ($128 \times 128$).}\\
	    & $t=10$ & $t=20$ & $t=30$ & $t=40$ & $t=50$ & $t=60$ & $t=70$ & $t=80$ & $t=90$ \\
	    [0.05cm]
		\raisebox{1.75\normalbaselineskip}[0pt][0pt]{\rotatebox[origin=c]{90}{$\textrm{Ctxt.}=0$}} &
		\includegraphics[width=.100\linewidth]{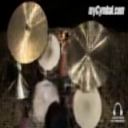} &
		\includegraphics[width=.100\linewidth]{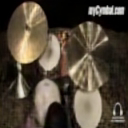} &
		\includegraphics[width=.100\linewidth]{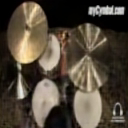} &
		\includegraphics[width=.100\linewidth]{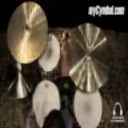} &
		\includegraphics[width=.100\linewidth]{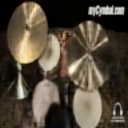} &
		\includegraphics[width=.100\linewidth]{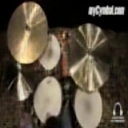} &
		\includegraphics[width=.100\linewidth]{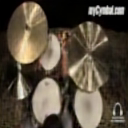} &
		\includegraphics[width=.100\linewidth]{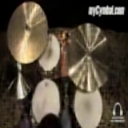} &
		\includegraphics[width=.100\linewidth]{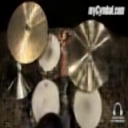} \\
		\raisebox{1.75\normalbaselineskip}[0pt][0pt]{\rotatebox[origin=c]{90}{$\textrm{Ctxt.}=1$}} &
		\includegraphics[width=.100\linewidth]{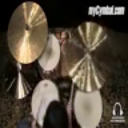} &
		\includegraphics[width=.100\linewidth]{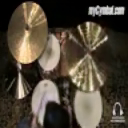} &
		\includegraphics[width=.100\linewidth]{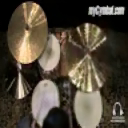} &
		\includegraphics[width=.100\linewidth]{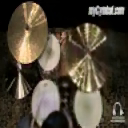} &
		\includegraphics[width=.100\linewidth]{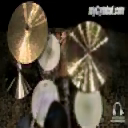} &
		\includegraphics[width=.100\linewidth]{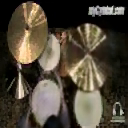} &
		\includegraphics[width=.100\linewidth]{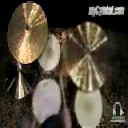} &
		\includegraphics[width=.100\linewidth]{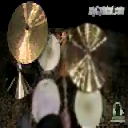} &
		\includegraphics[width=.100\linewidth]{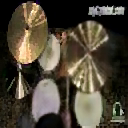} \\
		\raisebox{1.75\normalbaselineskip}[0pt][0pt]{\rotatebox[origin=c]{90}{$\textrm{Ctxt.}=4$}} &
		\includegraphics[width=.100\linewidth]{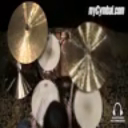} &
		\includegraphics[width=.100\linewidth]{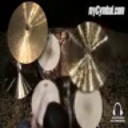} &
		\includegraphics[width=.100\linewidth]{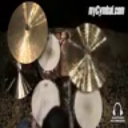} &
		\includegraphics[width=.100\linewidth]{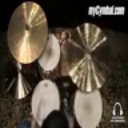} &
		\includegraphics[width=.100\linewidth]{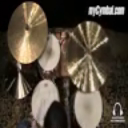} &
		\includegraphics[width=.100\linewidth]{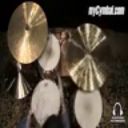} &
		\includegraphics[width=.100\linewidth]{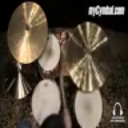} &
		\includegraphics[width=.100\linewidth]{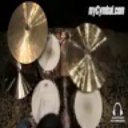} &
		\includegraphics[width=.100\linewidth]{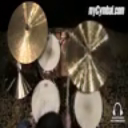} \\
		\raisebox{1.75\normalbaselineskip}[0pt][0pt]{\rotatebox[origin=c]{90}{$\textrm{Ctxt.}=16$}} &
		\includegraphics[width=.100\linewidth]{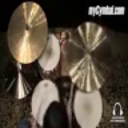} &
		\includegraphics[width=.100\linewidth]{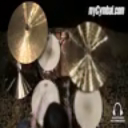} &
		\includegraphics[width=.100\linewidth]{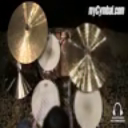} &
		\includegraphics[width=.100\linewidth]{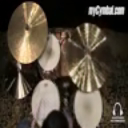} &
		\includegraphics[width=.100\linewidth]{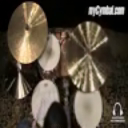} &
		\includegraphics[width=.100\linewidth]{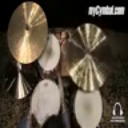} &
		\includegraphics[width=.100\linewidth]{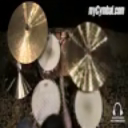} &
		\includegraphics[width=.100\linewidth]{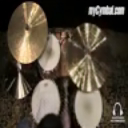} &
		\includegraphics[width=.100\linewidth]{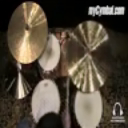} \\
	\end{tabular}
	\caption{\revision{Qualitative samples for long-term generalization of a model trained on 16-frame videos to the synthesis of 90-frame ones conditioned on one frame. We compare different context windows (``Ctx.'') for the flow module: 0 (not using the flow), 1, 4 and 16. Zoom in for details.}}
	\label{fig:long-term}
\end{figure}

\paragraph{Long-term synthesis.} \revision{We assess the ability of \method{} to generalize to long-term prediction through a qualitative study. We compare, for a model trained on sequences of length 16, the synthesis of 90-frame videos conditioned on a single frame and using different context windows for the flow module: 0 (not using the flow), 1, 4 and 16. Results on BAIR, Kinetics and AudioSet-Drums are available in Figure~\ref{fig:long-term}.  When the flow is not used (size 0, \ie, a naïve extension of~\citep{esser2020taming} or a setup close to~\citep{rakhimov2020latent}) videos are quite unstable and we see colors shifting over time. With a single-frame context window (size 1), there is a greater temporal consistency (\eg, the drum sticks are better preserved on AudioSet-Drums, the semantic structure shows higher fidelity to the input frame on BAIR or Kinetics), but in the long-term, synthesis artefacts add up due to the autoregressive decoding which leads to visual deterioration and saturation. These effects stand out the most on BAIR and Kinetics. The proposed multi-frame context extension (Appendix~\ref{sec:extension}) reduces this issue (size 4), and the latter becomes barely perceptible with an even larger context window (size 16). We found that even small context windows on AudioSet-Drums are sufficient thanks to the high inter-frame similarity.}

\section{Augmentation strategy}
\label{sec:augmentation}
\setcounter{table}{0}
\renewcommand{\thetable}{F\arabic{table}}
\setcounter{figure}{0}
\renewcommand{\thefigure}{F\arabic{figure}}

\begin{figure}
	\setlength\tabcolsep{2.0pt}
	\renewcommand{\arraystretch}{1.0}
	\footnotesize
	\begin{tabular}{C{0.10cm}ccccc}
	    && Context & Image & Mask & Flow \\
	    [0.05cm]
		
		\raisebox{1.75\normalbaselineskip}[0pt][0pt]{\rotatebox[origin=c]{90}{Real}} & 
		~ &
		\includegraphics[width=.230\linewidth]{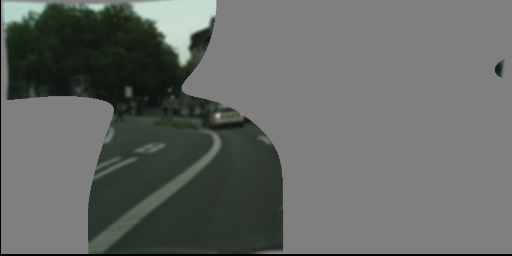} &
		\includegraphics[width=.230\linewidth]{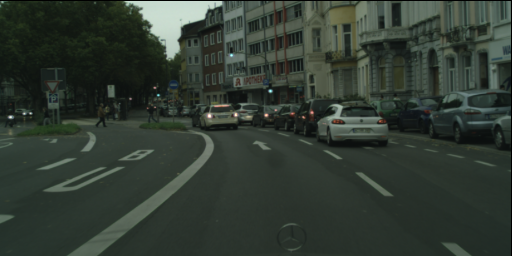} &
		\includegraphics[width=.230\linewidth]{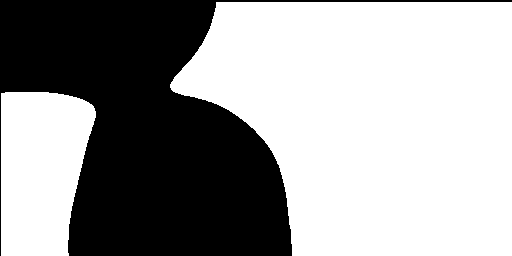} &
		\includegraphics[width=.230\linewidth]{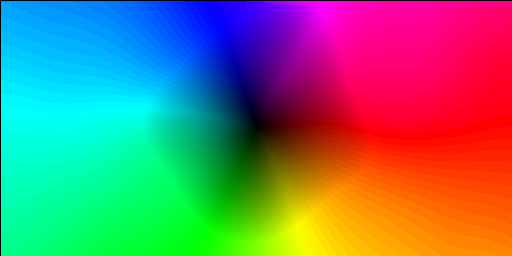} \\
		
		\raisebox{1.85\normalbaselineskip}[0pt][0pt]{\rotatebox[origin=c]{90}{Synthetic}} & 
		~ &
		\includegraphics[width=.230\linewidth]{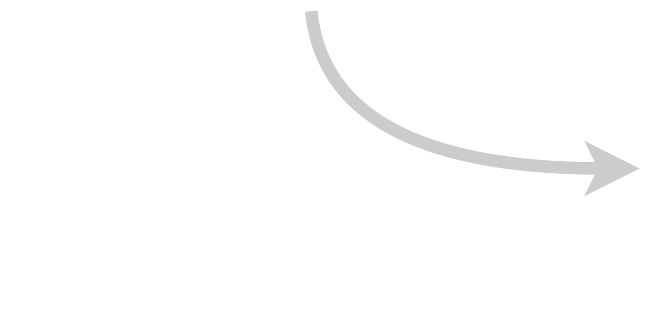} &
		\includegraphics[width=.230\linewidth]{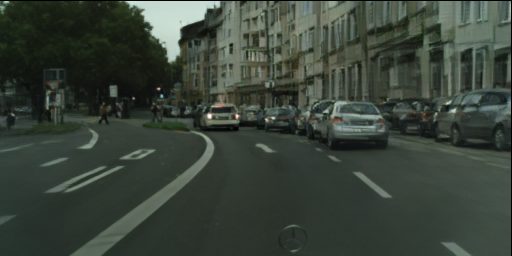} &
		\includegraphics[width=.230\linewidth]{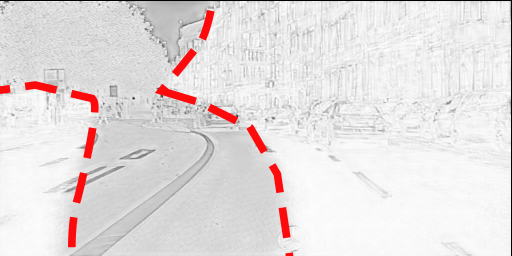} &
		\includegraphics[width=.230\linewidth]{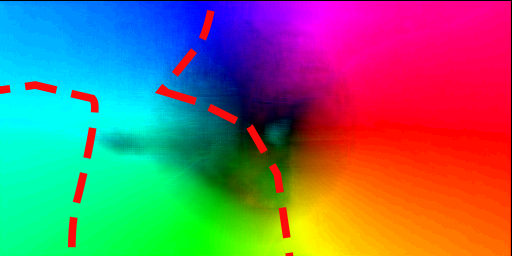} \\
		[0.3cm]
		
		\raisebox{1.75\normalbaselineskip}[0pt][0pt]{\rotatebox[origin=c]{90}{Real}} & 
		~ &
		\includegraphics[width=.230\linewidth]{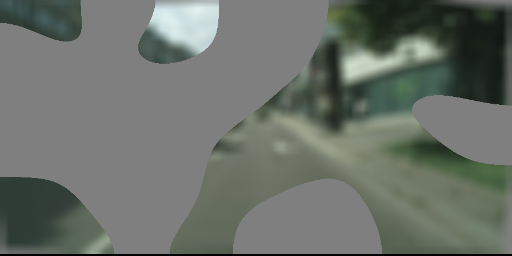} &
		\includegraphics[width=.230\linewidth]{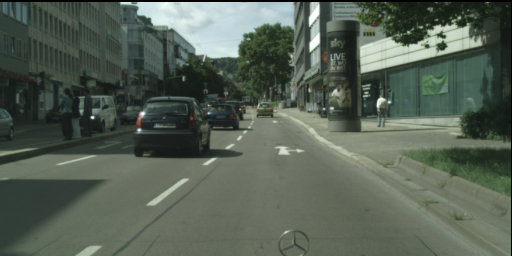} &
		\includegraphics[width=.230\linewidth]{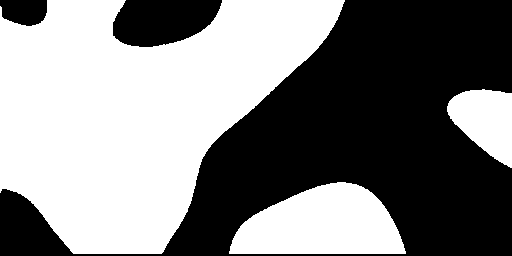} &
		\includegraphics[width=.230\linewidth]{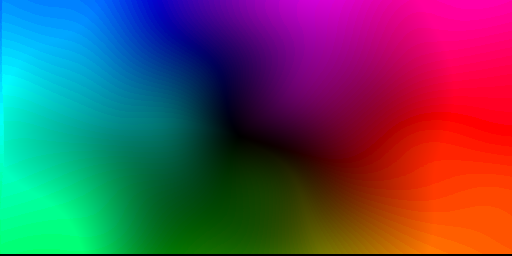} \\
		
		\raisebox{1.85\normalbaselineskip}[0pt][0pt]{\rotatebox[origin=c]{90}{Synthetic}} & 
		~ &
		\includegraphics[width=.230\linewidth]{neurips2021/figures/sup/aug/arrow.pdf} &
		\includegraphics[width=.230\linewidth]{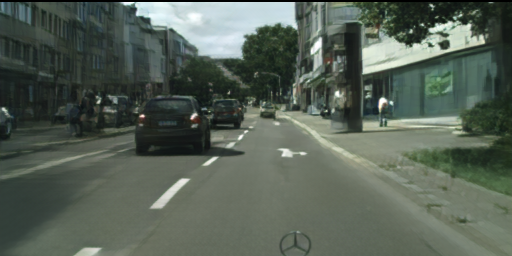} &
		\includegraphics[width=.230\linewidth]{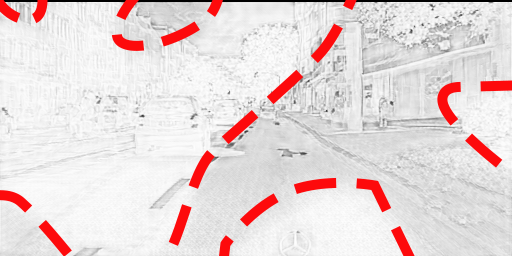} &
		\includegraphics[width=.230\linewidth]{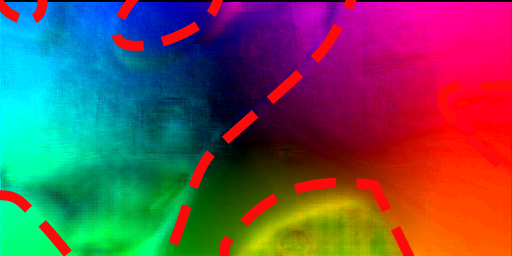} \\
		[0.3cm]
		
		\raisebox{1.75\normalbaselineskip}[0pt][0pt]{\rotatebox[origin=c]{90}{Real}} & 
		~ &
		\includegraphics[width=.230\linewidth]{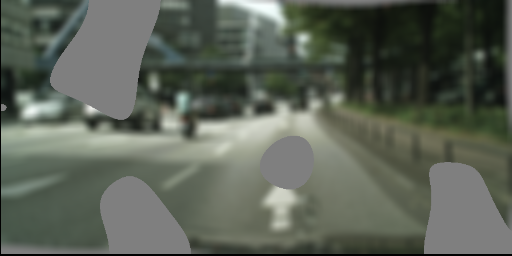} &
		\includegraphics[width=.230\linewidth]{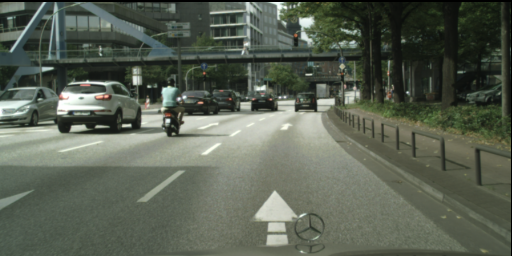} &
		\includegraphics[width=.230\linewidth]{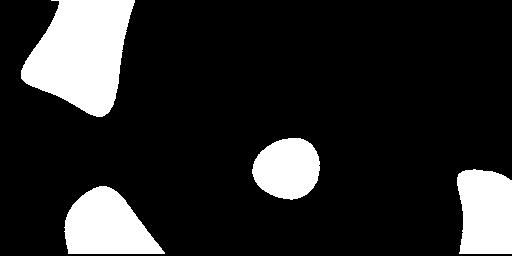} &
		\includegraphics[width=.230\linewidth]{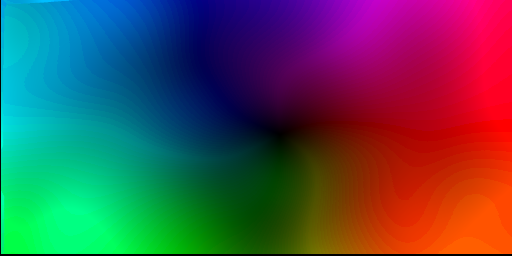} \\
		
		\raisebox{1.85\normalbaselineskip}[0pt][0pt]{\rotatebox[origin=c]{90}{Synthetic}} & 
		~ &
		\includegraphics[width=.230\linewidth]{neurips2021/figures/sup/aug/arrow.pdf} &
		\includegraphics[width=.230\linewidth]{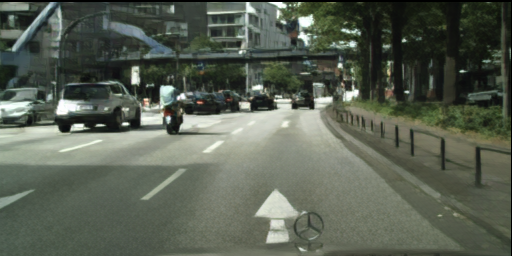} &
		\includegraphics[width=.230\linewidth]{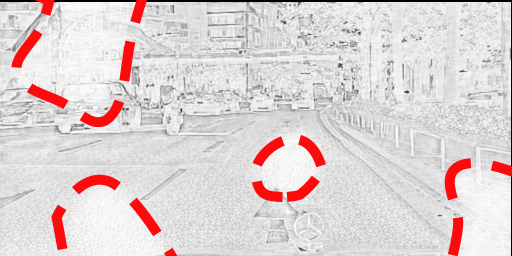} &
		\includegraphics[width=.230\linewidth]{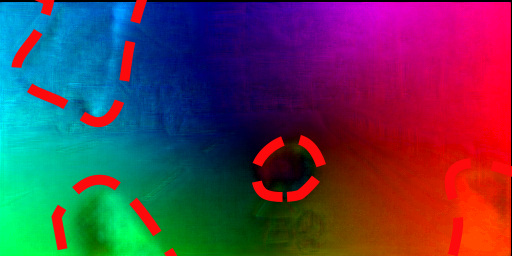} \\
	\end{tabular}
	\caption{Some examples of reconstructed image, fusion mask and optical flow from self-augmented views on Cityscapes ($256 \times 512$). For better visualization, we use a dashed red line in the synthetic mask and flow to indicate the outline of the real mask. Training losses encourage real and synthetic masks to be close within occluded context areas (white region), and real and synthetic flows to be close within the visible context areas (black region).}
	\label{fig:city-aug-256}
\end{figure}

As discussed in Section~\ref{sec:ae}, the self-supervision process used to train $\flowestimator$ reconstructs a static image $x$ from both its latent representation and an augmented view as context by estimating the flow and mask corresponding to that augmentation. We illustrate this process on the Cityscapes dataset in Figure~\ref{fig:city-aug-256}. We recall that, in this case, the context image can be writen as $x_c=o_c \otimes \blur(\aug(x))$, where $o_c$ is an occlusion mask, and $\blur$ and $\aug$ some blurring and augmentation (\ie, spatial transformation) functions respectively. We can invert the flow corresponding to transformation $\aug$ thanks to Algorithm~\ref{alg:flow-invert}. The output corresponds to the target ``Real Flow'' in the illustration. As for the target ``Real Mask'', it is obtained by warping the occlusion mask with the inverted flow. Looking at synthetic reconstructions, we see that visible details in the context are well recovered despite the augmentation. Indeed, real and synthetic flows match in non occluded areas (indicated by the dashed red line). Moreover, adversarial training produces high-quality images from blurred and occluded contexts. We notice that textures do not always match in real and synthetic images (\eg, building windows, car appearance). Yet, thanks to recovery losses, the semantic structure of the scene is preserved. Finally, we see that darker and lighter areas of the synthetic masks correspond to the same areas in the real ones. The darker the synthetic masks, the more features are updated from context. The \textit{whitish} part indicates that in occluded areas the final image is reconstructed from its latent representation alone, and the \textit{greyish} part shows that for other regions context is effectively reused (and latent representations still contribute to the final outcome). It is interesting to see that edges are visible in the synthetic masks. This indicates that those regions may be easier to detect (darker color) and play an important role in the image alignment process (estimation of the flow).

\end{document}